\pdfoutput=1
\documentclass[12pt,a4paper]{article}
\usepackage{siunitx}
\usepackage[english]{babel}
\usepackage{array}
\usepackage{booktabs}
\usepackage{tabularx}
\usepackage[table,xcdraw]{xcolor}
\usepackage{tcolorbox}
\usepackage{verbatim}
\usepackage[a4paper,top=2cm,bottom=2cm,left=2.5cm,right=2.5cm,marginparwidth=1.75cm]{geometry}
\usepackage{multirow}
\usepackage{array}
\usepackage{longtable}
\usepackage{pifont}
\newcommand{\cmark}{\textcolor{green!60!black}{\ding{51}}}
\newcommand{\xmark}{\textcolor{red}{\ding{55}}}
\usepackage{amsthm}

\theoremstyle{definition}

\usepackage{tcolorbox}

\usepackage{graphicx}
\usepackage{caption}
\usepackage{float}

\usepackage{graphicx}
\usepackage[normalem]{ulem}
\useunder{\uline}{\ul}{}
\usepackage{colortbl}
\usepackage{amsmath}
\usepackage{graphicx}
\usepackage[colorlinks=true, allcolors=blue, hyperfootnotes=false]{hyperref}
\usepackage{hyperref}
\usepackage[title]{appendix}
\usepackage{mathrsfs}
\usepackage{amsfonts}
\usepackage{booktabs} 
\usepackage{caption}  
\usepackage{threeparttable} 
\usepackage{algorithm}
\usepackage{algorithmicx}
\usepackage{algpseudocode}
\usepackage{listings}
\usepackage{hyperref}
\usepackage{enumitem}
\usepackage{chngcntr}
\usepackage{booktabs}
\usepackage{lipsum}
\usepackage{subcaption}
\usepackage{authblk}
\usepackage[T1]{fontenc}    
\usepackage{csquotes}       
\usepackage{diagbox}
\usepackage{tabularray}
\usepackage{geometry}
\usepackage{setspace}
\usepackage{makecell} 

\usepackage{titlesec}
\titleformat{\section} 
  {\normalfont\Large\bfseries}{\thesection.}{1em}{}

\usepackage{lineno} 

\rightlinenumbers 
\usepackage{booktabs}         
\usepackage{threeparttable}   
\usepackage{multirow}         
\linenumbers 
\usepackage{tabularx}
\usepackage{adjustbox}

\usepackage{float}   
\usepackage{caption} 
\usepackage[numbers]{natbib}


\nolinenumbers

\title{\large Bridging Online Behavior and Clinical Insight: A Longitudinal LLM-based Study of Suicidality on YouTube Reveals Novel Digital Markers }

\author[1]{\normalsize Ilanit Sobol \footnote{Corresponding author: ilanit1997@gmail.com}}
\author[1]{Shir Lissak}
\author[1]{Refael Tikochinski}
\author[2]{Tal Nakash}
\author[2]{Anat Brunstein Klomek}
\author[1]{Eyal Fruchter}
\author[1]{Roi Reichart}

\affil[1]{\small Technion – Israel Institute of Technology} \affil[2]{Reichman University}

\date{}  

\begin{document}
\maketitle

\newcommand{\myfootnote}{\footnote{\color{black}{Odds ratios (OR) quantify the change in the odds of topic prevalence \textit{after} versus \textit{before} the suicide attempt within the group of YouTubers who attempted suicide.}}}

\begin{abstract}

Suicide remains a leading cause of death in Western countries. As social media becomes central to daily life, digital footprints offer valuable insight into suicidal behavior.  Focusing on individuals who attempted suicide while uploading videos to their channels, we investigate:  \textit{How do linguistic patterns on YouTube reflect suicidal behavior, and how do these patterns align with or differ from expert knowledge?}

We examined linguistic changes around suicide attempts and compared individuals who attempted suicide while actively uploading to their channel with three control groups: those with prior attempts, those experiencing major life events, and matched individuals from the broader cohort.
Applying complementary bottom-up, hybrid, and expert-driven approaches, we analyzed a novel longitudinal dataset of 181 suicide-attempt channels and 134 controls.

In the \textbf{bottom-up} analysis, LLM-based topic-modeling identified 166 topics; five were linked to suicide attempts, two also showed attempt-related temporal changes (\textit{Mental Health Struggles}, $OR = 1.74$; \textit{YouTube Engagement}, $OR = 1.67$; $p < .01$)\myfootnote{}.

In the \textbf{hybrid} approach, clinical experts reviewed LLM-derived topics and flagged 19 as suicide-related. However, none showed significant effects beyond those identified bottom-up. \textit{YouTube Engagement}, a platform-specific indicator, was not flagged, underscoring the value of bottom-up discovery.
A \textbf{top-down} psychological assessment of suicide narratives revealed differing motivations: individuals describing prior attempts aimed to \textit{help others} ($\beta=-1.69$, $p<.01$), whereas those attempted during the uploading period emphasized \textit{personal recovery} ($\beta=1.08$, $p<.01$).

By integrating these approaches, we offer a nuanced understanding of suicidality, bridging digital behavior and clinical insights.

\end{abstract}

\newpage
\section{Introduction}
\label{sec:intro}

Suicide remains one of the most significant public health challenges, particularly among young people in western countries, where it consistently ranks as one of the leading causes of death \citep{naghavi2019global, glenn2020annual}. Traditionally, research on suicidality has focused on factors such as mental health issues (e.g., depression), social isolation, and family dynamics \citep{fergusson2000risk, pelkonen2003child, franklin2017risk}, as well as short-term affective fluctuations preceding suicidal behavior \citep{bagge2017trajectories}. These factors were predominantly predefined by experts and were typically examined by relying on patients' self-reports. However, as social media becomes increasingly prevalent and suicide rates continue to grow, established risk factors and their manifestation are changing, requiring further investigation into suicidal behavior in digital spaces \citep{castillo2020suicide, rabani2023detecting}.

Social media have become integral to communication and social development, especially among adolescents \citep{o2011impact}. Many young people turn to these platforms to share their experiences and seek support \citep{fu2013responses, lissak2024colorful}, potentially providing a digital footprint, documenting the nuances of everyday life, including moments of joy and distress \citep{de2013predicting}.

The intersection of suicide, social media, and Artificial Intelligence (AI) has gained traction in recent years, with much research utilizing natural language processing (NLP),  and especially Language Models (LM), to analyze large volumes of social media data \citep{ophir2020deep, lissak2024bored}. However, a common limitation is a reliance on unvalidated suicide risk assessments, such as identifying risk based on messages like "I want to kill myself" or user participation in suicide-related sub-reddits, which introduces biases in both data collection and reliability. Additionally, many studies lack model transparency and have inadequate control groups. Perhaps most importantly, they often fail to incorporate longitudinal analyses that are essential for capturing temporal shifts in relation to suicidal behavior.

Although social media platforms (e.g., Facebook or Reddit) are commonly used to study suicidality, video-sharing platforms like YouTube remain underexplored. As one of the most widely used multimedia-sharing platforms, YouTube enables users to communicate visually and verbally, providing a more direct and unmediated expression. Moreover, YouTube plays a significant role in both shaping and reflecting societal trends related to mental health and suicidality \citep{allgaier2020science, khan2022researching}. 
Still, a search on PubMed revealed only 28 articles containing the words "suicide" and "YouTube" in their title or abstract, with most focusing on the impact on viewers rather than on the creators \citep{dagar2020high, khasawneh2020examining}.

\begin{figure}[ht]
    \centering
    \includegraphics[width=0.6\textwidth]{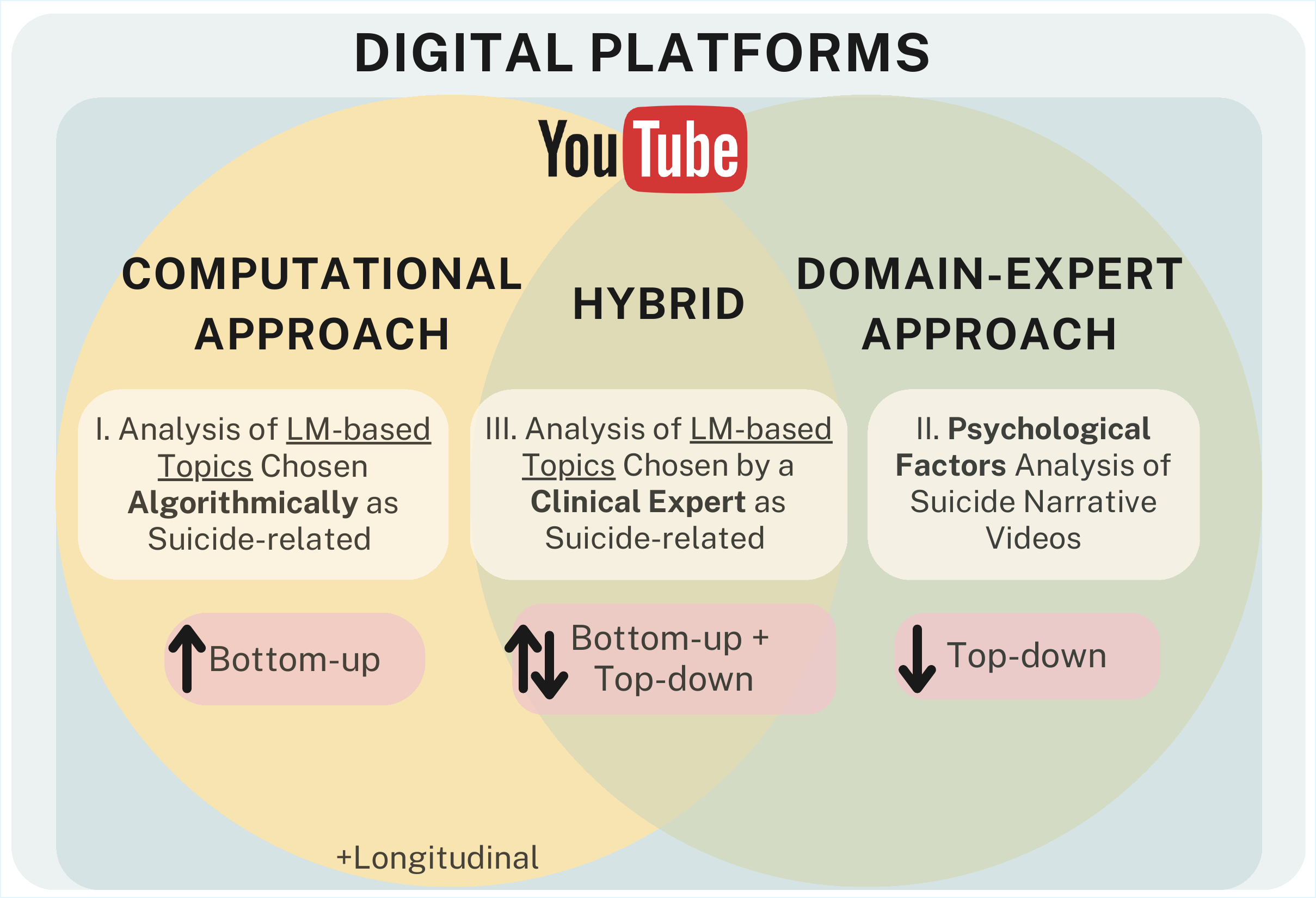} 
    \caption{An illustration of our research framework, showing the methodological approaches used in this study. Each circle represents a different approach, with the specific methods listed inside. The \textbf{computational approach} includes  Language Models (LM) based } topic modeling to identify behavioral patterns on YouTube (\romannumeral 1. \S\ref{results_buttom_up_comp}), analyzed through a psychological lens in the discussion section (\S\ref{sec:discussion}). The \textbf{domain-expert approach} involves a clinical assessment of suicide attempt narratives of predefined psychological factors   (\romannumeral 2. \S\ref{results_top_down_domain_expert}). In the \textbf{hybrid approach}, a domain expert reviewed all the LLM-derived topics and flagged those that may be considered as suicide risk factors, based on current clinical knowledge (\romannumeral 3. \S\ref{results_top_bottom_mixed}).
    \label{fig:intro}
\end{figure}

This study addresses existing research gaps by bridging the understanding of suicidality in offline and online contexts, considering both behavioral indicators and self-reported experiences, and by incorporating longitudinal analysis of the development of behaviors and feelings before and after suicide attempts. By applying three complementary approaches, as described in Figure \ref{fig:intro} and leveraging the strengths of both computational and domain-expert frameworks, we aim to explore our primary research question: \textit{How do linguistic patterns on YouTube reflect manifestations of suicidal behavior, and how do these patterns align with or differ from expert knowledge?}

\begin{figure}[ht]
    \centering
    \includegraphics[width=0.8\textwidth]{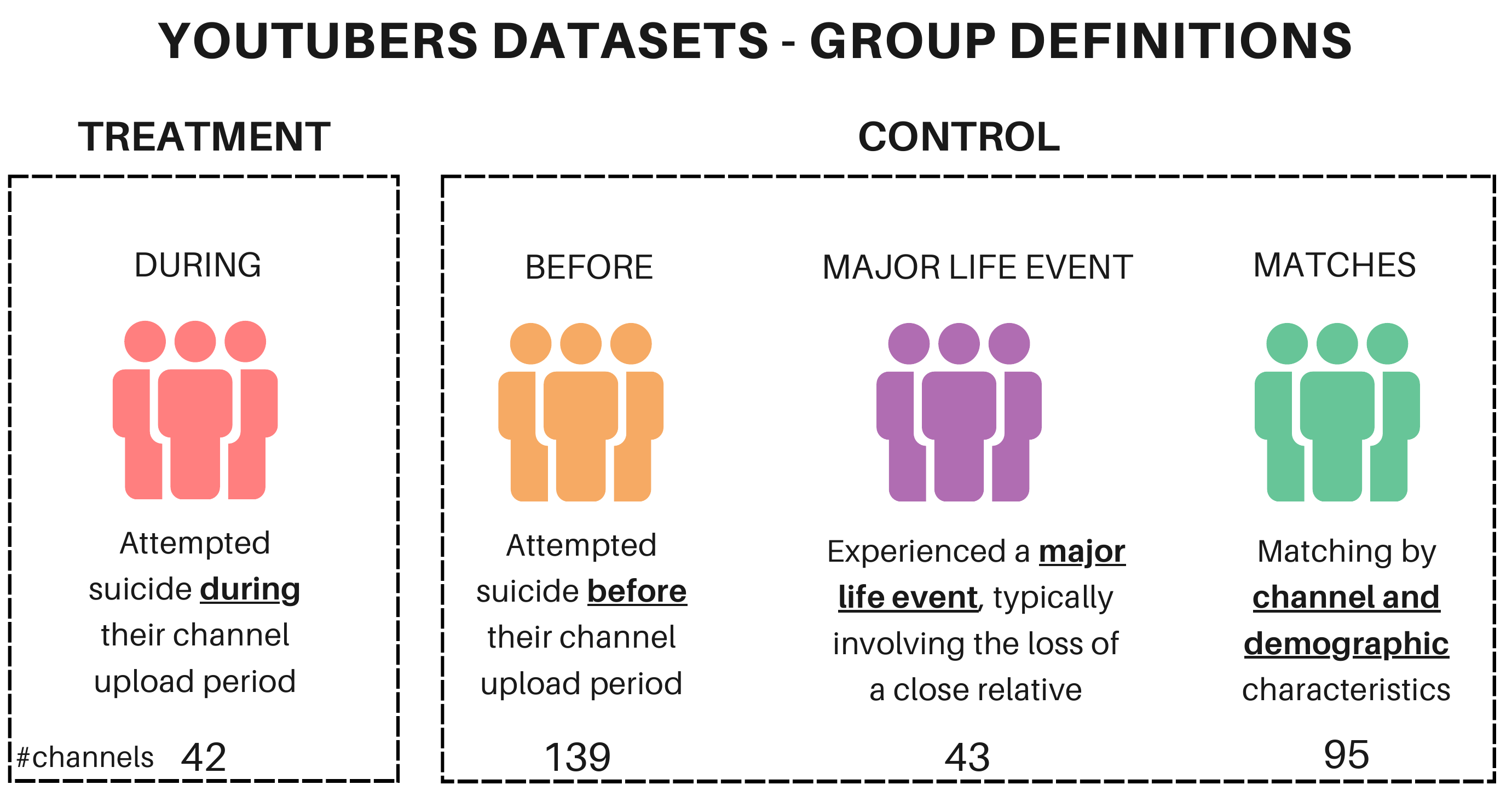}

   \caption{\small The research groups are categorized into treatment -\textbf{ Attempted (During)} - and three control groups - \textbf{Attempted (Before)}, \textbf{Control (Matches)} and \textbf{Control (Major Life Event)}. Both Attempted groups refer to those who attempted suicide during or before their channel upload period, while the other groups refer to control-based matches from the general population, or to those who experienced a major life event during their upload period (and didn't attempt suicide as far as we know). The sample size for each group is indicated below their respective categories. See data collection pipeline in Figure \ref{fig:high_level_method_pipeline}.}
    \label{fig:research_youtubers_groups_definition}
\end{figure}

To allow this investigation, we first constructed a \textbf{novel dataset} of YouTube channels belonging to individuals who had attempted suicide (Figure \ref{fig:research_youtubers_groups_definition}), with cases verified as life-threatening by a clinical psychologist (Figure A.1). We focused on individuals who attempted suicide after establishing their YouTube channel, referred as \textit{Attempted (During)}. Alongside the \textit{Attempted (During)} group, which we defined as the \textit{treatment} group, we gathered three \textit{control} groups representing varying levels of suicide risk.

In the \textbf{bottom-up approach}, we employed LM-based topic modeling and analyzed the topic distributions of each video to identify behavioral patterns on YouTube.
 By examining differences in pre-suicide attempt average topic probability in the treatment group compared to the control groups, we identified potential indicators of suicidal behavior. Then we performed a longitudinal analysis on the identified topics, revealing temporal trends before and after suicide attempts. 

Through this analysis, we answer the first part of the research question - \textit{how suicidal behaviors manifest on YouTube}, and partially address the second part, \textit{how are they similar or different from the clinical
literature}, by examining our findings in light of previous suicide research (see Discussion \S\ref{sec:discussion}).

In the \textbf{top-down approach}, we conducted a clinical psychological assessment of suicide attempt narrative videos to extract relevant predefined factors, including risk and resilience factors. We then statistically compared these factors between YouTubers with a general risk (i.e., attempted suicide in the past) and those at current risk (i.e., attempted during their upload period). This analysis enabled a traditional examination of self-reported suicide narratives.
 
Lastly, in the \textbf{hybrid approach}, all LLM-derived topics were reviewed by a clinical psychologist to manually identify suicide-related topics, and then subjected to a similar longitudinal analysis as the computational approach. Consequently, this provided a clinical perspective on how the current state of our knowledge about suicide, learned primarily from research in the real, offline world, may appear on YouTube. This approach also expanded the top-down method beyond suicide narrative videos to encompass the full content of YouTube channels.

This methodological framework enables the identification of emerging online behavioral trends and the examination of established psychological factors in a new online context, both of which offer novel insights into suicidality on YouTube (see results summary in Table \ref{tab:suicide_factors_comparison}).

Our \textbf{bottom-up}, fully automated computational analysis identified five topics as potential pre-attempt indicators, two of which also exhibited significant temporal shifts around the attempt (\S\ref{results_buttom_up_comp}): \textit{"Mental Health Struggles"} and \textit{"YouTube Engagement"} (see topics' representations: Figure \ref{fig:means_temporal_topics}). 
\textit{"Mental Health Struggles"} includes discussions about suicide, self-harm, and depression, aligned with established risk factors. \textit{“YouTube Engagement”}, which captures typical platform behavior such as audience interaction and content promotion, may serve as an indicator of “normal” YouTube activity, reflecting routine channel maintenance  \citep{zhang2023exploring, chung2023really, aquino2024youtube}. Notably, this topic appeared at lower levels in the treatment group prior to the suicide attempt—possibly signaling abnormal behavior and, therefore, a potential risk indicator. Both topics exhibited significantly higher prevalence following the suicide attempt, which we discuss at \S\ref{sec:discussion}. 

Additional three topics emerged as pre-attempt indicators and were not associated with a temporal pattern:  (1) "\textit{LGBTQ+ Identity and Acceptance}", (2) "\textit{Rape and Sexual Assault}", and (3) "\textit{Fashion and Style}", detailed in \ref{statisitcal_analysis_full} and discussed in \S\ref{sec:discussion}. The first two align with established suicide literature, reflecting the heightened distress faced by LGBTQ+ individuals due to discrimination and rejection \citep{sutter2016discrimination}, and by survivors of sexual violence due to trauma-related mental health challenges \citep{dworkin2022associations}. Both topics were more prevalent in the attempted groups compared to the other controls. In contrast, \textit{Fashion and Style}, a less familiar factor in suicide research, showed significantly lower prevalence in the attempted groups, potentially reflecting diminished engagement with self-expression or societal norms.

Moreover, the \textbf{top-down domain-expert approach} showed strong alignment in risk and resilience factors between YouTubers who attempted suicide before and during their upload period, reducing suicide-related tendencies confounders in the comparison between these groups (e.g., depression, family support). The only significant difference was the reason for sharing their attempt story: those who attempted suicide \textit{before} emphasized their \textit{Desire to Help Others}, while those who attempted \textit{during} framed it as part of their \textit{Personal Recovery Process} (Table \ref{psychological_stats_table}). 

 Finally, in the \textbf{hybrid approach}, although 19 topics were flagged as suicide-related by a clinical expert, three were already detected in the bottom-up approach, and only one, “Anxiety and Panic Attacks”, showed a significant group effect, but it did not exhibit a significant temporal change.

The discovery of significant indicators through the computational analysis, which were not identified in either the top-down or the hybrid analysis, underscores the value of bottom-up methods in uncovering hidden indicators.
The absence of some expected topics, along with the influence of platform-specific features, such as the YouTube-specific indicators observed in our study, highlights the need for online-focused research.
Furthermore, our findings emphasize the importance of distinguishing between current and past suicide attempts, as different patterns may emerge across these groups, reflected in both the computational and expert-driven analyses.

The key contributions of this study are: (1) novel insights into the manifestation of suicidal behavior in online contexts, including reduced engagement-seeking content as a strong behavioral marker; (2) integration of clinical and computational perspectives, revealing both aligned and distinct indicators; (3) uncovering psychological similarity between individuals who attempted suicide before and during their YouTube activity, differing mainly in their motivation for disclosure; (4) creation of a first-of-its-kind publicly available longitudinal dataset of 181 suicide-attempt channels and 134 controls, enriched with transcripts, metadata, demographics, and psychological assessments; (5) development of an AI-driven pipeline for analyzing longitudinal YouTube content; and (6) a novel matching framework for assembling high-quality YouTube control cohorts. We hope this work will inspire future efforts to integrate digital behavioral analysis with clinical insight in the study of suicidality.

\section{Methods}
\label{sec:methods}

This section outlines the methods used in this study, beginning with \textbf{data collection} and followed by a description of the \textbf{procedures} (Figure \ref{fig:method_procedures}). 
The \textbf{data collection} (\S{\ref{data_collection_section}}) involved constructing a textual dataset comprising four types of YouTube channels creators (Figure \ref{fig:research_youtubers_groups_definition}): belonging to creators who attempted suicide (before or during their YouTube activity), those who experienced a major life event, and matched control users from the general population. We searched YouTube for relevant channels, transcribed the videos and collected channel information, annotated demographic and psychological assessments, and matched them to control channels using a multi-step approach. The \textbf{procedures} (\S{\ref{Procedures}}) included applying an LM-based topic modeling to the channels' transcripts and implementing statistical procedures to address our research objective: understanding how suicidal behavior manifests on YouTube. 

\subsection{Data}\label{data_section}

\subsubsection{Data collection} \label{data_collection_section}

YouTube served as our primary data source. We identified users who uploaded a video sharing either their own suicide attempt story or a major life event using the YouTube API\footnote{API Documentation: \href{https://developers.google.com/youtube/v3/docs}{developers.google.com/youtube/v3/docs}}. These stories were typically not the main theme of the channels, but were often shared to raise awareness of \textit{suicide} or \textit{loss}. Additionally, we included a control group that matched the treatment users to facilitate a fair comparison with the general population. The data collection procedure is outlined in Figure \ref{fig:high_level_method_pipeline} and described in more detail below, with full documentation provided in Appendix (\ref{inclusion_exclusion_appendix} Search, \ref{data_preprocessing_appendix} Pre-processing, \ref{human_eval_protocol_full} Human Evaluation, \ref{matching_full} Matching). 

\phantomsection \label{data_collection_search}
First, using the YouTube API, we \textbf{searched} channels of people who attempted suicide (e.g., 'my suicide attempt') and people who experienced a major life event, mostly involving the loss of a close relative. These disclosure videos are named \textit{narrative} videos. 

\phantomsection \label{data_collection_human_eval}

We then conducted a \textbf{human evaluation} involving two levels of expertise. First, non-expert \textbf{channel validation} to ensure that only \textit{personal} channels were included, while excluding educational, news, and family channels. Following this, we extracted \textbf{user and event information} from the \textit{attempt} and \textit{major life event} stories (e.g., attempt date). Second, a \textbf{psychological assessment} of the \textit{attempt narrative videos} of YouTubers who reported a suicide attempt for two purposes: (1) \textbf{validating the intent of the attempt}, and (2) comparing Attempted (During) and Attempted (Before) to suicide-related tendencies to estimate potential confounding factors (e.g., mental illness, risk groups) and to perform a \textbf{suicide-related factor analysis}. See Tables \ref{tab:agreement}, \ref{tab:risk_factors_drilldown} for inter-annotator agreement which shows overall substantial to excellent reliability over a subset of examples. 

\phantomsection \label{data_collection_matching}

Next, we applied a multi-step \textbf{matching} process (Figure \ref{fig:matching_algorithm} visualizes the matching algorithm), which allowed us to first search for control cases from the general YouTube population and later match them to users who attempted suicide, based on demographic and channel characteristics. 

\phantomsection \label{data_collection_processing}

Finally, we employed \textbf{data processing}, which included collecting the users' channel videos and metadata (e.g., likes, description) and performing automatic transcription using Whisper \citep{radford2023robust}, the current speech-to-text state-of-the-art model.

\subsubsection{Study population }\label{dataset_demog_stats_section}

We categorized channels into two primary groups: treatment and control (Figure \ref{fig:research_youtubers_groups_definition}). The \textit{treatment} consisted of YouTubers who attempted suicide during their upload period, while the first \textit{control} group included YouTubers who attempted suicide before their upload period. These groups were labeled as \textbf{Attempted (During)} and \textbf{Attempted (Before)}. The additional \textit{control} groups comprised YouTubers who experienced a major life event during their upload period, labeled \textbf{Control (Major Life Event)}, and YouTubers matched to the treatment group based on demographic and content characteristics, labeled \textbf{Control (Matches)}. 
The inclusion of three control groups was designed to enable both temporal and cross-sectional comparisons. The Attempted (Before) group allows isolation of trait-level vulnerability effects of suicide, the Major Life Event group controls for non-suicidal distress, and the Matched control group captures baseline platform behavior. Together, these groups facilitate the disentanglement of suicide-specific, temporal, and contextual patterns.
See Table \ref{tab:inclusion_exclusion_criteria} for inclusion and exclusion criteria.

\textbf{The Final Sample} (Table \ref{tab:validity}): Of the initial 303 YouTubers in the \textit{Attempted} group, 104 users were excluded due to non-suicidal cases (e.g., clickbait-like videos like "Bust Myths About Suicide"), non-English content, or an insufficient number of videos, and 18 users were reclassified as Control (Major Life Event) due to the suicide attempt of a close relative. The final set of attempted channels consists of 181 channels (66\% female, average age = 25) who uploaded 34,407 videos ($M=190$, $Mdn=83$, $SD=286$). This dataset was later divided into two groups: 42 \textbf{Attempted (During)}, defined as the \textit{treatment}, and 139 \textbf{Attempted (Before)}, defined as one of the three \textit{controls}, based on whether the attempt occurred \textit{during} or \textit{before} their upload period  (Figure \ref{fig:research_youtubers_groups_definition}). 

A similar filtering process was applied to the \textbf{Control (Major Life Event)} group. The final Control (Major Life Event) dataset included 43 users (69\% female, average age = 27). These users collectively uploaded 13,470 videos ($M=320$, $Mdn=155$, $SD=372$). This subgroup included 25 users who experienced prolonged difficulties (e.g., relative's cancer), with the remaining 18 users experiencing the acute death of a family relative (e.g., heart attack).

The final \textbf{Control (Matches)} group included 95 channels (81\% female, average age = 23) that met the inclusion criteria and were found as most similar to the 42 Attempted (During) group according to the matching algorithm (Figure \ref{fig:matching_algorithm}). These 95 users collectively uploaded 14,860 videos ($M=182$, $Mdn=95$, $SD=223$).

\subsection{Procedures}\label{Procedures}

\begin{figure}[ht]
    \centering
    \includegraphics[width=\textwidth]{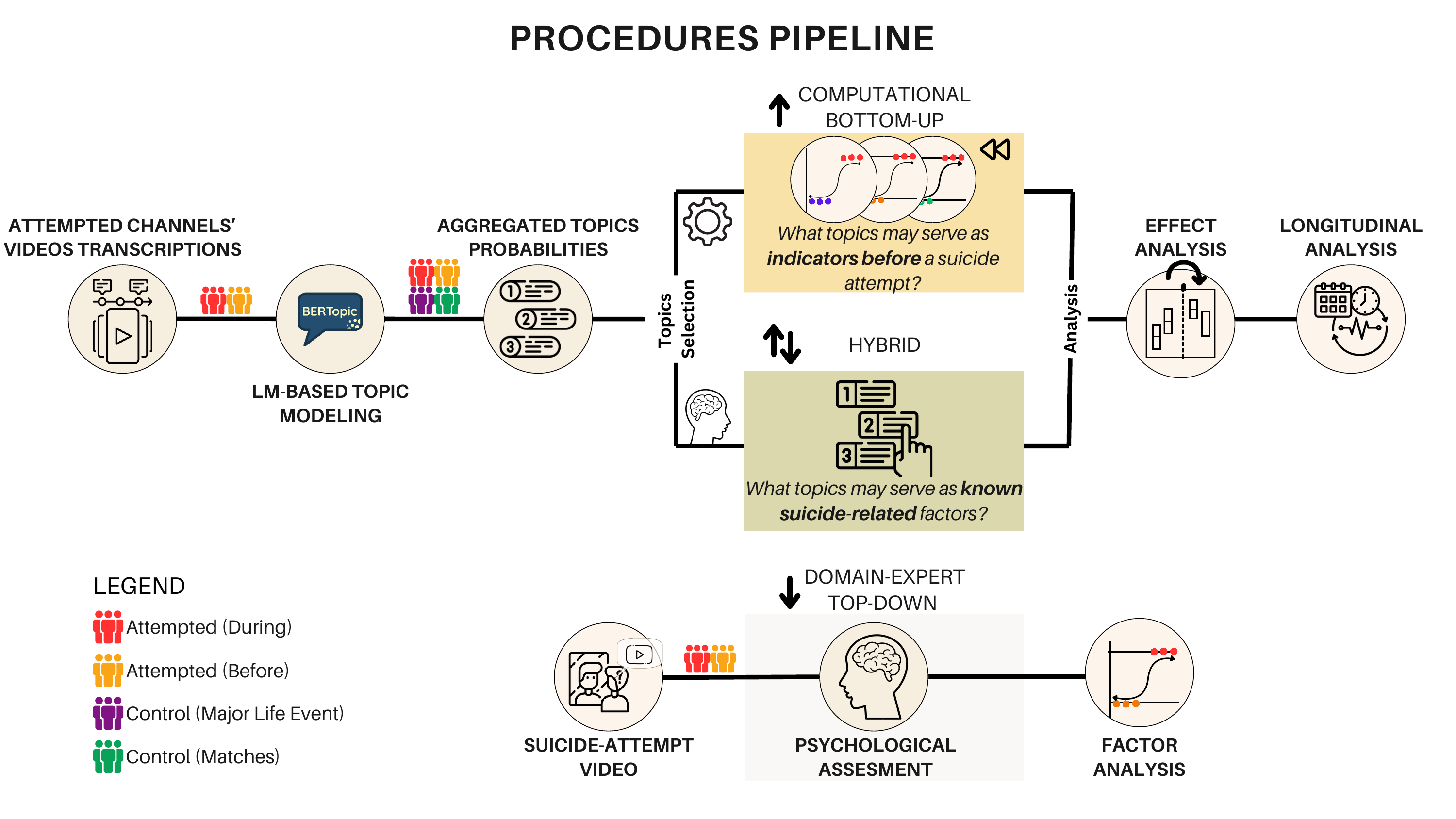} 
    \caption{The methodological pipeline consists of three approaches:
\textbf{1. Computational Bottom-Up -} Factors were \textit{generated} using an LM-based topic modeling algorithm, and topic relevance was evaluated using a bottom-up approach;
\textbf{2. Domain-Expert Top-Down -} Factors were \textit{predefined} by a clinical expert and assessed solely based on the suicide attempt narratives, and a
\textbf{3. Hybrid Approach -} A set of 166 LM-based topics was reduced to 19 based on their relevance to suicide as determined by a clinical expert.
The \textbf{bottom-up} and \textbf{hybrid} approaches were also followed by a two-way mixed effect ANOVA examining both temporal and group effects, while the \textbf{top-down} approach was limited to group differences.}
    \label{fig:method_procedures}
\end{figure}

This study introduces a novel \textit{longitudinal} analysis of linguistic content from YouTube channels, focusing on suicide attempts as a reference point for temporal changes. The procedures pipeline is visualized in Figure \ref{fig:method_procedures}, and outlined below.

\phantomsection \label{Procedures_topic_modeling}
\begin{tcolorbox}[title=Topic Definition, colback=gray!5!white, colframe=gray!50!black, fonttitle=\bfseries]
A \textbf{topic} is a cluster of semantically related text segments gathered by an unsupervised clustering method of sentence embeddings. Each topic is automatically named by an LLM considering top-associated words and representative texts.  
\end{tcolorbox} \label{definition_topic}

 First, we applied an \textbf{LM-based topic modeling} technique using \textbf{BERTopic} \citep{grootendorst2022bertopic}, which combines transformer-based sentence embeddings with density-based clustering. The pipeline uses Sentence-BERT \citep{reimers2019sentence} to encode text segments into semantic representations, followed by HDBSCAN clustering to identify thematic groups. Each cluster is then
 automatically labeled by Llama-2-7B \citep{touvron2023llama} based on representative
segments and keywords. We selected this approach because it enables stable, interpretable, and context-aware topic discovery across very large and unstructured datasets with no any earlier assumptions on the topics. It outperforms traditional topic models (e.g., LDA; \citealp{blei2003latent}) and other LLM-based topic modeling because it captures contextual semantic relationships while maintaining computational efficiency and topic stability at scale. Unlike traditional word co-occurrence methods, it detects semantically nuanced themes; unlike other LLM-based approaches, it systematically covers the full corpus and remains more computationally efficient and reliable for comprehensive topic extraction at scale
without over-generalizing dominant themes \citep{yang2025enhancing,mu2024large, zhang2022neural}. This method aligns with our goal of identifying non-explicit manifestations of suicidality in over 750,000 YouTube transcripts, where interpretability, stability,  and comprehensive coverage are essential (Figure \ref{fig:topic_modeling_pipeline}  provides an overview).

To capture the linguistic patterns typical of the at-risk population, the topic model was fitted exclusively on the textual content from the Attempted users’ channels, comprising 757,849 text segments, yielding 166 topics (see Definition~\ref{definition_topic}). This model was then used to infer topic distributions across the videos of all research groups (Figure \ref{fig:research_youtubers_groups_definition}). These topic distributions are used in bottom-up and hybrid approaches. Full technical details are provided in Appendix B.1 (prompts, models' hyper-parameters, and quality metrics over topics).

Next, we defined the \textit{reference event} of each group: a suicide attempt for Attempted (During), a major life event for Control (Major Life Event), and a synthetic event (midpoint of uploads) for Attempted (Before) and Control (Matches). Topic probabilities were averaged separately for the pre- and post-event periods. Although the attempt narratives were included in the topic model, they were excluded from subsequent statistical analyses to avoid bias, as they do not represent the typical content of the channels.

\phantomsection \label{Procedures_statistical_analysis}
The \textit{bottom-up} and \textit{hybrid} procedures followed a two-step protocol: \textbf{topic selection}, aimed at enabling more reliable statistical analysis in a high-dimensional setting, and a \textbf{statistical effect analysis} to evaluate group and time effects.

In the \textit{hybrid} approach, we asked a clinical expert:  \textit{What topics are known risk factors of suicide?} which reduced the topics from 166 to 19. In the \textit{bottom-up} approach, we asked: \textit{What topics may serve as indicators of the period before a suicide attempt?}, and used stepwise logistic regression to the pre-event period \citep{pace2009stepwise}, coding Attempted (During) as 1 and controls as 0, across three control populations and their combined pool, yielding four binary models and narrowing topics from 166 to five. \footnote{\label{fn:lasso} All stepwise regression findings were replicated using LASSO-regularized logistic regression, confirming the robustness of the identified predictors.}

Following topic selection, we fit generalized linear mixed models (GLMMs) \citep{bagiella2000mixed} with a Beta likelihood and a logit link function \citep{ferrari2004beta} to model topic probabilities. In total, we ran $24$ models, one per topic (hybrid: 19 topics, bottom-up: five). For each topic, we calculated the mean probability for the pre-event and post-event periods. The models included fixed effects for \textit{Time} (After vs.~Before the event), \textit{Group} (see Figure~\ref{fig:research_youtubers_groups_definition}), their interaction (\textit{Group} × \textit{Time}), and general control variables (Gender, Age, Ethnicity, and Upload activity). The \textit{Attempted (During)} group and
the pre-event period (\textit{Before}) served as reference categories.  We applied Benjamini-Hochberg (BH) FDR correction \citep{benjamini1995controlling} across topics. The complete model specification, including the mathematical formulation and interpretation of coefficients, is provided in Appendix \ref{appendix_effect_analysis}.

As a complementary analysis, we extended our time-aggregated approach by conducting two types of \textbf{longitudinal topic analyses} to explore temporal patterns of significant topics. The first was a \textit{within-group analysis}, where we explored changes over time within the Attempted (During) group. This analysis used $t=0$, allowing us to provide a standardized baseline from which to compare temporal changes. The second was a \textit{between-group analysis}, in which we compared the Attempted (During) group to the three control groups at each timestep. Appendix \ref{fig:temporal_analysis_pipeline} and \ref{appendix_temporal_testing} provide further in-depth description of the temporal alignment and testing procedure.

To ensure that our findings were not driven by external or methodological artifacts, we conducted a series of \textbf{robustness and validation analyses}. These analyses evaluated potential confounding factors at multiple levels: (1) a \textit{sensitivity analysis} excluding ambiguous cases of suicidal ideation to assess the stability of group effects; (2) a \textit{control measures analysis} examining content shifts during the COVID-19 pandemic; (3) an \textit{engagement metrics analysis} comparing aggregate audience interaction indicators (e.g., likes) across groups; and (4) an \textit{activity analysis} assessing individual upload behavior and frequency before and after the reference event. These analysis together with the general control variables included in the GLMM model reduces the possibility of confounding effects in our results (full details in Appendix \ref{validation_analysis}). 

\phantomsection
\label{Procedures_psychological_factors}

Finally, motivated by traditional clinical assessment practices, we conducted a \textit{top-down} analysis in the form of a \textbf{psychological factor analysis}, aiming to assess whether the behavioral signals identified in our study align with established clinical knowledge on suicidality. This analysis is based upon assessments conducted by a clinical psychologist on the Attempted groups (\S{\ref{data_collection_human_eval}}) and involved one-hot encoding of the psychological factors, then performing a stepwise logistic regression analysis, where the outcome variable was coded as 1 for Attempted (During) and 0 for Attempted (Before). \textsuperscript{\ref{fn:lasso}}

\newpage
\section{Results}
\label{sec:results}

Based on the research framework illustrated in Figure \ref{fig:intro}, this chapter presents the results of the three approaches (Table \ref{tab:suicide_factors_comparison}): bottom-up computational (\S\ref{results_buttom_up_comp}), top-down domain-expert (\S\ref{results_top_down_domain_expert}), and hybrid (\S\ref{results_top_bottom_mixed}), complemented by longitudinal topic analysis (\S\ref{results_longitudinal}). 

\begin{table}[h]
\caption{Comparison of analytical approaches and suicide-related factors. Each factor was identified through one or more of the following approaches: \textbf{Bottom-Up $\uparrow$} (LM-based topic modeling), \textbf{Hybrid $\uparrow \downarrow$} (expert evaluation of LM-based topics), or \textbf{Top-Down $\downarrow$} (psychological annotation of narrative videos). The leftmost column indicates whether the factor was generated by an LLM or defined by a human expert. The rightmost columns indicate statistically significant effects from the mixed ANOVA: \textbf{Group Effect} refers to the main effect of group; \textbf{Temporal Effect} refers to the group × time interaction. "–" marks untested effects. \textit{Attempted (During)} group.}
\label{tab:suicide_factors_comparison}
\centering
\scriptsize
\begin{tabular*}{0.98\linewidth}{@{\extracolsep{\fill}} lllllll@{}}
\toprule
Origin & Factor & Bottom-Up $\uparrow$ & Hybrid $\uparrow\downarrow$ & Top-Down $\downarrow$ & Group & Temporal \\
\midrule
\multirow{6}{*}{LM-Based Topics} 
& YouTube Engagement & \cmark & \xmark & – & \cmark & \cmark \\
& Mental Health Struggles & \cmark & \cmark & – & \cmark & \cmark \\
& Fashion and Style & \cmark & \xmark & – & \cmark & \xmark \\
& LGBTQ+ and Personal Identity & \cmark & \cmark & – & \xmark & \xmark \\
& Rape and Sexual Assault & \cmark & \cmark & – & \xmark & \xmark \\
\midrule
\multirow{2}{*}{Domain-Expert} 
& Personal Recovery Process & – & – & \cmark & \cmark & – \\
& Desire to Help Others & – & – & \cmark & \cmark & – \\
\bottomrule
\end{tabular*}
\end{table}

\subsection{The bottom-up analysis}

\label{results_buttom_up_comp}

\begin{figure}[ht]
    \centering
    \includegraphics[width=\textwidth]{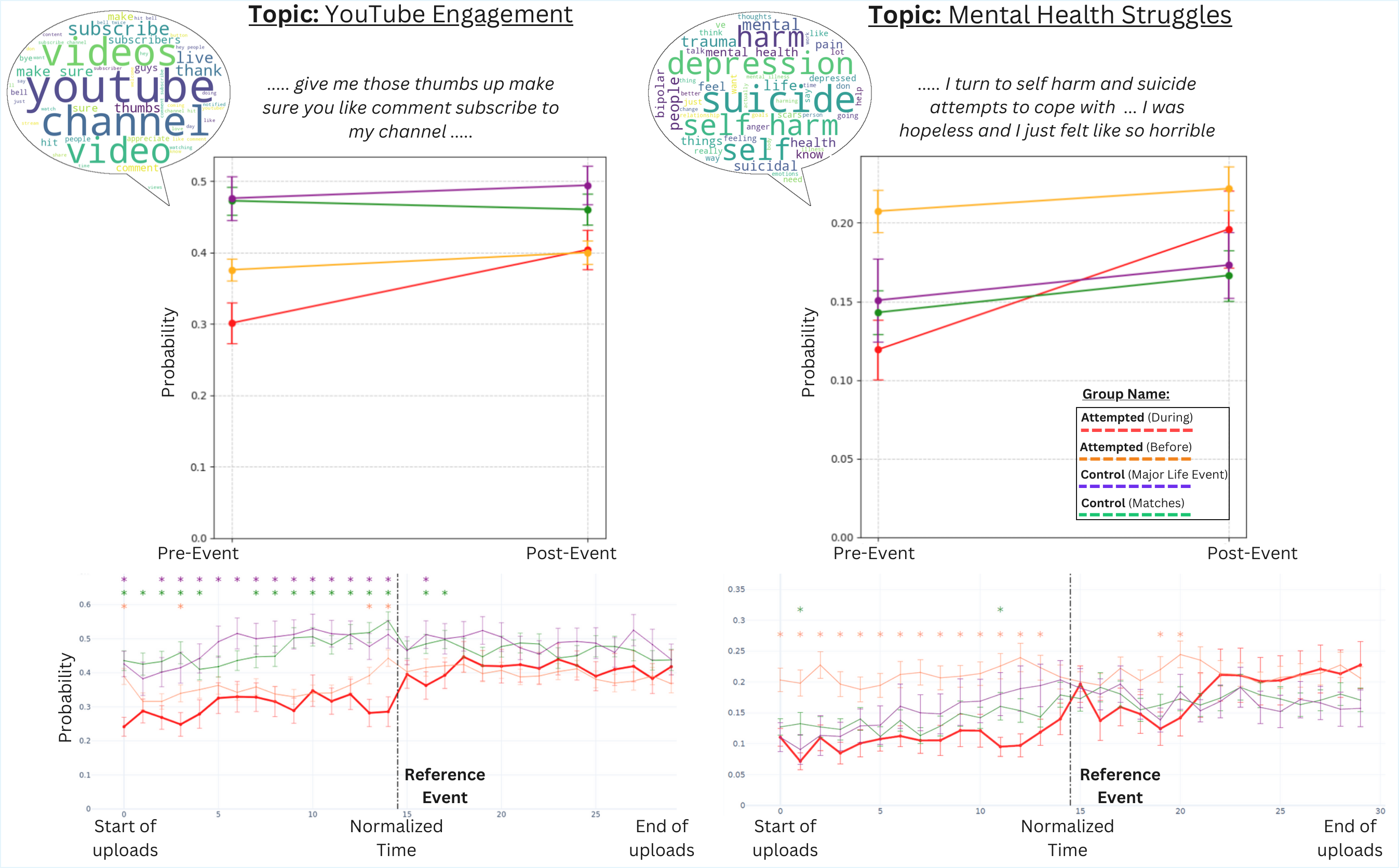}

\caption{\small \textbf{YouTube Engagement and Mental Health Struggles:} The figures present topics identified as significant in the bottom-up analysis (\S\ref{results_buttom_up_comp}): \textit{YouTube Engagement} (left) and \textit{Mental Health Struggles} (right). The top row shows average topic values across four groups—Attempted (During), Attempted (Before), Control (Major Life Event), and Control (Personal)—in Pre- and Post-Event periods. TF-IDF word clouds are displayed alongside representative transcript excerpts, with word size reflecting importance. Below, the second row presents longitudinal trends over normalized time, aligned to each group's reference event (e.g., suicide attempt). Asterisks ($*$) denote statistically significant differences between Attempted (During) and other groups at each time point (\textit{t}-test; see Appendix \ref{appendix_temporal_testing} for temporal testing procedures).} \label{fig:means_temporal_topics}
\end{figure}

Our primary computational analysis examines LM-based topic distributions (Figure \ref{topic_modeling_full}).
Statistical comparison between the \textit{treatment} Attempted (During) group and the \textit{control} groups found two topics - \textit{YouTube Engagement} and  \textit{Mental Health Struggles}, as both predictive of the pre-attempt period and significant indicators of temporal changes (pre to post-attempt). Representations of these topics are shown in Figure \ref{fig:means_temporal_topics}.

\subsubsection{What topics may serve as indicators of the period before a suicide attempt?}\label{results_buttom_up_comp_reducing_topics}

As part of the topic selection procedure, we identified topics predictive of the period \textit{preceding} a suicide attempt with a forward-backward stepwise logistic regression\textsuperscript{\ref{fn:lasso}}, as these factors \textit{may} act as indicators of current risk for suicide attempt.

The analysis revealed five significant topics (Table \ref{regression_results_table}):  (1) \textit{YouTube Engagement}, (2) \textit{Mental Health Struggles}, (3) \textit{Fashion and Style},
(4) \textit{Rape and Sexual Assault} and (5) \textit{LGBTQ+ Identity and Acceptance}. Figures \ref{full_tf_idf_wordclouds} and \ref{fashion_and_lgbtq_means} illustrate their word cloud and temporal dynamics. 

\begin{table}[ht]
\caption{Regression results predicting membership in the Attempted (During) group versus individual or pooled control groups (Four separate models were trained). Models use pre-event average topic probabilities. $R^2$ refers to McFadden's pseudo-R-squared}
\label{regression_results_table}
\scriptsize
\footnotesize
\begin{tabular*}{0.98\linewidth}{@{\extracolsep{\fill}} llcccccr@{}}
\toprule
Treatment Group & Control Group & Topic & $\beta$ Coeff. & $z$ Score & $p$ Value & 95\% CI & $R^2$ \\
\midrule
\multirow{7}{*}{Attempted (During)}
  & \multirow{2}{*}{Attempted (Before)}
  & Mental Health Struggles & -1.49 & -3.62 & $<0.001$ & [-2.29, -0.68] & \multirow{2}{*}{0.11} \\
  &                         & Rape and Sexual Assault & 0.68 & 2.68 & 0.007 & [0.185, 1.186] & \\
  \cline{2-8}
  & \multirow{2}{*}{Control (Major Life Event)}
  & YouTube Engagement & -1.46 & 4.23 & $<0.001$ & [-2.142, -0.787] & \multirow{2}{*}{0.23} \\
  &                             & LGBTQ+ Identity and Acceptance & 0.96 & 2.58 & 0.010 & [0.233, 1.7] & \\
  \cline{2-8}
  & \multirow{2}{*}{Control (Matches)}
  & YouTube Engagement & -0.808 & -3.19 & 0.001 & [-1.304, -0.313] & \multirow{2}{*}{0.22} \\
  &                             & Fashion and Style & -0.85 & -2.74 & 0.006 & [-1.462, -0.243] & \\
  \cline{2-8}
  & Combined Pool & YouTube Engagement & -0.73 & -3.92 & $<0.001$ & [-1.098, -0.366] & 0.05\\
\bottomrule
\end{tabular*}
\end{table}

\subsubsection{Temporal effect analysis of pre-attempt indicator topics}
\label{results_buttom_up_comp_effect_analysis}
\begin{table*}[ht]
\centering
\scriptsize
\caption{Fixed effects from GLMMs for topic‐level outcomes. 
Positive coefficients indicate higher topic prevalence relative to baseline. 
Reference group: Attempted (During); Reference time: Before event.}
\label{tab:glmm_results}
\begin{tabular}{lccccccc}
\toprule
\textbf{Topic} & \textbf{Parameter} & \textbf{$\beta$} & \textbf{OR} & \textbf{SE} & \textbf{z} & \textbf{p} & \textbf{$p_{adj}$}\\
\midrule
\multicolumn{7}{c}{\textbf{Temporal Effects (Time and Group × Time Interactions)}} \\
\midrule
\multirow{4}{*}{Mental Health Struggles} 
 & Time (After vs. Before) & 0.554 & 1.74 & 0.145 & 3.82 & $<.001$ &  $<.001$ \textcolor{red}{**} \\
 & Group × Time: Attempted (Before) & $-0.440$ & 0.64 & 0.162 & $-2.72$ & .006 & .016 \textcolor{red}{*}  \\
 & Group × Time: Control (Matches) & $-0.356$ & 0.70 & 0.172 & $-2.08$ & .038 & .095 \\
 & Group × Time: Control (Major Life Event) & $-0.397$ & 0.67 & 0.199 & $-1.99$ & .046 & .11\\
\midrule
\multirow{4}{*}{YouTube Engagement}
 & Time (After vs. Before) & 0.515 & 1.67 & 0.107 & 4.80 & $<.001$ & $<.001$ \textcolor{red}{**} \\
 & Group × Time: Attempted (Before) & $-0.353$ & 0.70 & 0.122 & $-2.90$ & .004 & .016 \textcolor{red}{*} \\
 & Group × Time: Control (Matches) & $-0.527$ & 0.59 & 0.128 & $-4.13$ & $<.001$ & $<.001$ \textcolor{red}{**}  \\
 & Group × Time: Control (Major Life Event) & $-0.415$ & 0.66 & 0.149 & $-2.78$ & .005 & .027 \textcolor{red}{**} \\
\midrule
\multicolumn{7}{c}{\textbf{Non‐Temporal Effects (Baseline and Covariates)}} \\
\midrule
\multirow{2}{*}{Fashion and Style}
 & Group: Control (Matches) & 0.891 & 2.44 & 0.311 & 2.86 & .004 & .01 \textcolor{red}{**} \\
 & Gender (Male) & $-0.661$ & 0.52 & 0.259 & $-2.55$ & .011 & .026 \textcolor{red}{*} \\
\midrule
\multirow{4}{*}{YouTube Engagement}
 & Group: Attempted (Before) & 0.336 & 1.40 & 0.138 & 2.43 & .015 & .075 \\
 & Group: Control (Matches) & 0.827 & 2.29 & 0.137 & 6.05 & $<.001$ & $<.001$ \textcolor{red}{**} \\
 & Group: Control (Major Life Event) & 0.764 & 2.15 & 0.204 & 3.75 & $<.001$ & .001 \textcolor{red}{**} \\
 & Gender (Male) & $-0.314$ & 0.73 & 0.110 & $-2.84$ & .005  & .023 \textcolor{red}{*}  \\
\bottomrule
\end{tabular}
\vspace{0.4em}
\footnotesize{Coefficients ($\beta$) are log‐odds from Beta GLMMs; 
odds ratios (OR) are computed as $\exp(\beta)$. $p$-values are reported for each model coefficient. To account for multiple comparisons across topics, $p_{adj}$ shows the FDR-corrected p-values (Benjamini-Hochberg) applied separately for each coefficient type across all topics.
Significance stars correspond to $p_{adj}$ thresholds: * $<0.05$, ** $<0.01$.
}
\end{table*}

Following the topic selection procedure (\S\ref{results_buttom_up_comp_reducing_topics}), which identified five topics, we fit a generalized linear mixed model (GLMM) to examine the effects of group and time, as well as their interaction, along with demographic and upload activity covariates.

As illustrated in Figure~\ref{fig:means_temporal_topics}, the Attempted (During) group exhibited a post-event increase in both topics, aligning with the Attempted (Before) group. For \textit{YouTube Engagement}, both Attempted groups (Before and During) showed lower levels before and after the event compared to the control groups.

As shown in Table \ref{tab:glmm_results} and illustrated in Figure~\ref{fig:means_temporal_topics}, significant group-by-time interactions were observed for the topics \textit{YouTube Engagement} and \textit{Mental Health Struggles}. This analysis found that the Attempted (During) group showed significantly different temporal changes in \textit{YouTube Engagement} compared to all control groups, and in \textit{Mental Health Struggles}, only compared to the Attempted (Before) group.  We provide a more detailed description of the GLMM model and results, in terms of odds change and non-temporal significant results in Appendix \ref{appendix_effect_analysis} (Effect Analysis).
We also conducted a sensitivity analysis excluding ambiguous cases (Appendix \ref{validation_analysis}), which confirmed that temporal effects remained statistically significant and consistent in direction.

\subsection{The longitudinal topic analysis}
\label{results_longitudinal}

We conducted a longitudinal analysis focusing on the topics \textit{Mental Health Struggles} and \textit{YouTube Engagement}, as they exhibited significant temporal changes following a suicide attempt (\S{\ref{results_buttom_up_comp_effect_analysis}}). 

Figure \ref{fig:temporal_topics_within} shows the results of the within-group analysis, which illustrates how the prevalence of \textit{YouTube Engagement} and \textit{Mental Health Struggles} significantly increased after the attempt. Additionally, the between-group analysis (Figure \ref{fig:means_temporal_topics}) reveals that \textit{YouTube Engagement} significantly differs between the Attempted (During) group and all control groups in the time-steps right before the attempt. Meanwhile, significant differences in \textit{Mental Health Struggles} were primarily observed between the Attempted (During) and Attempted (Before) groups. These differences were further validated as relative to suicide attempt, as a control analysis using COVID-19 as a reference event showed no significant differences between-group effects (Appendix \ref{validation_analysis}).

\subsection{The top-down domain-expert analysis}

\label{results_top_down_domain_expert}

While an AI-driven bottom-up approach uncovers topics without relying on clinical knowledge, traditional clinical assessments provide structured interpretations grounded in psychological literature. To combine these perspectives, we incorporated a domain-expert top-down analysis. In this analysis, a clinical psychologist assessed psychological factors by analyzing only the suicide attempt narrative videos, whereas the bottom-up approach was applied to the entire channel. This method allowed us to construct a psychological profile of the attempted groups, which we later incorporated into a regression analysis to differentiate between the two groups. 

We found that the Attempted (During) and Attempted (Before) groups reported similar risk and resilience factors in their suicide narratives (Table \ref{psychological_stats_table}). Correspondingly, the users in both the Attempted (During) and the Attempted (Before), display a similar distribution of mental health conditions (e.g., 78.6\% and 79.9\% self-reported depression), resilience factors (40.5\% and 37.4\% self-reported social/family support), and feelings of hopelessness/despair  (71.4\% and 64\%) and unbearable mental pain  (38.1\% and 46.8\%).  The only significant difference between the groups lies in their attempt sharing motivations: Attempted (During) users mainly discuss \textit{Personal recovery} (66.7\% vs. 38.1\%), while Attempted (Before) users mentioned the \textit{Desire to help others} (77\% vs. 40.3\%).  

Subsequently, we examined the significance of these differences by applying one-hot encoding to the psychological factors and conducting a stepwise logistic regression analysis on group membership. The regression identified the motivation \textit{Desire to help others}, as a significant factor ($\beta = -1.69$, $p < .01$, $z = -4.49$), followed by \textit{Personal recovery process} ($\beta = 1.08$, $p < .01$, $z = 2.93$) after the former was excluded. These were the only two significant factors; excluding both revealed no additional factors (see \S\ref{sec:discussion}).

Since the main bottom-up approach was not specific to the suicide attempt narratives, we conducted an additional analysis using a separate LM-based topic model focused solely on suicide attempt narratives (Appendix C.2). This complementary analysis enabled us to identify thematic patterns within the narratives, and initial exploration shows alignment between LM-based topics and predefined suicide-related psychological factors. 

\subsection{The hybrid bottom-up and top-down analysis} \label{results_top_bottom_mixed}

The \textit{bottom-up} approach automatically generated LM-based topics, which were analyzed computationally and longitudinally to identify behavioral indicators of suicide attempts (§\ref{results_buttom_up_comp}). In contrast, the \textit{top-down} approach involved manual assessment of pre-defined psychological factors and analysis of suicide attempt narrative videos (\S\ref{results_top_down_domain_expert}). To integrate both approaches, we asked a clinical psychologist to evaluate all 166 LM-based topics (\ref{definition_topic}) and flag those associated with suicidality. We then conducted the same statistical effect analysis used in the bottom-up approach on this expert-curated subset (\S{\ref{results_buttom_up_comp_effect_analysis}}).

\subsubsection{What topics are known risk factors for suicide?}
\label{results_top_bottom_mixed_reducing_topics}

The clinical psychologist identified 19 topics as related to suicidality based on topic descriptions, three of which were already identified by the computational approach (\S{\ref{results_buttom_up_comp}}). In this section, we focus on the remaining 16 topics. The full list is in Table \ref{tab:topic_info}.

\subsubsection{Temporal Effect Analysis of Suicide-Related Topics}
\label{results_top_bottom_mixed_effect}

Following the topic selection procedure (\S{\ref{results_top_bottom_mixed_reducing_topics}}), we fit a GLMM on the 19 expert-selected topics  (as in \S\ref{results_buttom_up_comp_effect_analysis}). The goal was to identify topics that not only differed between groups but also changed differently over time across those groups. The analysis didn't reveal new significant topics with reference to \textit{Attempted (During)}.  

A qualitative analysis did reveal an interesting pattern of one topic—\textit{Anxiety and Panic Attacks}.
As shown in Figure \ref{fig:anxiety_means_word_cloud}, the Attempted (During) and Control (Major Life Event) groups had lower topic prevalence before the event, with some increase observed post-event. However, the observed temporal increase in the Attempted (During) group was not statistically significant ($p_{adj} > .05$).

Interestingly, well-established clinical suicide-related factors, such as \textit{Religion}, \textit{Addiction}, and \textit{Bullying} (topic information can be seen in Table \ref{tab:topic_info}), did not yield statistically significant effects in our analysis, despite being flagged as relevant by a clinical expert. Additionally, other risk factors, such as \textit{Hopelessness} or \textit{Social Isolation}, did not appear in the LM-based topics.

\section{Discussion}\label{sec:discussion}

This study introduces a unique, publicly available longitudinal dataset of 181 YouTube channels from individuals with verified suicide attempts, along with two additional control groups (n = 138), analyzed through a multi-method framework combining bottom-up, top-down, and hybrid approaches. Anchoring the analysis around the suicide attempt, the bottom-up computational approach identified \textbf{Mental Health Struggles} and \textbf{YouTube Engagement} as key behavioral markers, both significantly lower before the attempt compared to control groups and increasing afterward. The top-down psychological assessment of suicide narratives confirmed the suicide attempts as life-threatening, and revealed strong alignment between the narratives of \textit{Attempted (Before)} and \textit{Attempted (During)} groups in psychological factors such as depression and loneliness. The only notable difference was motivation to share their suicide attempt, with the latter framing it as part of a \textbf{Personal Recovery Process}. Lastly, the hybrid approach identified 19 out of 166 LLM-derived topics as suicide-related. Together, these complementary perspectives provide a richer, more nuanced understanding of how suicidality manifests on YouTube.

A particularly intriguing finding was the increase in the discussion of \textbf{Mental Health Struggles} \textit{after} a suicide attempt. This finding is in line with previous studies indicating that mental health struggles may increase after a suicide attempt, since mental health disorders (depression, anxiety, PTSD) are often persistent or worsen post-attempts \citep{mcgill2019information}. Moreover, the period after an attempt is a high-risk window for another attempt \citep{mcgill2019information}. Individuals after an attempt may feel increased feelings of shame and guilt, and their interpersonal relationships, may be challenged \citep{sheehy2019examination}. 
However, this observation might appear counterintuitive based on some traditional psychological expectations, which suggest that discussions about mental health struggles typically increase prior to an attempt, as individuals at risk are more likely to seek help from mental health services \citep{luoma2002contact}, and may diminish post-attempt due to a sense of relief or emotional avoidance \citep{bryan2019examining, kaplow2014emotional}.

Assuming the latter theory holds, possible explanations for the discrepancy between this theory and our findings may relate to the nature of the data we rely on or may be explained by psychological theory.
First, consider the characteristics of the study population and platform. As noted by \citep{luoma2002contact}, “...alternate approaches to suicide-prevention efforts may be needed for those less likely to be seen in primary care or mental health specialty care, specifically young men.” Our research population differs from those typically studied, focusing on young people and YouTube creators in the context of their online behavior. Moreover, YouTube content may not fully capture the breadth of an individual’s emotional experiences, as video creation depends on both the technical and emotional readiness of the creator.
Second, the clinical literature suggests that the increased descussion on mental health could be related to the process of \textbf{post-traumatic growth} \citep{frey2019recipients}. Individuals who survive a suicide attempt may experience psychological growth through sharing their struggles \citep{kirchner2024experiences}. This hypothesis is further supported by our psychological analysis, which found that the Attempted (During) group was significantly more motivated to share their stories as part of a \textbf{Personal Recovery Process}.

The second main finding, regarding \textbf{YouTube Engagement} (content aimed at driving audience interaction), is an intriguing behavioral marker, as it may reflect normal platform usage patterns rather than being directly indicative of distress. Given that creators often seek interaction to maintain their audience \citep{zhang2023exploring, chung2023really, aquino2024youtube}, this finding underscores the importance of contextualizing social media behaviors within platform norms.
The finding regarding higher YouTube engagement after a suicide attempt may be explained in several ways. Engaging others may be a form of seeking connection. Individuals often feel isolated after a suicide attempt and turn to others in order to understand and validate their struggles and feelings. A social media community can provide empathy and reassurance. Another option is that the engagement is actually a cry for help \citep{maple2020nobody}. This may be a way to signal that they are still struggling and need support. They might not feel comfortable asking for help directly, but hope that someone will notice and intervene. Future studies may also consider the connection between online engagement and addictive social media use, as recently described in youth suicidal behaviors \citep{xiao2025addictive}.

Additional findings emerged regarding three topics identified as potential pre-attempt indicators in the computational approach: (1) \textbf{LGBTQ+ Identity and Acceptance}, (2) \textbf{Rape and Sexual Assault}, and (3) \textbf{Fashion and Style}. The first two align with existing literature on suicide risk factors and were independently identified as suicide-related by a clinical psychologist in the mixed-method approach. \textbf{LGBTQ+} individuals are at higher risk for suicide attempts compared to the general population \citep{russell2016mental, trevor}. The chronic stress experienced by LGBTQ+ individuals due to social rejection and discrimination is extremely high \citep{sutter2016discrimination}. In addition, LGBTQ+ individuals face many barriers to mental health care \citep{meyer2003prejudice}.  Similarly, \textbf{Rape and Sexual Assault} are profoundly traumatic experiences that elevate the risk for attempts \citep{dworkin2022associations}. These traumas often lead to mental health disorders, including depression and substance abuse, which are themselves risk factors. Moreover, the victims often suffer from social isolation and stigma, which intensifies feelings of loneliness and despair.
In contrast to the first two topics, \textbf{Fashion and Style}, a less familiar factor in suicide research, showed a significantly lower prevalence in the attempted groups. This may reflect reduced engagement with self-expression or social norms. Prior studies link between mood and clothing and suggest that fashion may indicate on positivity \citep{moody2010exploratory, masuch2014understanding}. This pattern may also align with Joiner's Interpersonal Theory, where social withdrawal and loss of belonging contribute to suicidal behavior \citep{joiner2005people}. The decrease in fashion-related content may indicate broader patterns of social disengagement and loss of interest in activities that typically foster interpersonal connection. However, given that Fashion and Style are not conventionally examined in suicide research, further studies are needed to clarify this relationship.

Despite these promising findings, this study has several limitations. First, the sample is inherently biased toward individuals who are active on YouTube and may not represent the broader population at risk for suicide. Second, the reliance on public content means that more private or intimate expressions of distress are not captured. Third, our analyses are restricted to speech-to-text transcripts rather than multi-modal cues such as tone, facial expression, or movement. Finally, although our temporal alignment procedure standardized all users to 15 pre- and 15 post-event intervals and the number of valid transcripts was included as a parameter in the GLMM, differences in the amount of linguistic data before and after the attempt may still affect variance estimation. Users with fewer valid transcripts in one period contribute less information, which may influence standard errors.

Looking ahead, future research could explore and compare how suicidal behaviors manifest across different social media platforms. A particularly compelling direction would be to examine engagement techniques across platforms and analyze their connection to mental health-related behaviors, as YouTube Engagement is identified in this research as a strong and consistent behavioral indicator.
Another promising direction is leveraging YouTube's multi-modal content, including speech, audio, and visual, to enhance predictive power and capture nuanced behaviors that may have been overlooked in this study.

In summary, our study adds to the growing body of literature on social media and mental health while also advancing the use of YouTube for suicide research. By integrating qualitative and quantitative methods and emphasizing longitudinal analysis considering suicide attempts, we highlight the significance of a temporal approach compared to broader group-based analyses. Although not intended for suicide-risk prediction or clinical assessment, this study underscores the potential of incorporating digital traces, such as social media language, into future ethically governed frameworks for suicide-risk assessment beyond traditional clinical settings to ensure responsible clinical use. We hope this work encourages further interdisciplinary research in this critical area.

\textbf{Ethical Considerations \& Data Management}
This study relied exclusively on publicly available YouTube content. No private or restricted data were accessed, and only aggregated analyses are reported, ensuring that no individual creator can be re-identified. The dataset will be released only after formal approval is obtained from the institution’s ethics and legal departments, and under a protocol that ensures full compliance with data protection and privacy regulations.

\bibliography{mybibliography}

@article{grootendorst2022bertopic,
  title={BERTopic: Neural topic modeling with a class-based TF-IDF procedure},
  author={Grootendorst, Maarten},
  journal={arXiv preprint arXiv:2203.05794},
  year={2022}
}

@article{salton1988term,
  title={Term-weighting approaches in automatic text retrieval},
  author={Salton, Gerard and Buckley, Christopher},
  journal={Information processing \& management},
  volume={24},
  number={5},
  pages={513--523},
  year={1988},
  publisher={Elsevier}
}

@article{ferrari2004beta,
    title={Beta regression for modelling rates and proportions},
    author={Ferrari, Silvia and Cribari-Neto, Francisco},
    journal={Journal of Applied Statistics},
    volume={31},
    number={7},
    pages={799--815},
    year={2004},
    publisher={Taylor \& Francis},
    doi={10.1080/0266476042000214501}
  }

@article{zhang2023siren,
  title={Siren’s song in the ai ocean: A survey on hallucination in large language models},
  author={Zhang, Yue and Li, Yafu and Cui, Leyang and Cai, Deng and Liu, Lemao and Fu, Tingchen and Huang, Xinting and Zhao, Enbo and Zhang, Yu and Chen, Yulong and others},
  journal={arXiv preprint arXiv:2309.01219},
  volume={2},
  number={5},
  year={2023}
}

@article{mcinnes2017hdbscan,
  title={hdbscan: Hierarchical density based clustering.},
  author={McInnes, Leland and Healy, John and Astels, Steve and others},
  journal={J. Open Source Softw.},
  volume={2},
  number={11},
  pages={205},
  year={2017}
}

@article{mcinnes2018umap,
  title={Umap: Uniform manifold approximation and projection for dimension reduction},
  author={McInnes, Leland and Healy, John and Melville, James},
  journal={arXiv preprint arXiv:1802.03426},
  year={2018}
}

@article{blei2001latent,
  title={Latent dirichlet allocation},
  author={Blei, David and Ng, Andrew and Jordan, Michael},
  journal={Advances in neural information processing systems},
  volume={14},
  year={2001}
}

@misc{LabelStudio,
  title={{Label Studio}: Data labeling software},
  url={https://github.com/HumanSignal/label-studio},
  note={Open source software available from https://github.com/HumanSignal/label-studio},
  author={
    Maxim Tkachenko and
    Mikhail Malyuk and
    Andrey Holmanyuk and
    Nikolai Liubimov},
  year={2020},
}

@article{benjamini1995controlling,
  title={Controlling the false discovery rate: a practical and powerful approach to multiple testing},
  author={Benjamini, Yoav and Hochberg, Yosef},
  journal={Journal of the Royal statistical society: series B (Methodological)},
  volume={57},
  number={1},
  pages={289--300},
  year={1995},
  publisher={Wiley Online Library}
}

@inproceedings{radford2023robust,
  title={Robust speech recognition via large-scale weak supervision},
  author={Radford, Alec and Kim, Jong Wook and Xu, Tao and Brockman, Greg and McLeavey, Christine and Sutskever, Ilya},
  booktitle={International conference on machine learning},
  pages={28492--28518},
  year={2023},
  organization={PMLR}
}

@article{pace2009stepwise,
  title={Stepwise logistic regression},
  author={Pace, Nathan L and Briggs, William M},
  journal={Anesthesia \& Analgesia},
  volume={109},
  number={1},
  pages={285--286},
  year={2009},
  publisher={LWW}
}

@article{reimers2019sentence,
  title={Sentence-bert: Sentence embeddings using siamese bert-networks},
  author={Reimers, Nils and Gurevych, Iryna},
  journal={arXiv preprint arXiv:1908.10084},
  year={2019}
}

@article{bagiella2000mixed,
  title={Mixed-effects models in psychophysiology},
  author={Bagiella, Emilia and Sloan, Richard P and Heitjan, Daniel F},
  journal={Psychophysiology},
  volume={37},
  number={1},
  pages={13--20},
  year={2000},
  publisher={Cambridge University Press}
}

@article{lissak2024bored,
  title={Bored to death: Artificial Intelligence research reveals the role of boredom in suicide behavior},
  author={Lissak, Shir and Ophir, Yaakov and Tikochinski, Refael and Brunstein Klomek, Anat and Sisso, Itay and Fruchter, Eyal and Reichart, Roi},
  journal={Frontiers in psychiatry},
  volume={15},
  pages={1328122},
  year={2024},
  publisher={Frontiers Media SA}
}

@article{caliendo2008some,
  title={Some practical guidance for the implementation of propensity score matching},
  author={Caliendo, Marco and Kopeinig, Sabine},
  journal={Journal of economic surveys},
  volume={22},
  number={1},
  pages={31--72},
  year={2008},
  publisher={Wiley Online Library}
}

@article{openai2023gpt,
  title={GPT-4 technical report},
  author={OpenAI, R and others},
  journal={ArXiv},
  volume={2303},
  pages={08774},
  year={2023}
}

@article{ophir2020deep,
  title={Deep neural networks detect suicide risk from textual facebook posts},
  author={Ophir, Yaakov and Tikochinski, Refael and Asterhan, Christa SC and Sisso, Itay and Reichart, Roi},
  journal={Scientific reports},
  volume={10},
  number={1},
  pages={16685},
  year={2020},
  publisher={Nature Publishing Group UK London}
}

@article{glenn2020annual,
  title={Annual research review: A meta-analytic review of worldwide suicide rates in adolescents},
  author={Glenn, Catherine R and Kleiman, Evan M and Kellerman, John and Pollak, Olivia and Cha, Christine B and Esposito, Erika C and Porter, Andrew C and Wyman, Peter A and Boatman, Anne E},
  journal={Journal of child psychology and psychiatry},
  volume={61},
  number={3},
  pages={294--308},
  year={2020},
  publisher={Wiley Online Library}
}

@article{naghavi2019global,
  title={Global, regional, and national burden of suicide mortality 1990 to 2016: systematic analysis for the Global Burden of Disease Study 2016},
  author={Naghavi, Mohsen},
  journal={bmj},
  volume={364},
  year={2019},
  publisher={British Medical Journal Publishing Group}
}

@article{fergusson2000risk,
  title={Risk factors and life processes associated with the onset of suicidal behaviour during adolescence and early adulthood},
  author={Fergusson, David M and Woodward, Lianne J and Horwood, L John},
  journal={Psychological medicine},
  volume={30},
  number={1},
  pages={23--39},
  year={2000},
  publisher={Cambridge University Press}
}

@article{pelkonen2003child,
  title={Child and adolescent suicide: epidemiology, risk factors, and approaches to prevention},
  author={Pelkonen, Mirjami and Marttunen, Mauri},
  journal={Pediatric Drugs},
  volume={5},
  pages={243--265},
  year={2003},
  publisher={Springer}
}

@article{franklin2017risk,
  title={Risk factors for suicidal thoughts and behaviors: A meta-analysis of 50 years of research.},
  author={Franklin, Joseph C and Ribeiro, Jessica D and Fox, Kathryn R and Bentley, Kate H and Kleiman, Evan M and Huang, Xieyining and Musacchio, Katherine M and Jaroszewski, Adam C and Chang, Bernard P and Nock, Matthew K},
  journal={Psychological bulletin},
  volume={143},
  number={2},
  pages={187},
  year={2017},
  publisher={American Psychological Association}
}

@article{castillo2020suicide,
  title={Suicide risk assessment using machine learning and social networks: a scoping review},
  author={Castillo-S{\'a}nchez, Gema and Marques, Gon{\c{c}}alo and Dorronzoro, Enrique and Rivera-Romero, Octavio and Franco-Mart{\'\i}n, Manuel and De la Torre-D{\'\i}ez, Isabel},
  journal={Journal of medical systems},
  volume={44},
  number={12},
  pages={205},
  year={2020},
  publisher={Springer}
}

@article{rabani2023detecting,
  title={Detecting suicidality on social media: Machine learning at rescue},
  author={Rabani, Syed Tanzeel and Khanday, Akib Mohi Ud Din and Khan, Qamar Rayees and Hajam, Umar Ayoub and Imran, Ali Shariq and Kastrati, Zenun},
  journal={Egyptian Informatics Journal},
  volume={24},
  number={2},
  pages={291--302},
  year={2023},
  publisher={Elsevier}
}

@article{fu2013responses,
  title={Responses to a self-presented suicide attempt in social media},
  author={Fu, King-wa and Cheng, Qijin and Wong, Paul WC and Yip, Paul SF},
  journal={Crisis},
  year={2013},
  publisher={Hogrefe Publishing}
}

@article{lissak2024colorful,
  title={The Colorful Future of LLMs: Evaluating and Improving LLMs as Emotional Supporters for Queer Youth},
  author={Lissak, Shir and Calderon, Nitay and Shenkman, Geva and Ophir, Yaakov and Fruchter, Eyal and Klomek, Anat Brunstein and Reichart, Roi},
  journal={arXiv preprint arXiv:2402.11886},
  year={2024}
}

@article{o2011impact,
  title={The impact of social media on children, adolescents, and families},
  author={O'Keeffe, Gwenn Schurgin and Clarke-Pearson, Kathleen and others},
  journal={Pediatrics},
  volume={127},
  number={4},
  pages={800--804},
  year={2011},
  publisher={American Academy of Pediatrics}
}

@article{dagar2020high,
  title={High viewership of videos about teenage suicide on YouTube.},
  author={Dagar, Anjali and Falcone, Tatiana},
  year={2020},
  publisher={Elsevier Science}
}

@article{khasawneh2020examining,
  title={Examining the self-harm and suicide contagion effects of the Blue Whale Challenge on YouTube and Twitter: qualitative study},
  author={Khasawneh, Amro and Chalil Madathil, Kapil and Dixon, Emma and Wi{\'s}niewski, Pamela and Zinzow, Heidi and Roth, Rebecca},
  journal={JMIR mental health},
  volume={7},
  number={6},
  pages={e15973},
  year={2020},
  publisher={JMIR Publications Toronto, Canada}
}

@article{allgaier2020science,
  title={Science and medicine on YouTube},
  author={Allgaier, Joachim},
  journal={Second international handbook of Internet research},
  pages={7--27},
  year={2020},
  publisher={Springer}
}

@article{khan2022researching,
  title={Researching YouTube: Methods, tools, and analytics},
  author={Khan, M Laeeq and Malik, Aqdas},
  journal={The sage handbook of social media research methods},
  pages={651--663},
  year={2022}
}

@article{touvron2023llama,
  title={Llama 2: Open foundation and fine-tuned chat models},
  author={Touvron, Hugo and Martin, Louis and Stone, Kevin and Albert, Peter and Almahairi, Amjad and Babaei, Yasmine and Bashlykov, Nikolay and Batra, Soumya and Bhargava, Prajjwal and Bhosale, Shruti and others},
  journal={arXiv preprint arXiv:2307.09288},
  year={2023}
}

@article{luoma2002contact,
  title={Contact with mental health and primary care providers before suicide: a review of the evidence},
  author={Luoma, Jason B and Martin, Catherine E and Pearson, Jane L},
  journal={American Journal of Psychiatry},
  volume={159},
  number={6},
  pages={909--916},
  year={2002},
  publisher={Am Psychiatric Assoc}
}

@article{bryan2019examining,
  title={Examining emotion relief motives as a facilitator of the transition from suicidal thought to first suicide attempt among active duty soldiers.},
  author={Bryan, Craig J and May, Alexis M and Harris, Julia},
  journal={Psychological services},
  volume={16},
  number={2},
  pages={293},
  year={2019},
  publisher={Educational Publishing Foundation}
}

@article{kaplow2014emotional,
  title={Emotional suppression mediates the relation between adverse life events and adolescent suicide: Implications for prevention},
  author={Kaplow, Julie B and Gipson, Polly Y and Horwitz, Adam G and Burch, Bianca N and King, Cheryl A},
  journal={Prevention science},
  volume={15},
  pages={177--185},
  year={2014},
  publisher={Springer}
}

@article{frey2019recipients,
  title={Recipients of suicide-related disclosure: the link between disclosure and posttraumatic growth for suicide attempt survivors},
  author={Frey, Laura M and Drapeau, Christopher W and Fulginiti, Anthony and Oexle, Nathalie and Stage, Dese’Rae L and Sheehan, Lindsay and Cerel, Julie and Moore, Melinda},
  journal={International journal of environmental research and public health},
  volume={16},
  number={20},
  pages={3815},
  year={2019},
  publisher={MDPI}
}

@article{kirchner2024experiences,
  title={Experiences of suicide survivors of sharing their stories about suicidality and overcoming a crisis in media and public talks: a qualitative study},
  author={Kirchner, Stefanie and Niederkrotenthaler, Thomas},
  journal={BMC public health},
  volume={24},
  number={1},
  pages={142},
  year={2024},
  publisher={Springer}
}

@article{aquino2024youtube,
  title={YouTube Influencers Fostering Audience Engagement Through Parasocial Relationships},
  author={Aquino, Elaina and Yang, Kiseol and Brandon, Lynn},
  journal={Journal of Marketing Development and Competitiveness},
  volume={18},
  number={2},
  pages={67--81},
  year={2024},
  publisher={North American Business Press}
}

@article{chung2023really,
  title={I really know you: how influencers can increase audience engagement by referencing their close social ties},
  author={Chung, Jaeyeon and Ding, Yu and Kalra, Ajay},
  journal={Journal of Consumer Research},
  volume={50},
  number={4},
  pages={683--703},
  year={2023},
  publisher={Oxford University Press}
}

@article{zhang2023exploring,
  title={Exploring audience engagement in YouTube vlogs through consumer engagement theory: the case of UK beauty vlogger Zoe Sugg.},
  author={Zhang, Hantian and Lee, John},
  journal={First Monday},
  volume={28},
  number={4},
  year={2023},
  publisher={University of Illinois at Chicago Library}
}

@article{sheehy2019examination,
  title={An examination of the relationship between shame, guilt and self-harm: A systematic review and meta-analysis},
  author={Sheehy, Kate and Noureen, Amna and Khaliq, Ayesha and Dhingra, Katie and Husain, Nusrat and Pontin, Eleanor E and Cawley, Rosanne and Taylor, Peter J},
  journal={Clinical psychology review},
  volume={73},
  pages={101779},
  year={2019},
  publisher={Elsevier}
}

@article{mcgill2019information,
  title={Information needs of people after a suicide attempt: A thematic analysis},
  author={McGill, Katie and Hackney, Sue and Skehan, Jaelea},
  journal={Patient education and counseling},
  volume={102},
  number={6},
  pages={1119--1124},
  year={2019},
  publisher={Elsevier}
}

@article{maple2020nobody,
  title={“Nobody hears a silent cry for help”: Suicide attempt survivors’ experiences of disclosing during and after a crisis},
  author={Maple, Myfanwy and Frey, Laura M and McKay, Kathy and Coker, Sarah and Grey, Samara},
  journal={Archives of suicide research},
  volume={24},
  number={4},
  pages={498--516},
  year={2020},
  publisher={Taylor \& Francis}
}

@article{russell2016mental,
  title={Mental health in lesbian, gay, bisexual, and transgender (LGBT) youth},
  author={Russell, Stephen T and Fish, Jessica N},
  journal={Annual review of clinical psychology},
  volume={12},
  number={1},
  pages={465--487},
  year={2016},
  publisher={Annual Reviews}
}

@article{trevor,
  title={Trevor Project explores MH of multiracial LGBTQ youth},
  author={Canady, Valerie A},
  journal={Mental Health Weekly},
  volume={32},
  number={33},
  pages={7--8},
  year={2022},
  publisher={Wiley Online Library}
}

@article{meyer2003prejudice,
  title={Prejudice, social stress, and mental health in lesbian, gay, and bisexual populations: conceptual issues and research evidence.},
  author={Meyer, Ilan H},
  journal={Psychological bulletin},
  volume={129},
  number={5},
  pages={674},
  year={2003},
  publisher={American Psychological Association}
}

@article{dworkin2022associations,
  title={Associations between sexual assault and suicidal thoughts and behavior: A meta-analysis.},
  author={Dworkin, Emily R and DeCou, Christopher R and Fitzpatrick, Skye},
  journal={Psychological trauma: theory, research, practice, and policy},
  volume={14},
  number={7},
  pages={1208},
  year={2022},
  publisher={Educational Publishing Foundation}
}

@article{sutter2016discrimination,
  title={Discrimination, mental health, and suicidal ideation among LGBTQ people of color.},
  author={Sutter, Megan and Perrin, Paul B},
  journal={Journal of counseling psychology},
  volume={63},
  number={1},
  pages={98},
  year={2016},
  publisher={American Psychological Association}
}

@book{joiner2005people,
  title={Why people die by suicide},
  author={Joiner, Thomas},
  year={2005},
  publisher={Harvard University Press}
}

@article{moody2010exploratory,
  title={An exploratory study: Relationships between trying on clothing, mood, emotion, personality and clothing preference},
  author={Moody, Wendy and Kinderman, Peter and Sinha, Pammi},
  journal={Journal of Fashion Marketing and Management: An International Journal},
  volume={14},
  number={1},
  pages={161--179},
  year={2010},
  publisher={Emerald Group Publishing Limited}
}

@article{masuch2014understanding,
  title={Understanding the links between positive psychology and fashion: A grounded theory analysis},
  author={Masuch, Christoph-Simon and Hefferon, Kate},
  journal={International Journal of Fashion Studies},
  volume={1},
  number={2},
  pages={227--246},
  year={2014},
  publisher={Intellect}
}

@article{blei2003latent,
  title={Latent dirichlet allocation},
  author={Blei, David M and Ng, Andrew Y and Jordan, Michael I},
  journal={Journal of machine Learning research},
  volume={3},
  number={Jan},
  pages={993--1022},
  year={2003}
}

@article{yang2025enhancing,
  title={Enhancing topic coherence and diversity in document embeddings using LLMs: A focus on BERTopic},
  author={Yang, Chibok and Kim, Yangsok},
  journal={Expert Systems with Applications},
  volume={281},
  pages={127517},
  year={2025},
  publisher={Elsevier}
}

@article{mu2024large,
  title={Large language models offer an alternative to the traditional approach of topic modelling},
  author={Mu, Yida and Dong, Chun and Bontcheva, Kalina and Song, Xingyi},
  journal={arXiv preprint arXiv:2403.16248},
  year={2024}
}

@article{xiao2025addictive,
  title={Addictive screen use trajectories and suicidal behaviors, suicidal ideation, and mental health in US youths},
  author={Xiao, Yunyu and Meng, Yuan and Brown, Timothy T and Keyes, Katherine M and Mann, J John},
  journal={JAMA},
  year={2025}
}

@inproceedings{de2013predicting,
  title={Predicting depression via social media},
  author={De Choudhury, Munmun and Gamon, Michael and Counts, Scott and Horvitz, Eric},
  booktitle={Proceedings of the international AAAI conference on web and social media},
  volume={7},
  number={1},
  pages={128--137},
  year={2013}
}

@article{bagge2017trajectories,
  title={Trajectories of affective response as warning signs for suicide attempts: An examination of the 48 hours prior to a recent suicide attempt},
  author={Bagge, Courtney L and Littlefield, Andrew K and Glenn, Catherine R},
  journal={Clinical Psychological Science},
  volume={5},
  number={2},
  pages={259--271},
  year={2017},
  publisher={Sage Publications Sage CA: Los Angeles, CA}
}

@article{zhang2022neural,
  title={Is neural topic modelling better than clustering? an empirical study on clustering with contextual embeddings for topics},
  author={Zhang, Zihan and Fang, Meng and Chen, Ling and Namazi-Rad, Mohammad-Reza},
  journal={arXiv preprint arXiv:2204.09874},
  year={2022}
}

@misc{brightdata_academic,
  author       = {{Bright Data}},
  title        = {Bright Initiative Academy},
  howpublished = {\url{https://brightinitiative.com/academy}},
  note         = {Accessed: November 11, 2025}
}

\newpage
\appendix

\section*{Appendix Roadmap}

The Appendix provides methodological and supplementary information supporting the findings presented in the main manuscript; organized into five sections:
\begin{itemize}
    \item \textbf{Appendix A – Data Collection:} Describes the procedures used for data acquisition, inclusion and exclusion criteria, human evaluation, and the multi-step matching algorithm applied to construct the Control (Matches).
    \item \textbf{Appendix B – Procedures:} Details the computational and statistical pipelines, including topic modeling, temporal alignment, and temporal testing procedures. 
\item \textbf{Appendix C – Supplementary Results:} Provides extended descriptive results and illustrative figures that complement the main findings. This section includes additional topic modeling illustrations, visual examples from the bottom-up, top-down, and hybrid analyses, and longitudinal visualizations.
\item \textbf{Appendix D – Statistical and Robustness Analyses:} Presents the full statistical formulation underlying the main results, and robustness checks conducted to assess potential confounding factors.
    \item \textbf{Appendix E – Topic Information:} Provides the full list of topics identified in the study, including representative transcript segments, top words, and topic-level correlations.
\end{itemize}

\tableofcontents

\section{Appendix A - Data Collection}

\noindent
This appendix provides a detailed account of the data collection and cohort construction processes supporting the methods section (\S2) of the main manuscript. The data collection pipeline (Figure~\ref{fig:high_level_method_pipeline}) was designed to ensure both clinical validity and methodological rigor, using inclusion and exclusion criteria to define eligible samples (\S\ref{inclusion_exclusion_appendix}), automated data preprocessing and transcription steps (\S\ref{data_preprocessing_appendix}), a two-tiered human evaluation protocol combining non-expert and clinical psychological review (\S\ref{human_eval_protocol_full}), and a multi-step matching framework for constructing demographically and contextually comparable control cohorts (\S\ref{matching_full}). The resulting dataset and subgroup characteristics are summarized in Table \ref{dataset_stats}.

\begin{table}[ht]
\caption{Descriptive statistics of the research groups. Summary of demographics and YouTube activity. See \S{\ref{validation_analysis}} for upload trends pre- and post-event. Minority groups are defined as individuals who are non-Caucasian or non-heterosexual.}
\label{dataset_stats}
\scriptsize
\begin{tabular*}{0.95\linewidth}{@{\extracolsep{\fill}} lcccccc@{}}
\toprule
Group & \# Users & Uploads & Mean / Med / SD & Female & Age & Minority \\
 & & (Total) & & (n, \%) & (Mean) & (n, \%) \\
\midrule
Attempted (During) & 42 & 6,354 & 151 / 89 / 169 & 30 (71\%) & 24 & 27 (59\%) \\
Attempted (Before) & 139 & 28,053 & 201 / 83 / 312 & 91 (65\%) & 26 & 78 (56\%) \\
Control (Major Life Event) & 43 & 13,470 & 320 / 155 / 372 & 29 (69\%) & 27 & 15 (35\%) \\
Control (Matches) & 95 & 14,860 & 182 / 95 / 223 & 79 (81\%) & 23 & 52 (54\%) \\
\bottomrule
\end{tabular*}
\end{table}

\setcounter{table}{0}
\renewcommand{\thetable}{A.\arabic{table}}
\setcounter{figure}{0}
\renewcommand{\thefigure}{A.\arabic{figure}}

\begin{figure}[ht]
    \centering
    \includegraphics[width=\textwidth]{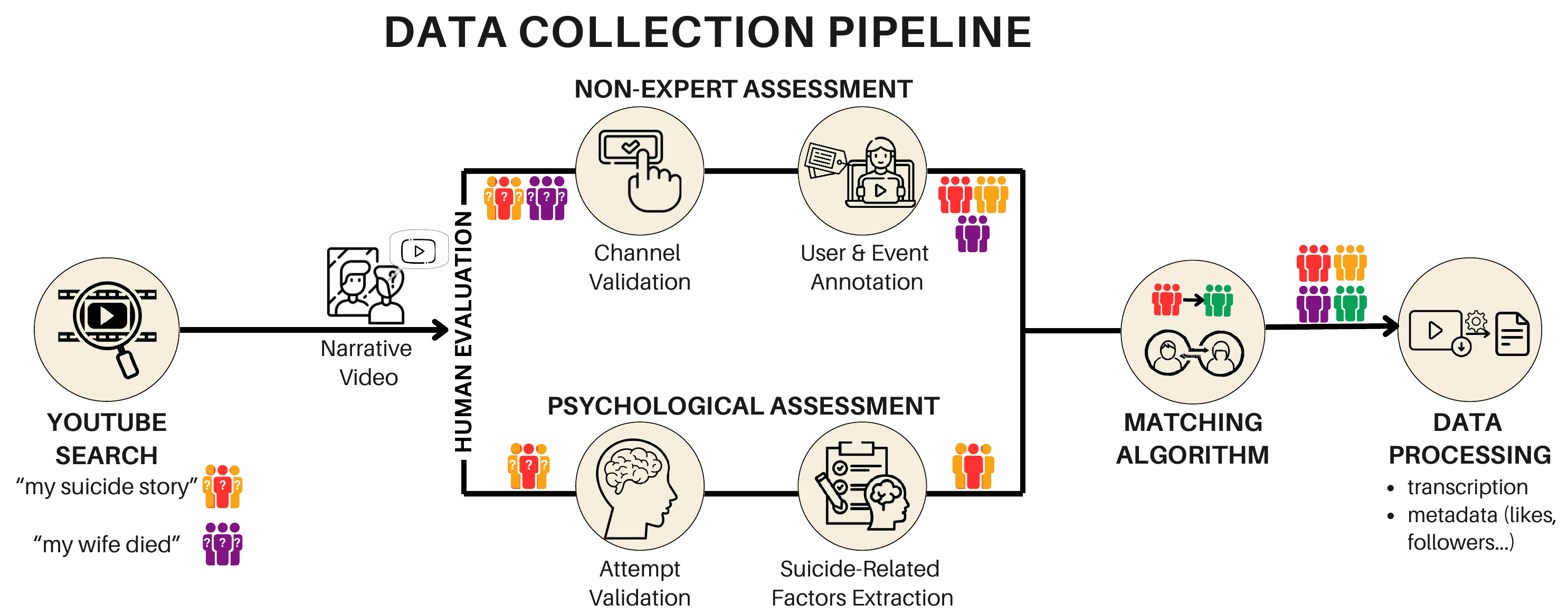} 
    \caption{The data collection pipeline includes \textbf{searching} for relevant YouTubers (e.g., YouTubers who attempted suicide) (\ref{inclusion_exclusion_appendix}), a two-step \textbf{human evaluation} (\ref{human_eval_protocol_full}), involving \textit{non-expert assessment} (personal channel validation and extracting demographic and event-related information), and \textit{psychological assessment} (attempt intent validation and suicide-related factor extraction); both steps validating and then analyzing the narrative stories describing a \textit{suicide attempt} or a \textit{major life event}. The final steps involved control-based \textbf{matching algorithm}, which involved finding "similar" YouTube channels to the \textit{treatment} group (Figure \ref{fig:matching_algorithm}). The last step included performing \textbf{data processing} (including filtering uploads and automatic transcription) on the 4 research groups (\ref{data_preprocessing_appendix}).}
    \label{fig:high_level_method_pipeline}
\end{figure}

\subsection{Inclusion and exclusion criteria} \label{inclusion_exclusion_appendix}

\begin{table*}[h]
\centering
\scriptsize
\renewcommand{\arraystretch}{1.2}
\begin{tabular}{|p{1.65cm}|p{1.7cm}|p{2.3cm}|p{2.3cm}|p{2.5cm}|p{2cm}|}
\hline
\textbf{Stage (Scope)} & \textbf{Filtering Step} & \multicolumn{4}{c|}{\textbf{Group}} \\
\cline{3-6}
& & \makecell{\textbf{Attempted}\\(During)} & \makecell{\textbf{Attempted}\\(Before)} & \makecell{\textbf{Control}\\(Major Life Event)} & \makecell{\textbf{Control}\\(Matches)} \\
\hline

Search (Video) & Search Queries (OR) & 
\multicolumn{2}{c|}{\begin{tabular}[c]{@{}c@{}}``my suicide attempt''\\``my suicide story''\\``I am a suicide attempt survivor''\\``attempted suicide''\\``take my life''\end{tabular}} & 
\multicolumn{1}{c|}{\begin{tabular}[c]{@{}c@{}}
``my \textit{relative} died'' \\
``losing my \textit{relative}'' \\
``I lost my \textit{relative}'' \\
\textit{relative}  $\in$ \{brother, sister,\\ father, mother\}
\end{tabular}} & 
\begin{tabular}[c]{@{}c@{}}Automated channel- \\specific queries using:\\• Frequent bigrams\\• Video categories\\• Upload periods\end{tabular} \\
\hline

Search  (Video) & Category &\multicolumn{4}{c|}{\begin{tabular}[c]{@{}c@{}}
\textbf{Excluded} query videos from the following categories: \\
\texttt{["Gaming", "Sports", "Comedy", "Music", "News \& Politics",} \\
\texttt{"Pets \& Animals", "Science \& Technology", "Travel \& Events", "Film \& Animation"]}
\end{tabular}} \\ \\
\hline

Search (Video) & Content & \multicolumn{4}{p{13cm}|}{
\begin{tabular}[c]{@{}c@{}} \textbf{Excluded} query videos that appeared to be: \\ • clickbait (e.g., \textit{"attempt... [Mha gacha] [Warning:Suicide]"}) \\ • news reports (e.g., \textit{"Coco Lee dies after suicide attempt"}) \\ • non-personal stories (e.g., \textit{"TONI Ep.104 | Molested and Attempted Suicide"}) \end{tabular}
} \\
\hline

Search (Channel) & Personal Accounts & \multicolumn{4}{c|}{\textbf{Included} personal accounts validated via profile picture, channel link and sample video} \\
\hline

Preprocessing (Video) & Content & \multicolumn{4}{c|}{\textbf{Included} Only public, English, downloadable videos, longer than 60 seconds.} \\
\hline

Preprocessing (Video) & File Size Limit & \multicolumn{3}{c|}{Not applied} & \textbf{Excluded} $>$50MB \\
\hline

Preprocessing (Transcript) & Transcripts & \multicolumn{4}{p{12.5cm}|}{
\textbf{Excluded} transcripts if: (1) $<$3 sentences, (2) $>$50\% repeated (hallucinated) content, (3) $>$30\% non-English, or (4) $>$2 speakers. Videos which weren't filtered are considered \textit{valid}.} \\
\hline

Preprocessing (Channel) & Activity Threshold & \multicolumn{4}{c|}{\textbf{Excluded} channels with $<$10 \textit{valid} videos} \\
\hline

Assessment (Mixed) & Group-Specific Procedure & \multicolumn{2}{c|}{Psychological assessment + Event validation} & Event validation & Matching Procedure \\
\hline

-- & Final Sample Size & \textit{n}=42 & \textit{n}=139 & \textit{n}=43 & \textit{n}=95 \\
\hline

\end{tabular}

\caption{\textbf{Inclusion and exclusion criteria for the four research groups}. The leftmost column indicates the pipeline stage and data scope at which each filtering step was applied (e.g., during search, query filtering, or preprocessing). Label Studio was used for manual review.}
\label{tab:inclusion_exclusion_criteria}
\end{table*}

As part of the data collection, we used the YouTube API, \href{https://developers.google.com/youtube/v3/docs}{Google YouTube API Documentation}, to identify channels relevant to our study. These included creators who (a) reported a suicide attempt — \textit{Attempted (During)} and \textit{Attempted (Before)} groups, (b) experienced a major life event — \textit{Control (Major Life Event)}, and (c) represented the general YouTube population — \textit{Control (Matches)}. The full inclusion and exclusion criteria for each research group are provided in Table~\ref{tab:inclusion_exclusion_criteria}. The sample consists exclusively of English-language YouTube channels identified through English search queries (Table~\ref{tab:inclusion_exclusion_criteria}). Geographic location was not systematically coded, limiting our ability to describe the precise regional distribution. Demographic characteristics including ethnicity were assessed by annotators based on explicit self-disclosure in videos where available (Appendix~\ref{human_eval_protocol_full}). The classification of "minority" (Table~\ref{dataset_stats}) combines ethnicity (non-Caucasian) and sexual orientation (non-heterosexual). We acknowledge that these limitations constrain generalizability to non-English-speaking and non-Western populations.

\subsection{Data preprocessing \label{data_preprocessing_appendix}}

As part of the data collection, we implemented the following preprocessing steps:
\begin{enumerate}
    \item \textbf{Downloading}: Channel uploads were obtained via the ~\citep{brightdata_academic} service under its Academic Program.
    \item \textbf{Whisper Transcription:} We used the Whisper (medium) \citep{radford2023robust}, model for audio-to-text conversion with specific parameters to enhance transcription accuracy (no speech threshold=0.3, condition on previous text=True).
    \item \textbf{Data Cleaning:} \label{data_cleaning} 
To ensure dataset quality and relevance, we \textbf{excluded} transcripts that match one of the following criteria :
\begin{enumerate}
\item Too short - fewer than three transcribed sentences or less than 1 minutes of audio.
\item Contain hallucinations - repeated words or sentences comprising more than 40\% of the document, often resulting from non-verbal content like background music \citep{zhang2023siren}.
\item Non-English Content - contain more than 50\% non-English content (detected automatically).
\end{enumerate}
The remaining videos were labeled as \textbf{valid}.  See Table \ref{tab:validity} which summarizes uploads validity.  
\item \textbf{Transcript Segmentation:} To accommodate the token limits of modern embedding models like Sentence-BERT \citep{reimers2019sentence} (which has a 512-token threshold), we divided transcripts into chunks of a maximum of 80 tokens, ensuring each chunk contained complete sentences. This segmentation allows for better representation of multiple topics within a single video.

\begin{table}[ht!]
\scriptsize
\begin{tabular}{lcc|cccc}
\toprule
\textbf{Group} & \textbf{Valid} & \textbf{Invalid} & \textbf{Short} & \textbf{Non-Eng} & \textbf{Short} & \textbf{Halluc.} \\
 & \textbf{(\%)} & \textbf{(\%)} & \textbf{Text (\%)} & \textbf{(\%)} & \textbf{Audio (\%)} & \textbf{(\%)} \\
\midrule
Attempted (Before) & 86.4 & 13.6 & 20.3 & 42.1 & 46.5 & 5.7 \\
Attempted (During) & 86.7 & 13.3 & 17.0 & 41.4 & 45.0 & 3.8 \\
Control (Major Life Event) & 90.1 & 9.9 & 15.9 & 65.4 & 0.18 & 0.9 \\
Control (Matches) & 94.3 & 5.7 & 10 & 53 & 10 & 11 \\
\bottomrule
\end{tabular}
\caption{Summary of upload validity by research group, expressed as percentages. 
Columns \textit{Valid} and \textit{Invalid} indicate the proportion of total uploads meeting validity criteria, 
while \textit{Short Text}, \textit{Non-English}, \textit{Short Audio}, and \textit{Hallucinated} show the distribution of failure reasons among invalid uploads. 
Values represent means across YouTubers in each group.}
\label{tab:validity}
\end{table}

\end{enumerate}

\subsection{Human evaluation}  \label{human_eval_protocol_full} 

As part of our annotation process, we used the Label Studio \citep{LabelStudio} platform to create two protocols. In both, the annotators were provided with a YouTuber image, Link to the YouTuber's channel, Attempt/Major Life Event narrative's video, accompanied with audio, and transcription of the video, and summary (automatically generated by GPT-4 \citep{openai2023gpt}).
\phantomsection \label{data_collection_human_eval_basic}

The \textbf{Non-Expert Assessment} step covered three broad areas: demographic information (e.g., gender, age, ethnicity), channel-based features (e.g., confirmation of first-person accounts, personal channels), and event-related details (e.g., method of suicide attempt, such as medication, for \textit{attempts}; or type of major life event, e.g., sudden death of relative). In addition, the date of the suicide attempt and the major life event was extracted as part of the event-related information, and defined as the \textit{reference points}. The dates were recorded, with some users providing specific dates (e.g., 02-05-2017) and others providing approximate estimates (e.g., beginning of the year, October 2017). Not all groups had a specific event date during their upload period (i.e., Attempted (Before) and Control (Matches)), thus, we also created \textit{synthetic reference events} for different groups (e.g., middle of the uploading period). These \textit{reference events} allowed us to control for the significance of the suicide attempt compared to other reference events. 

This annotation was performed on Attempted (During), Attempted (Before), and Control (Major Life Event). 

The annotators were asked to answer the following questions:
\begin{itemize}
    \item \textbf{Relevance} - 
    \begin{itemize}
        \item What type of channel is it? Personal (individual who uploads content about themselves) or General (e.g., family channel, news outlet, educational content)
    \end{itemize}
    \item \textbf{Demographic Information} - age, ethnicity, gender.
    \item \textbf{Event information} -    
    \begin{itemize}
        \item Does the story talk about a suicide attempt/major life event?
        \item Who did the attempt/died?
        \item What is the channel owner's age at the time of the event?
        \item What are the date(s) of event(s)? we asked for both specific and estimated date(s). 
        \item How did the attempt/major life event happen? e.g., suicide attempt using medication, or sudden death of a relative.
    \end{itemize}

\phantomsection \label{data_collection_human_eval_psyc}

The \textbf{Psychological Assessment} step included psychological analysis of predefined relevant \textit{factors} including suicide risk factors and additional info, and \textit{verification} of suicide intent. A trained clinical psychologist reviewed the \textit{suicide attempt narratives}, verifying that the individual describes a suicide attempt rather than self-harm and also conducting a qualitative analysis of risk factor-related information.  This annotation was performed \textit{only} on the Attempted (During) and Attempted (Before) groups and included the following questions:

\textbf{    \item Risk Factors Information:
}    \begin{itemize}
        \item What risk factors for suicide are mentioned (e.g., depression)?
        \item What resilience factors are mentioned (e.g., family support)?
        \item What risk group does the individual belong to  (e.g., LGBTQ+ individuals)?
        \item What appears to be the motivation for posting this video  (e.g., raising awareness)?
    \end{itemize}
    \item \textbf{  Clinical Information}
    \begin{itemize}
        \item Does the text's main speaker mention having any mental illness  (e.g., PTSD)?
        \item Does the text's main speaker mention being treated by a professional psychologist or psychiatrist and/or being hospitalized?
        \item What is the suicide attempt intent (Suicide Attempt, Suicidal Ideation, or Non-Suicidal Self-Harm)?
    \end{itemize}
\end{itemize}

\phantomsection \label{inter_ann_agreement}
\paragraph{Inter-Annotator Agreement}

To evaluate the reliability of the psychological annotations, a second clinical psychologist independently re-annotated a subset of 20 suicide-attempt videos. Before this process, five videos were jointly reviewed to harmonize interpretation and resolve potential ambiguities. Agreement was quantified across 15 annotated fields grouped into four domains: \textit{Demographics}, \textit{Attempt Information}, \textit{Risk and Protective Factors}, and \textit{Mental Health Indicators}.  The second annotation, conducted approximately one year after the initial assessment, served as an independent validation. 

Inter-rater reliability scores (Table~\ref{tab:agreement}) indicated overall \textbf{substantial to excellent reliability} for the core variables in this suicide-narrative annotation task.
  
Panel-level analysis revealed the highest reliability for \textit{Mental Health Information} (Jaccard $= 0.82$, $\kappa = 0.79$) and \textit{Risk Factors} (Jaccard $= 0.79$, $\kappa = 0.62$), followed by \textit{Demographics} (Jaccard $= 0.68$, $\kappa = 0.37$) and \textit{Attempt Information}
(Jaccard $= 0.64$, $\kappa = 0.43$).  At the field level, excellent agreement (Jaccard $\geq 0.8$) was achieved for primary variables including \textit{video motivation} ($0.94$), \textit{gender} ($0.92$), \textit{attempt intent} ($0.88$), \textit{mental illness} ($0.84$), and \textit{risk groups} ($0.84$). Lower agreement (Jaccard $< 0.6$) was observed for more interpretive constructs such as \textit{protective factors} ($0.58$) and \textit{age estimation} ($0.46-0.55$).  Category-level analysis (Table~\ref{tab:risk_factors_drilldown}) further revealed that common psychiatric symptoms (\textit{depression}, \textit{hopelessness}, \textit{unbearable mental pain}) achieved excellent agreement ($0.81--0.88$), while protective factors showed substantial variability, with  \textit{treatment} ($0.33$) and \textit{responsibility for others} ($0.50$) exhibiting particularly low reliability.

\begin{table}[htbp]
\centering
\caption{Inter-Annotator Agreement Metrics by Panel and Feature}
\label{tab:agreement}
\small
Agreement metrics: Jaccard similarity $J = \frac{A \cap B}{A \cup B}$;
Percent Agreement $P = \frac{n_{\text{agree}}}{n_{\text{total}}}$;
Cohen's $\kappa = \frac{P_o - P_e}{1 - P_e}$.
\vspace{0.3em}

\begin{tabular}{llcccc}
\toprule
\textbf{Panel} & \textbf{Feature} & \textbf{Jaccard} & \textbf{\% Agree} & \textbf{Cohen's $\kappa$} & \textbf{Agreement Level} \\
\midrule
\multirow{6}{*}{\begin{tabular}[c]{@{}l@{}}Attempt\\Information\end{tabular}}
& Attempt thoughts & 0.875 & 0.750 & 0.621 & Excellent \\
& Number of attempts & 0.792 & 0.792 & 0.602 & Good \\
& Instrument/method & 0.667 & 0.583 & 0.555 & Good \\
& Age at attempt & 0.552 & 0.500 & 0.350 & Moderate \\
\midrule
\multirow{2}{*}{\begin{tabular}[c]{@{}l@{}}Mental Health\\Information\end{tabular}}
& Mental illness & 0.840 & 0.750 & 0.865 & Excellent \\
& Hospital/therapy & 0.806 & 0.667 & 0.707 & Excellent \\
\midrule
\multirow{4}{*}{\begin{tabular}[c]{@{}l@{}}Risk\\Factors\end{tabular}}
& Video motivation & 0.938 & 0.917 & 0.637 & Excellent \\
& Risk groups & 0.840 & 0.792 & 0.621 & Excellent \\
& Risk factors & 0.807 & 0.375 & 0.701 & Excellent \\
& Protective factors & 0.583 & 0.375 & 0.505 & Moderate \\
\midrule
\multirow{3}{*}{Demographics}
& Gender & 0.917 & 0.917 & 0.600 & Excellent \\
& Ethnicity & 0.667 & 0.625 & 0.390 & Good \\
& Age at story & 0.458 & 0.458 & 0.128 & Moderate \\
\bottomrule
\end{tabular}
\begin{tablenotes}
\small
\item \textit{Agreement Levels:} Excellent (Jaccard $\geq$ 0.80); Good (0.60 $\leq$ Jaccard $<$ 0.80); Moderate (Jaccard $<$ 0.60)
\item Cohen's Kappa is undefined (N/A) when there is insufficient variation in the data
\end{tablenotes}
\end{table}

\begin{table}[htbp]
\centering
\caption{Category-level agreement for multiple-choice fields within the Risk Factors panel. Agreement represents the proportion of concordant annotations.
}
\label{tab:risk_factors_drilldown}
\vspace{0.3em}

\begin{tabular}{llccc}
\toprule
\textbf{Field} & \textbf{Category} & \textbf{Agreement} & \textbf{N} & \textbf{Level} \\
\midrule
\multirow{6}{*}{\begin{tabular}[c]{@{}l@{}}Risk\\Factors\end{tabular}}
& Unbearable mental pain & 0.882 & 17 & Excellent \\
& Hopelessness/despair & 0.842 & 19 & Excellent \\
& Depression & 0.810 & 21 & Excellent \\
& Life events & 0.778 & 9 & Good \\
& Comorbidity & 0.778 & 9 & Good \\
& Loneliness & 0.667 & 18 & Good \\
& Lack of belonging & 0.667 & 6 & Good \\
\midrule
\multirow{4}{*}{\begin{tabular}[c]{@{}l@{}}Protective\\Factors\end{tabular}}
& Social/family support & 0.769 & 13 & Good \\
& Responsibility for others & 0.500 & 10 & Moderate \\
& Treatment & 0.333 & 9 & Moderate \\
& Other & 0.400 & 5 & Moderate \\
\midrule
\multirow{4}{*}{\begin{tabular}[c]{@{}l@{}}Risk\\Groups\end{tabular}}
& Immigrants & 1.000 & 8 & Excellent \\
& Men & 1.000 & 6 & Excellent \\
& Teens & 0.786 & 14 & Good \\
& None & 0.500 & 6 & Moderate \\
\midrule
\multirow{3}{*}{\begin{tabular}[c]{@{}l@{}}Video\\Motivation\end{tabular}}
& Desire to help others & 0.947 & 19 & Excellent \\
& Personal recovery process & 0.923 & 13 & Excellent \\
& Raising awareness & 0.667 & 6 & Good \\
\bottomrule
\end{tabular}
\begin{tablenotes}
\footnotesize
\item \textit{Agreement Levels:} Excellent (Agreement $\geq$ 0.80); Good (0.60 $\leq$ Agreement $<$ 0.80); Moderate (Agreement $<$ 0.60)
\item Only categories with N $\geq$ 5 instances are included to ensure statistical reliability
\item N = number of instances where at least one annotator selected the category
\end{tablenotes}
\end{table}

\begin{figure}[htbp]
    \centering
   \includegraphics[height=0.85\textheight]{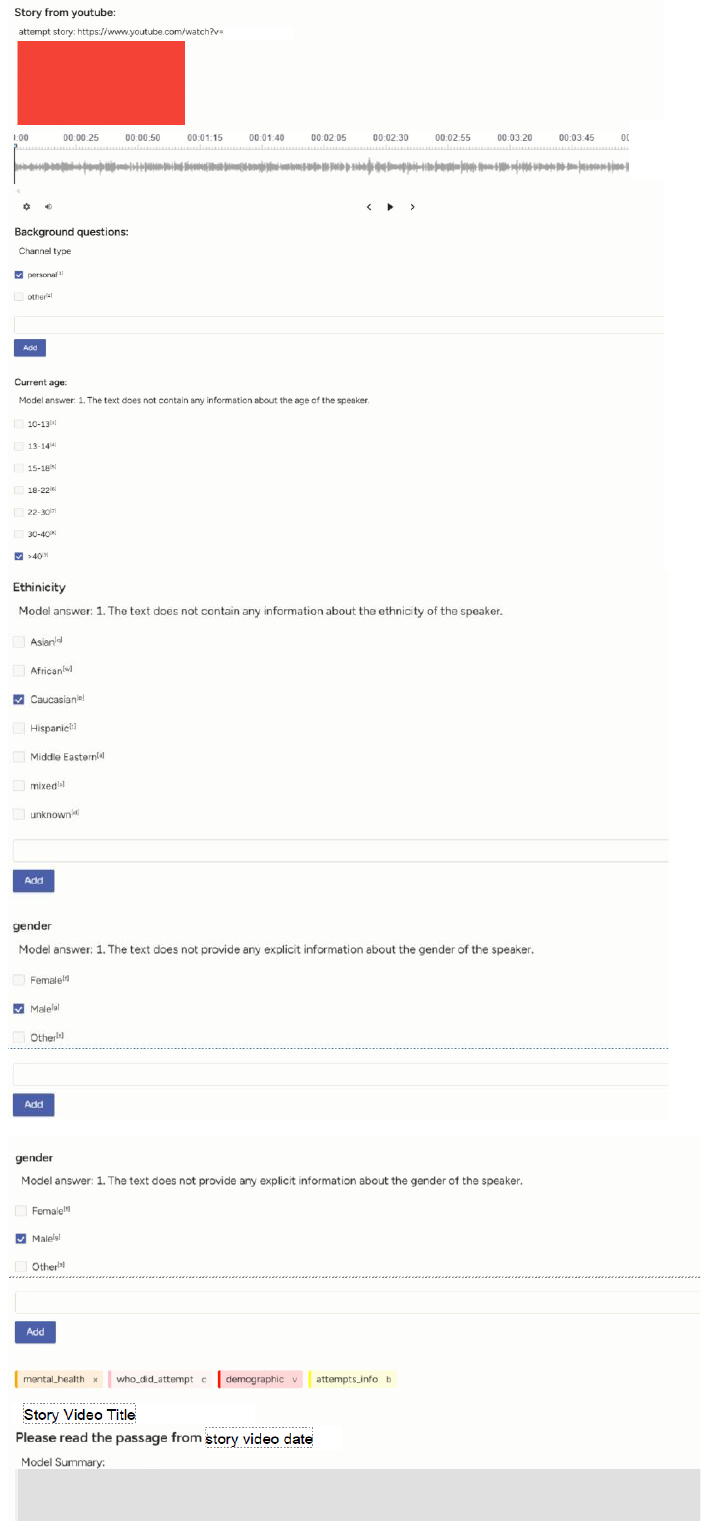}
\caption{The annotation interface from Label-Studio \citep{LabelStudio}, displaying a partial example provided to the annotators. The interface supported two main annotation processes: (1) \textbf{Non-Expert Assessment}, which involved classifying personal channels, estimating demographic information, assessing the relevance of the story, extracting event-related details (e.g., suicide attempt date); and (2) \textbf{Psychological assessment} of suicide attempt narratives (conducted by a clinical psychologist), which included evaluating risk factors, risk groups, mental health illness, and distinguishing suicide attempts from non-suicidal self-harm. Full example of annotation interface is attached in a separate file.}    \label{fig:annotation_interface}
\end{figure}

\subsection{Matching procedure}\label{matching_full}

\begin{figure}[!ht]
    \centering
    \includegraphics[width=0.8\textwidth]{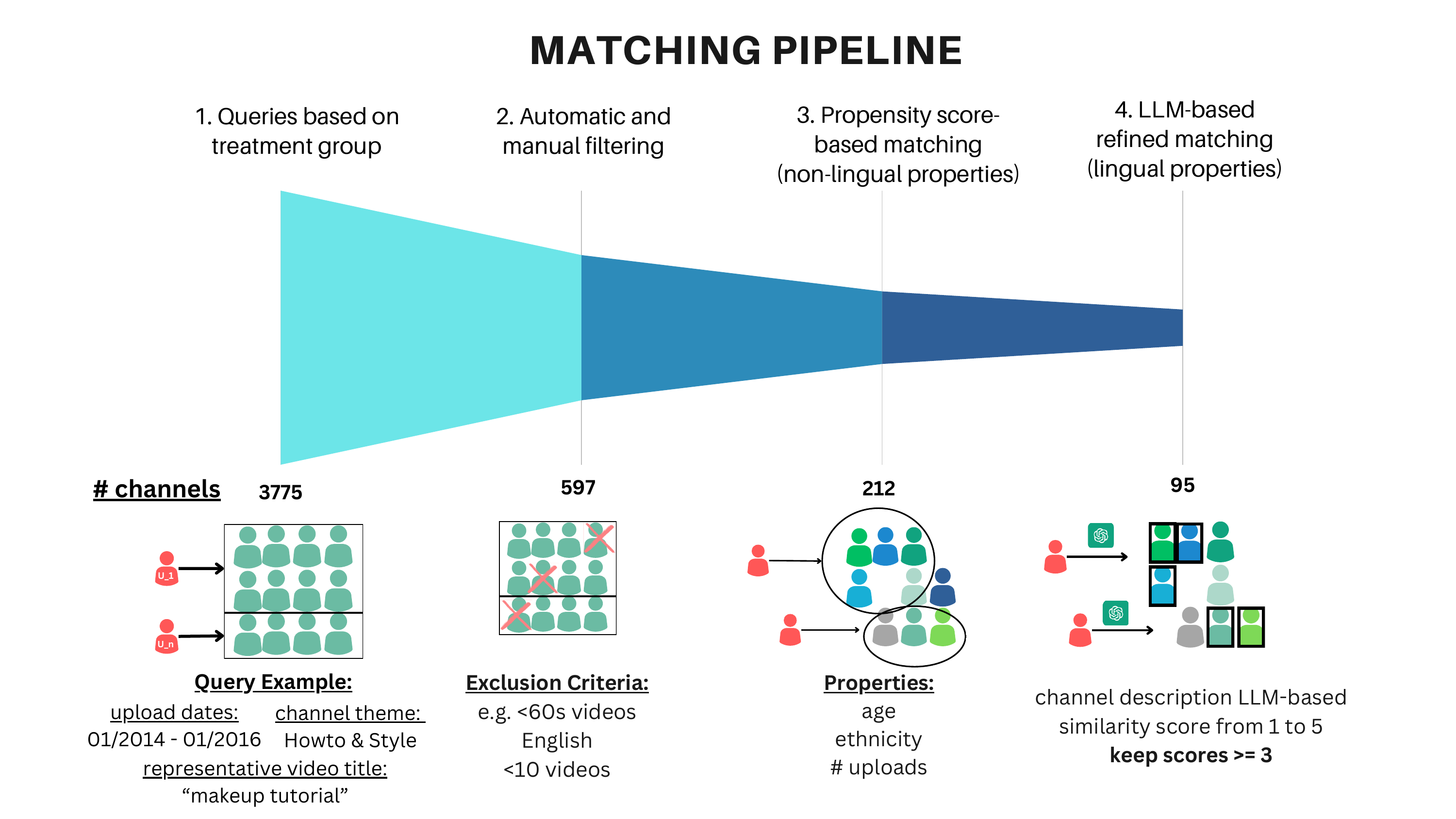}
    \caption{\small 
    The matching algorithm, comprising of four steps: 
    (1) creating \textbf{queries based on the treatment group}, using the YouTube API (\S\ref{matching_full_queries}); 
    (2) \textbf{automatic and manual filtering} resulting in relevant \textbf{personal channels} (\S\ref{matching_full_filtering}); 
    (3) \textbf{propensity score-based matching} based on demographic and channel data, using (a) logistic regression and (b) KNN matching of treatment to control users (\S\ref{matching_full_prop_Estimation}); and 
    (4) \textbf{LLM-based refined matching}, using GPT-4 to score the potential matching, retaining good matches (scores > 3) (\S\ref{matching_full_llm}).
    }
    
    \label{fig:matching_algorithm}
\end{figure}

The matching process ultimately identified 95 users as the Control (Matches) group. This group represents a general cohort of the YouTube population matched by different attributes to the Attempted (During) group. While this group was initially collected and matched to the Attempted group, the analysis in the main paper focuses on group-level effects to estimate the prevalence of different topics.  The full methodology is illustrated in Figure 5, and the sections below provides both a high-level overview and technical details regarding each step (\S{\ref{matching_technical_details}}).

\paragraph{Overview} \phantomsection \label{matching_overview}

Each step in the following section is briefly described, with references to the relevant detailed sections:

(1) We first automatically constructed \textbf{queries based on treatment group} properties (\S\ref{matching_full_queries}), encompassing channel characteristics such as themes and upload period. This process produced 20–30 candidate matching queries per channel, resulting in a final group of 3,775 channels.

(2) This group was then \textbf{filtered manually and automatically} to identify \textbf{matched channels} (\S\ref{matching_full_filtering}) per the inclusion and exclusion criterion detailed in Table \ref{tab:inclusion_exclusion_criteria}, yielding a pool of 597 potential control cases.

We applied a two-step matching algorithm to find an optimal subset of matches:

(3) The first step used \textbf{propensity score-based matching} (\S\ref{matching_full_prop_Estimation}) \citep{caliendo2008some}, which estimates similarity based on numerical and categorical features using a logistic regression model trained on demographic attributes (e.g., age) and channel-related features (e.g., number of uploads). Matching was performed using k-nearest neighbor on the resulting propensity scores.

(4) The second step involved \textbf{LLM-based refined matching} (\S\ref{matching_full_llm}), leveraging GPT-4 \citep{openai2023gpt}. This step incorporated channel content, aiming to control for general themes reflected in the textual content of the channel (e.g., gaming, coding, fashion). Matches were scored on a 1–5 scale based on textual content descriptions, and only channels rated 3 or higher were retained.

This process yielded 95 matched control cases for 42 users who attempted suicide during their upload period, with 1 to 3 matches per treatment channel.

\paragraph{Technical Details} \phantomsection\label{matching_technical_details}

\subsubsection{Queries based on treatment group} \label{matching_full_queries}

The initial pool of potential matches was generated by creating YouTube search queries automatically tailored to each user in the Attempted group. These queries incorporated channel-specific features: frequent bi-grams from video titles, video categories, and upload time frames. This process yielded a total of 3,775 channels.

\subsubsection{Automatic and manual filtering personal channels} \label{matching_full_filtering} 

Then, we used the \textit{yt-dlp} package and downloaded audio files of public videos that were longer than 60 seconds and smaller than 50 MB, and used Whisper (medium) to transcribe the audio. The 3775 channels were later \textbf{automatically} and \textbf{manually} filtered to exclude non-compliant channels and videos (per the inclusion criteria detailed in Table \ref{tab:inclusion_exclusion_criteria}), yielding 597 channels as potential control matches.

\subsubsection{Propensity Score-Based Matching } 
\label{matching_full_prop_Estimation}

 \paragraph {\textbf{Propensity Score Estimation:}  }

Propensity score estimation is used to calculate the likelihood of treatment assignment based on observed covariates. A logistic regression model was employed to estimate the propensity score \(P(T=t|X)\). Here, \(t=1\) represents treatment, Attempted (During), while \(t=0\) denotes the Control (Matches) group. The covariates (\(X\)) consisted of two categories of features:  
\begin{itemize}  
    \item Demographic features: gender and age (manually annotated).  
    \item Channel-related features: video categories, number of uploads and followers, average upload duration, views, likes, and comments (automatically extracted from YouTube).  
\end{itemize}  

 \paragraph {\textbf{Propensity Score Matching:}  }

Based on the estimated propensity scores, k-nearest neighbor matching was applied to identify a set of potential matches for each treatment user. We tested different values of \(k\) (3, 5, and 8) and selected \(k=8\), which demonstrated higher overall ranks in the subsequent LLM-based evaluation stage.

 \subsubsection{LLM-based Refined Matching}
\label{matching_full_llm}

Lastly, we incorporated textual features, channel descriptions, upload titles, and video descriptions, to further refine the matches using GPT-4  \citep{openai2023gpt}. For cost and efficiency, three uploads per channel were preprocessed and summarized using GPT-4.  Then, we used GPT-4 again to evaluate and rate each match on a scale of 1 to 5 (using the prompt described in \ref{prompts}), providing explanations for its choices. Matches rated 3 or higher were retained, as they were deemed to be of sufficient quality for analysis. Matches rated below 3 were excluded to reduce noise and improve the reliability of the results. 
\clearpage

\clearpage
\section{Appendix B - Procedures} \label{appendix_procedures}

\noindent
This appendix provides a detailed description of the computational and statistical procedures that complement the \textbf{procedures} section (~\S3) of the main text. It begins with the topic modeling framework (\S\ref{topic_modeling_full}), followed by temporal alignment (\S\ref{temporal_align_full}) and the associated testing procedures (\S\ref{appendix_temporal_testing}). 

\setcounter{table}{0}
\renewcommand{\thetable}{B.\arabic{table}}
\setcounter{figure}{0}
\renewcommand{\thefigure}{B.\arabic{figure}}

\subsection{Topic modeling} \label{topic_modeling_full}

\begin{figure}[h]
    \centering
    \includegraphics[width=0.95\textwidth]{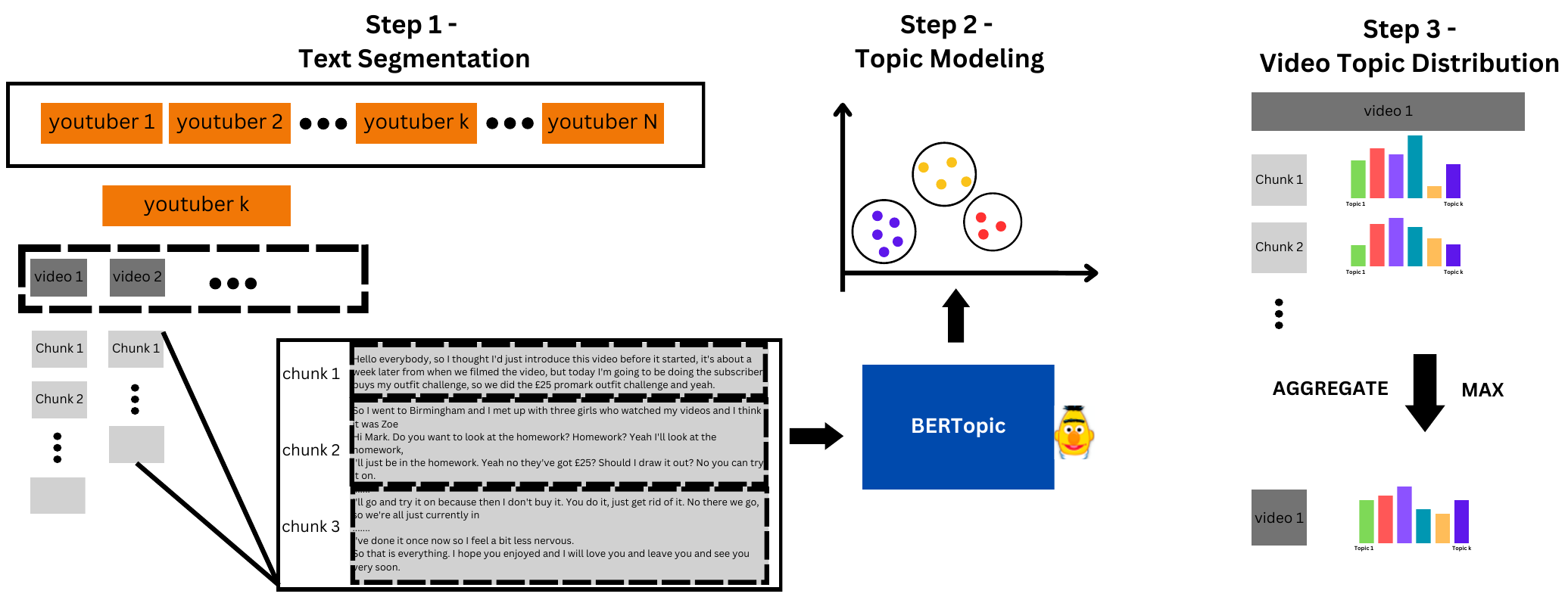}
    \caption{Overview of the Topic Modeling Pipeline. The pipeline includes text segmentation, topic modeling using BERTopic, and video topic distribution to analyze YouTube video content.}
    \label{fig:topic_modeling_pipeline}
\end{figure}
\subsubsection{Technical Pipeline}

The topic modeling pipeline was applied to transcripts from the full-channel videos uploaded by the Attempted (During) and Attempted (Before) groups. Each video was transcribed and preprocessed according to the methods detailed in \S{\ref{data_preprocessing_appendix}}.

First, we obtained embeddings of textual segments using Sentence-BERT (SBERT), a modification of the Bidirectional Encoder Representations from Transformers (BERT), network to efficiently create short text embeddings. Uniform Manifold Approximation Projection (UMAP) \citep{mcinnes2018umap}, was used for dimensionality reduction, reducing the embeddings from $300$ to $5$ dimensions. Next, the embeddings were clustered with the Hierarchical Density-based spatial clustering of applications with noise (HDBSCAN) algorithm \citep{mcinnes2017hdbscan}. HDBSCAN identifies and excludes noisy data points from the clusters, contributing to greater clarity of the topic model. In addition, HDBSCAN provides probability distribution over the topics, as a result of soft clustering (mimicking traditional topic modeling techniques like Latent Dirichlet Allocation (LDA) \citep{blei2001latent}), in which each document is modeled as a distribution of topics. Finally, each cluster is represented by its top $10$ words and $3$ most representative textual segments, which were used as input to an LLM (specifically LLama-7B \citep{touvron2023llama}), and facilitated in naming the clusters (see \ref{topic_prompt} for full prompt used in naming the clusters). 

While fitting different topic models, we explored various hyper-parameters and concluded with the following set of parameters as seen in Table \ref{tab:model_umap_hdbscan_params}.

\begin{table}[h!]
\centering
\caption{Model and Technical Hyperparameters for Topic Modeling, Dimensionality Reduction, and Clustering}
\begin{tabular}{p{5cm}p{8cm}}
\toprule
\multicolumn{2}{l}{\textbf{Topic Modeling Hyperparameters}} \\ 
\midrule
Embedding Model & Sentence-BERT (SBERT)\\
Representation Model & meta-llama/Llama-2-7b-chat-hf \\
Minimum Document Frequency & 0.001 \\
Number of Topics & auto \\
N-gram Range & (1, 2) \\
Minimum Topic Size & 180 \\
\midrule[\heavyrulewidth]
\multicolumn{2}{l}{\textbf{UMAP Parameters}} \\ 
\midrule
Metric & cosine \\
Minimum Distance & 0 \\
Low Memory & true \\
Number of Neighbors & 15 \\
Number of Components & 5 \\
Random State & 33 \\
\midrule[\heavyrulewidth]
\multicolumn{2}{l}{\textbf{HDBSCAN Parameters}} \\ 
\midrule
Metric & euclidean \\
Minimum Samples & 60 \\
Prediction Data & true \\
Minimum Cluster Size & 180 \\
Generate Minimum Spanning Tree & true \\
Cluster Selection Method & eom \\
\bottomrule
\end{tabular}
\label{tab:model_umap_hdbscan_params}
\end{table}

\subsubsection{Prompts for Topic Modeling and Matching} \label{prompts}

This section contains the prompt we used as part of the topic modeling and matching processes.  The first prompt used in the LM-based topic modeling pipeline is described in \ref{topic_prompt}. The LLM (LLama-7B) was provided with the 10 top TF-IDF terms and 3 most representative textual segments, and tasked with generating concise topic labels based on the provided content. The second prompt used in the matching process (\ref{matching_full_llm}), as part of the LLM-refined matching by GPT-4, can be seen at \ref{llm_matching_prompt}.

\label{topic_prompt}
\begin{tcolorbox}[title=Topic Modeling Naming - LLama-7B, colframe=blue, colback=blue!8!white, coltitle=white, fonttitle=\bfseries, width=\textwidth]
<s>[INST] <<SYS>>
You are a helpful, respectful and honest assistant for labeling topics.
<</SYS>>

I have a topic that contains the following documents:

- Traditional diets in most cultures were primarily plant-based with a little meat on top, but with the rise of industrial-style meat production and factory farming, meat has become a staple food.

- Meat, but especially beef, is the word food in terms of emissions.

- Eating meat doesn't make you a bad person, not eating meat doesn't make you a good one.

The topic is described by the following keywords: 'meat, beef, eat, eating, emissions, steak, food, health, processed, chicken'.

Based on the information about the topic above, please create a short label of this topic. 

Make sure you only return the label and nothing more.

[/INST] Environmental impacts of eating meat

[INST]
I have a topic that contains the following documents:
[DOCUMENTS]

The topic is described by the following keywords: '[KEYWORDS]'.

Based on the information about the topic above, please create a short label of this topic. 
Make sure you only return the label and nothing more.
[/INST]
\end{tcolorbox}

\begin{tcolorbox}[title=Matching Prompt - GPT-4o, colframe=blue, colback=blue!8!white, coltitle=white, fonttitle=\bfseries, width=\textwidth]
You will receive information about a treatment YouTuber and <k> potential match YouTubers. For each YouTuber, the following details will be provided:

\begin{enumerate}
    \item \textbf{Channel Description:} A brief overview of the channel’s content and focus.
    \item \textbf{Summary of Video Titles:} A general summary of the types of video titles the channel features.
    \item \textbf{Summary of Video Descriptions:} A synopsis of the typical video descriptions found on the channel.
    \item \textbf{Video Tags:} Common tags used across the videos.
\end{enumerate}

Evaluate the similarity between the treatment YouTuber and each potential match by considering the content, style, and demographic context. Rate each potential match on a scale of 1 to 5, where:

\begin{itemize}
    \item \textbf{1:} Bad match – The channels are very different.
    \item \textbf{2:} Poor match – The channels have some differences.
    \item \textbf{3:} Average match – The channels have some similarities but also notable differences.
    \item \textbf{4:} Good match – The channels are quite similar.
    \item \textbf{5:} Excellent match – The channels are very similar and the individuals seem similar in terms of demographics and context.
\end{itemize}

Provide a brief justification for each rating to explain your reasoning.
\end{tcolorbox}
\label{llm_matching_prompt}

\subsubsection{Topic Modeling Stability and Validation}  \label{topic_modeling_stability}
\begin{table}[h!]
\centering
\begin{tabular}{cccccc}
\hline
Seed & \#Topics & Coherence & Perplexity & Label Similarity & Words Similarity \\
\hline
30 & 64  & 0.42 & 1.40 & - & - \\
32 & 154 & 0.39 & 3.20 & - & - \\
33 & 167 & 0.38 & 4.35 & - & - \\
34 & 179 & 0.49 & 3.16 & - & - \\
35 & 177 & 0.40 & 3.16 & - & - \\
36 & 180 & 0.38 & 4.65 & - & - \\
\hline
Mean & 153 & 0.41 & 3.32 & 0.69 & 0.67 \\
Std  & 41  & 0.04 & 1.11 & 0.03 & 0.03 \\
\hline
\end{tabular}
\caption{Topic modeling metrics across six random seeds. Coherence and perplexity reflect semantic quality, while label and word similarity quantify stability. Similarity metrics were computed between all pairs of models, which is why the table reports mean and standard deviation across the pairs.}
\label{tab:topic_stability}
\end{table}

To assess the robustness of the derived topics, we evaluated both quantitative and qualitative measures. We fitted topic models using six random seeds and compared the resulting topics using coherence, perplexity, and similarity of labels and top words.  

\noindent \textbf{Topic Quality Metrics:}  
\begin{itemize}
    \item \textbf{Coherence:} Semantic consistency of top words within a topic; higher values indicate more interpretable topics.
    \item \textbf{Perplexity:} How well the model predicts unseen documents; lower values indicate better fit to the corpus.
    \item \textbf{Label Similarity:} Consistency of LLM-based topic labels across seeds; higher values reflect stable thematic identification.
    \item \textbf{Words Similarity:} Similarity of top words across seeds; higher values indicate stable topic composition.
\end{itemize}

Table~\ref{tab:model_umap_hdbscan_params} summarizes these metrics.

Overall, coherence and label/word similarity were consistent across seeds, indicating stable topic representations. We selected the model with seed 33 for downstream analyses, as it best reflected the overall topic distribution.  Finally, human validation of the selected model was performed. 

In addition, a clinical psychologist reviewed the topic labels generated by the LLM. Her overall assessment was that several topic names were overly specific to the narratives or contexts from which they were derived. For instance, she suggested simplifying labels such as “Online Therapy” to “Therapy,” “Racial tensions and discrimination within the Black community” to “Black community,” and “Sleep Disorders and Habits” to “Sleep Habits.” Among the 19 topics identified as relevant to suicide risk factors, 4 were considered overly specific. Following this evaluation, we chose to retain the original LLM-assigned topic names throughout the paper for the sake of consistency.

\subsection{Temporal Alignment \label{temporal_align_full}}

\begin{figure}[h]
    \centering
    \includegraphics[width=0.9\textwidth]{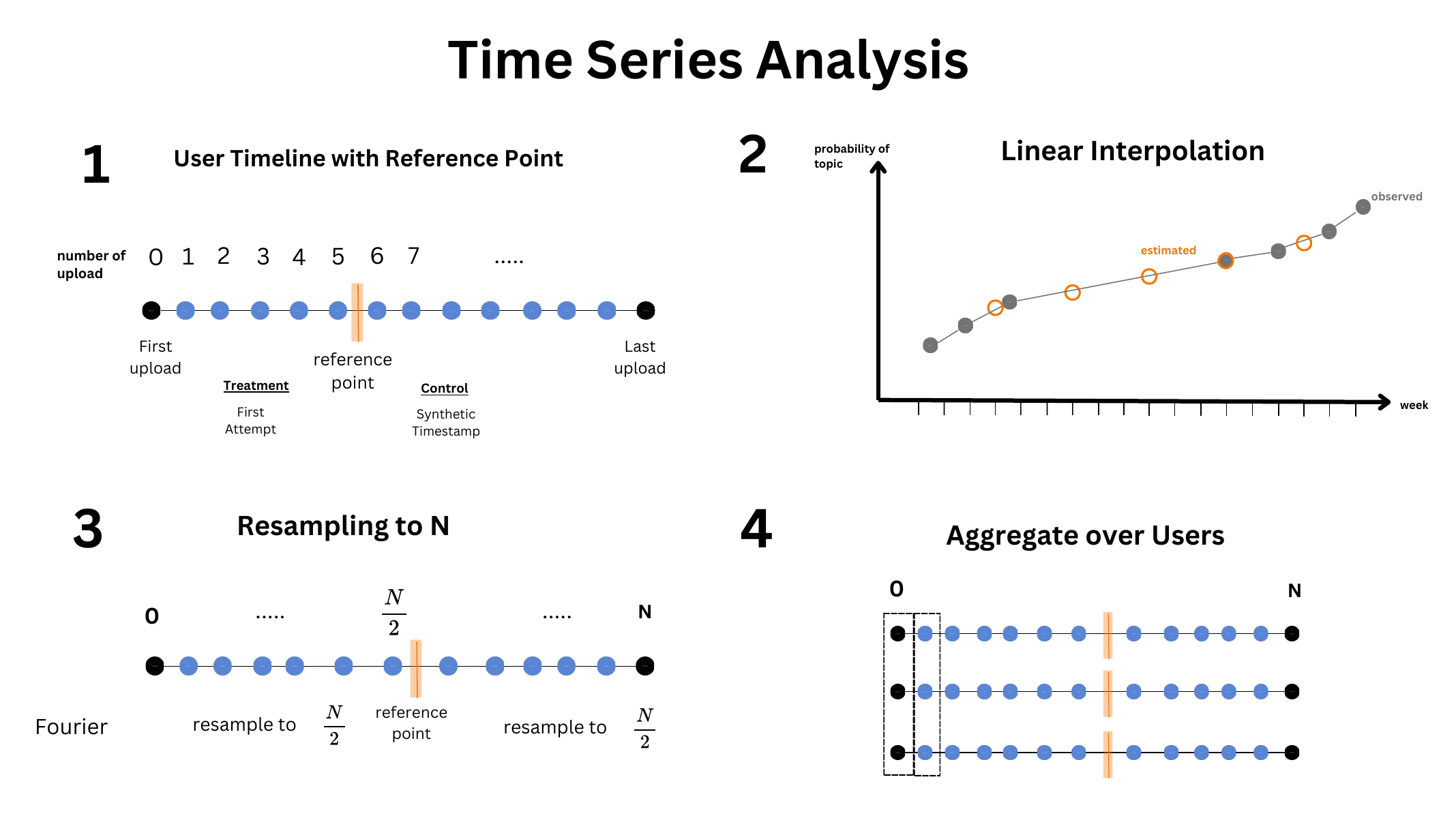}
    \caption{Overview of the Temporal Analysis Pipeline.}
    \label{fig:temporal_analysis_pipeline}
\end{figure}

In Figures 9 and 10, we analyzed the temporal dynamics of the topics \textit{YouTube Engagement} and  \textit{Mental Health Struggles} pre- and post-event. A key challenge in this analysis was aligning uploads from different users to a shared temporal reference frame. This alignment was performed relative to a specific reference point, which varied by group. For the Attempted (During) and Control (Major Life Event) groups, we used the manually extracted \textit{attempt} or \textit{major life} event date, while for the Attempted (Before) and Control (Matches) group, we used the midpoint of upload period. As part of this analysis, we excluded uploads from the attempt/major life event date for all applicable users.  Only videos within an 18-month window before and after the reference event were included to standardize temporal coverage across users.

Once reference points were established, we interpolated uploads to weekly values, applying linear interpolation for each topic probability to generate scores for every calendar week and topic. We then standardized the data by up-sampling and down-sampling to obtain 15 samples pre- and post-reference point (total 30). Finally, we aggregated these values across all users, calculating the mean and standard error for each normalized timestamp. This comprehensive approach enabled consistent temporal analysis across diverse user groups, facilitating comparative studies despite varying reference points and upload frequencies.

\subsection{Temporal Testing Procedure} \label{appendix_temporal_testing}

In Figures 9 and 10, we performed two types of statistical analyses to compare topic trends:  

\paragraph{Procedure 1: Within-Group Comparison}  
This procedure compared time steps within the Attempted (During) group to the baseline at \( t = 0 \):  
\begin{itemize}
    \item \textbf{Aligned Values}: The aligned values of all users in the Attempted (During) group were utilized, with \( t = 0 \) serving as the baseline.  
    \item \textbf{Statistical Test}: Paired \( t \)-tests were conducted to compare the aligned values at each time step (\( t = 1, \ldots, 30 \)) to the baseline values at \( t = 0 \).  
    \item \textbf{Post-Hoc Correction}: Benjamini-Hochberg (BH) FDR correction was applied to account for multiple comparisons across the 30 time steps.  
\end{itemize}  

\paragraph{Procedure 2: Between-Group Comparison}  
This procedure compared the Attempted (During) group to each of the three control groups (Attempted (Before), Control (Matches), Control (Major Life Event)) at each time step:  
\begin{itemize}
    \item \textbf{Aligned Values}: The aligned values of users in the Attempted (During) group were compared to the aligned values of users in the control groups at each time step (\( t = 1, \ldots, 30 \)).  
    \item \textbf{Statistical Test}: Welch’s \( t \)-test, which accounts for unequal variances, was used due to differing sample sizes between groups.  
    \item \textbf{Post-Hoc Correction}: Benjamini-Hochberg (BH) FDR correction was applied to control for multiple hypothesis testing across 30 time steps and 3 control groups (a total of 90 comparisons).  
\end{itemize}  

Both procedures relied on aligned values, as described in \S{\ref{temporal_align_full}}, to ensure comparisons were conducted in a consistent timeline framework.

\clearpage
\section{Appendix C - Supplementary Results}
\label{appendix_supp_results}
This appendix presents supplementary figures, tables, and analyses that support the main results section~\S4 of the manuscript. It expands on the topic modeling results (\S\ref{results_topic_modeling}), complementary analyses of narrative suicide videos (\S\ref{topic_model_attempts}), and provides supplementary figures and tables (\S\ref{statisitcal_analysis_full}). 
\setcounter{table}{0}
\renewcommand{\thetable}{C.\arabic{table}}
\setcounter{figure}{0}
\renewcommand{\thefigure}{C.\arabic{figure}}

\subsection{Topic Modeling of YouTube Channels} \label{results_topic_modeling}

\begin{figure}[h]
    \centering
    \includegraphics[width=0.95\linewidth]{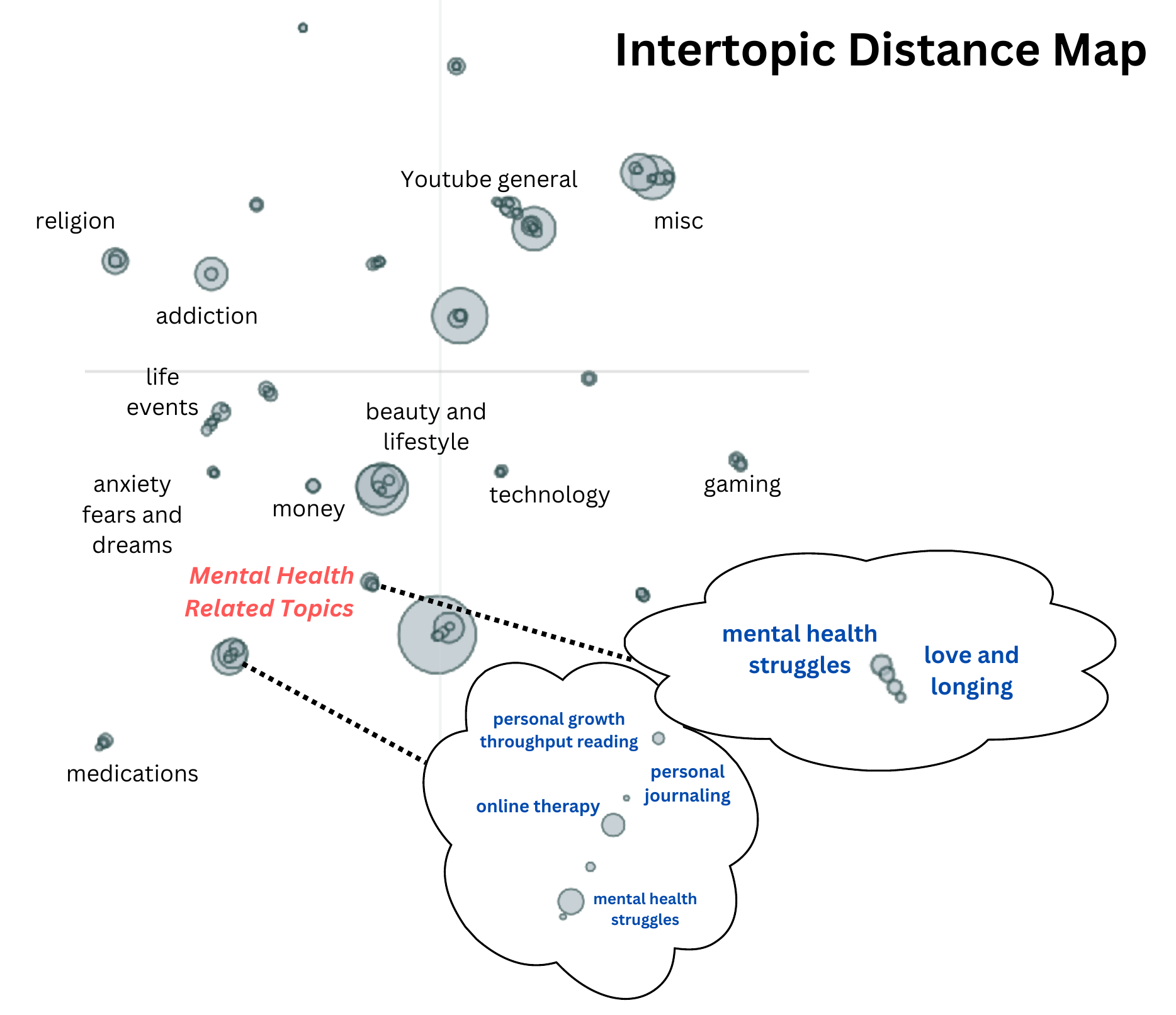}
    \caption{\small Inter-topic Distance Map showing the results of topic modeling trained on the \textit{Attempted (During + Before)} group. The figure displays the 100 most frequent topics, with bubble size indicating topic frequency and location determined by the average word embeddings of each topic after dimensionality reduction to 2D using UMAP algorithm. The view is zoomed in to highlight Mental Health topic clusters, with their LLM-generated names visible in blue. The rest of the cluster names were manually assigned for readability.}
    \label{fig:topic_modeling_all}
\end{figure}

Our analysis leveraged an LM-based topic modeling approach to analyze the transcripts from Attempted (Before) and Attempted (During) channels. This analysis resulted in the clustering of these transcripts and the identification of 166 topics, which were automatically named using the Llama-2-7B model \citep{touvron2023llama} (refer to \ref{prompts} for the full prompt used). The spatial distribution of the topics' embedding vectors is illustrated in Figure \ref{fig:topic_modeling_all}.

Our quantitative analysis of topic quality found that topics with similar names, such as \textit{Mental Health Struggles} and \textit{Mental Health Awareness}, were semantically related. This relationship is also reflected in their high cosine similarity \citep{salton1988term}, with a score of 0.72, indicating that topics with similar LLM-generated names also tend to have similar underlying content. 

Additionally, certain topics were user-specific, appearing in less than 10\% of the users (e.g., \textit{Filipino Culture and Language}), while others were more prevalent, appearing in more than 70\% of the users' videos (e.g., \textit{YouTube Engagement}). A detailed description of the topics, including their names, frequencies, most representative textual segments, and correlations between topics, is provided in the Supplementary Materials Appendix \ref{topic_information}.

\subsection{Topic Modeling of Narrative Suicide Stories} \label{topic_model_attempts}

\begin{figure}[h]
    \centering
    \includegraphics[width=0.9\textwidth]{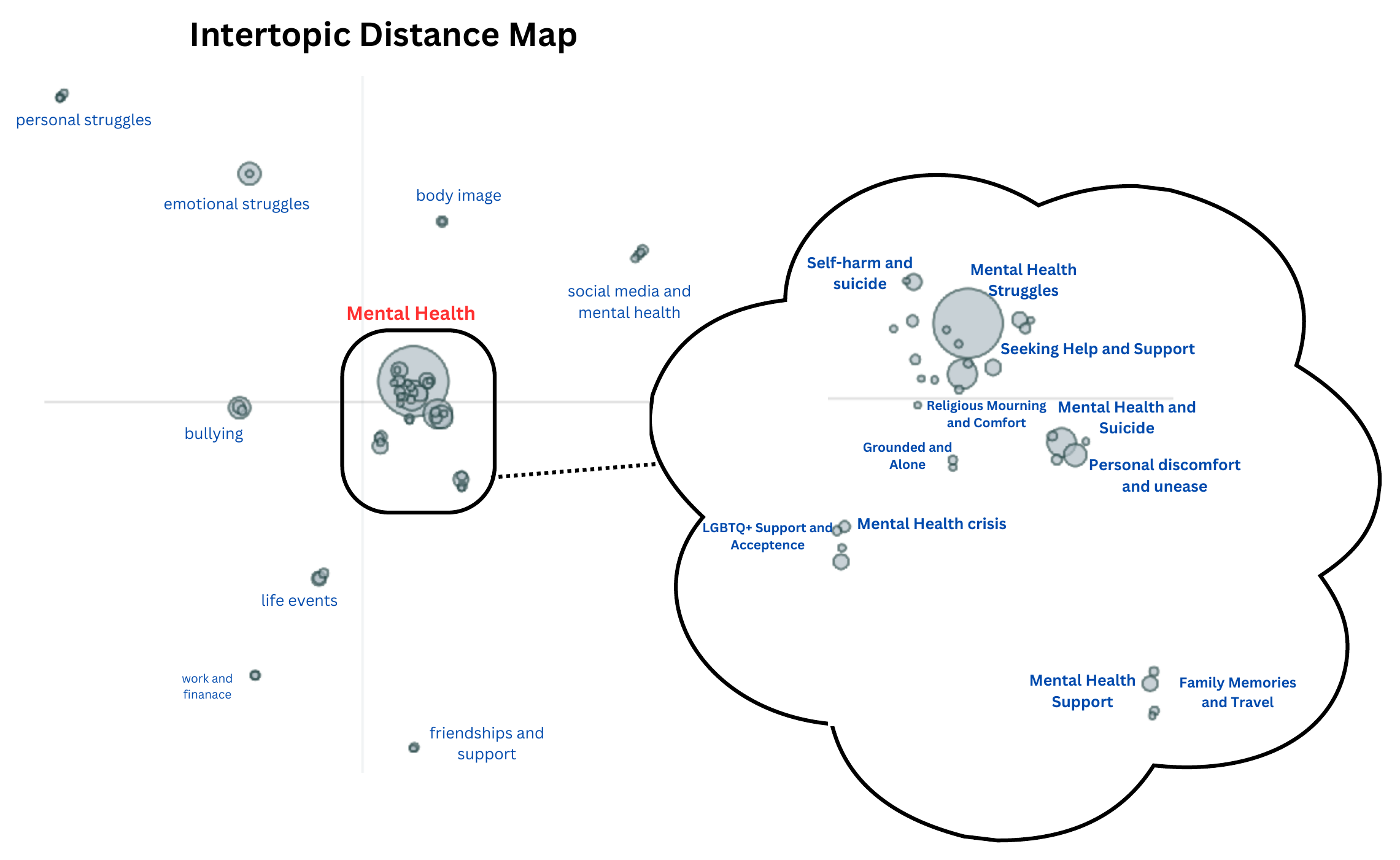}
    \caption{\textbf{Attempt Stories Inter-topic Distance Map} showing the results of topic modeling trained on the \textbf\textbf{attempt stories}. The map visualizes clusters of different topics, named by LLM, where the size of each topic bubble indicates the number of document chunks classified under that topic. The figure also includes zoomed-in views of a cluster of topics centered around mental health experience (e.g., Mental Health Support, Seeking Help and Support, Religious Mourning and Comfort)}
    \label{fig:topic_modelig_attempts}
\end{figure}

The primary topic model analyzed in this study (\ref{results_topic_modeling}) was trained on data from the treatment group (Appendix \S{\ref{data_preprocessing_appendix}} describes data collection for these groups), including all uploads from both the Attempted (Before) and Attempted (During) groups (including the attempt narrative videos).

As a complementary analysis to the psychological assessment (ref), we trained a separate topic model specifically on the 181 \textit{attempt story} videos reviewed by the psychologist. We applied a sliding-window text segmentation approach, advancing one sentence at a time, to generate 32,347 overlapping chunks. This method allowed for a more granular representation of the narratives, supporting richer topic detection.

The clustering process yielded 56 distinct topics (Figure \ref{fig:topic_modelig_attempts}), covering themes such as finances, family, personal reflections, friendships, emotional support, crisis management, and resilience. A significant cluster focused on individuals' mental health journeys, including experiences with self-harm, spirituality, and identity exploration. The clustering results provide a quantitative basis for exploring the interplay between different aspects of individuals' experiences in the context of suicide attempts. 

Future work could explore whether these emergent topics—particularly those not captured in standard clinical assessments—hold predictive value for assessing suicide risk or guiding intervention design.
\clearpage

\subsection{Supplementary Figures and Tables} \label{statisitcal_analysis_full}

This section presents supplementary analyses intended primarily for illustration and transparency. It expands on the main results of the manuscript by providing additional figures and visual examples that help contextualize the findings. 

\begin{figure}[h]
    \centering
    \includegraphics[width=0.95\textwidth]{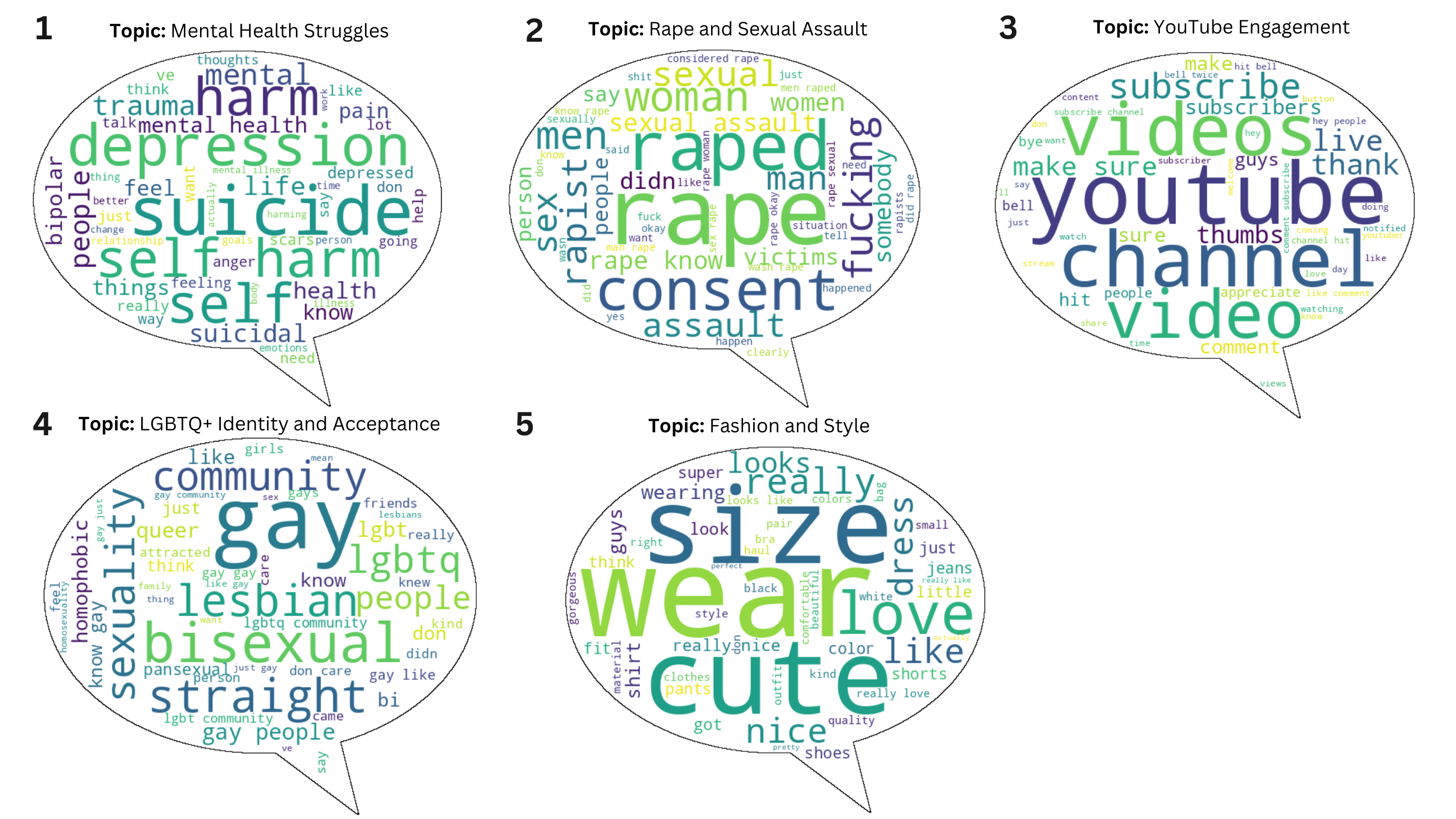}
  \caption{Word clouds based on the TF-IDF scores of the words in the five topics identified as predictive of the period prior to a suicide attempt (\S{3.1}). The size of the words in each cloud is proportional to their TF-IDF score.} \label{full_tf_idf_wordclouds}
   
\end{figure}

\begin{figure}[h]
    \centering
    \includegraphics[width=0.9\textwidth]{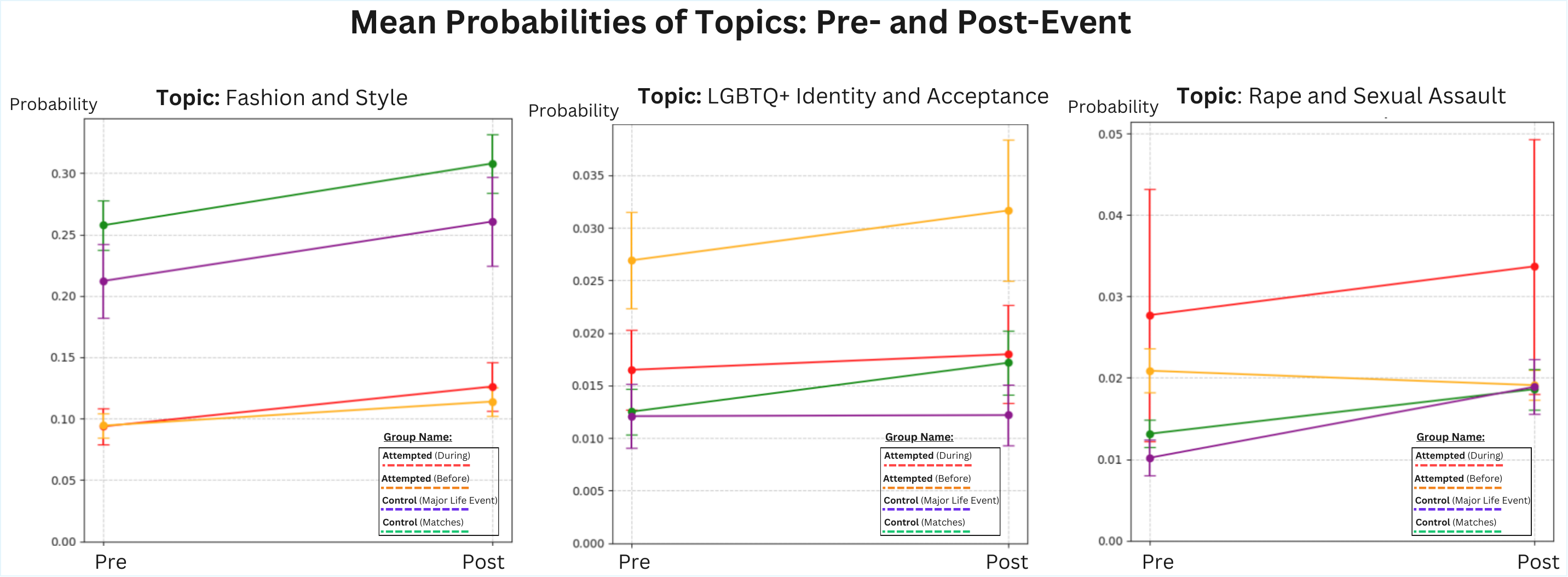}

\caption{\textbf{Fashion and Style (left), LGBTQ+ Identity and Acceptance (middle), and Rape and Sexual Assault (right): }Average Comparison Across Groups. The figures present the average topic values across four groups comparing Pre- and Post-Event (suicide attempt, major life event, or synthetic condition). These results
highlight the observed trends of lower engagement with \textit{Fashion and Style} topics in the Attempted groups and higher engagement with \textit{LGBTQ+ Identity and Acceptance} topics in the Attempted (Before) group. These topics were identified in answering: \textit{What topics may serve as indicators of the period before a suicide attempt? } in \S{3.1.1}. } \label{fashion_and_lgbtq_means}   
\end{figure}

The \textbf{bottom-up} analysis identified five topics as pre-attempt indicators, illustrated in Figure~\ref{full_tf_idf_wordclouds}. Figure~\ref{fashion_and_lgbtq_means} presents the mean aggregated values of selected topics before and after the reference event, showing that the \textit{Fashion and Style} topic distinguishes the two Attempted groups, with lower prevalence compared to the Controls. 

The \textbf{top-down} domain-expert analysis, summarized in Table~\ref{psychological_stats_table}, outlines the psychological profiles derived from narrative assessments of suicide attempts. 

\begin{figure}[ht]
    \centering
    \includegraphics[width=0.85\textwidth]{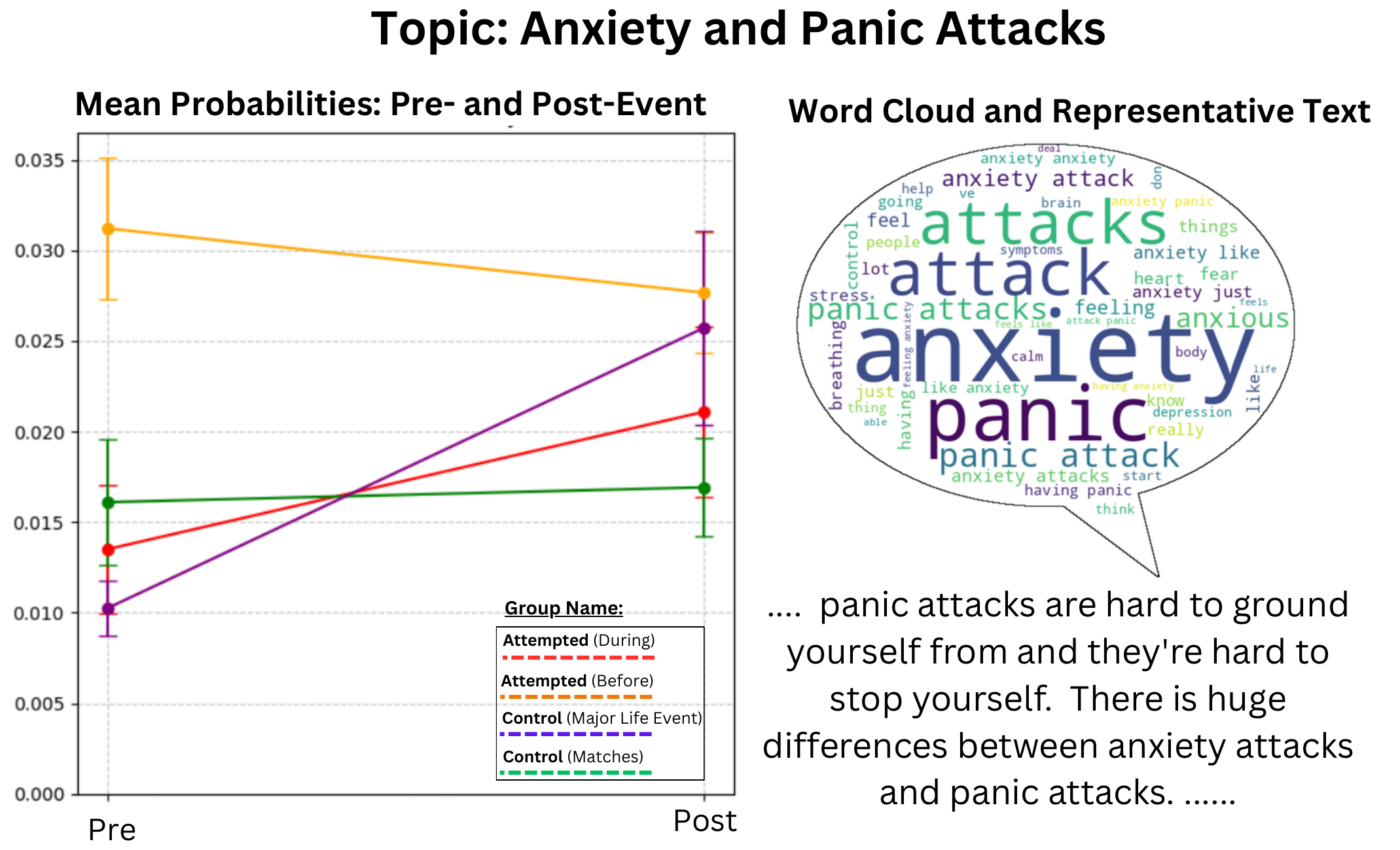}
\caption{\small \textbf{Anxiety and Panic Attacks}. The \textbf{left} figure presents the average topic values across our four groups, comparing Pre- and Post-Event (suicide attempt, major life event, or synthetic condition). The \textbf{right} figure presents the topic's word cloud. }

\label{fig:anxiety_means_word_cloud}

\end{figure}

The \textbf{hybrid} approach integrates computational and expert perspectives, with visualizations shown in Figure~\ref{fig:anxiety_means_word_cloud}. 

\begin{figure}[ht]
    \centering
    \includegraphics[width=\textwidth]{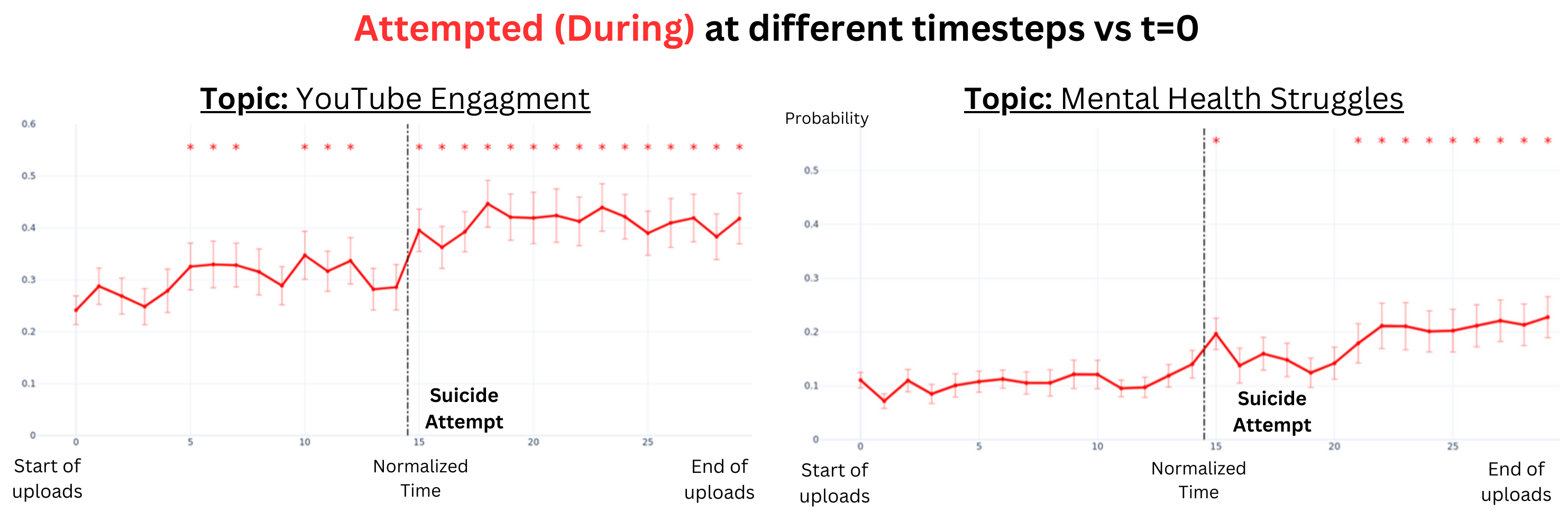}

\caption{\small \textbf{Longitudinal Comparison of YouTube Engagement (left) and Mental Health Struggles (right) relative to $t=0$ in the Attempted (During) Group}. The figures depict temporal trends of these topics within the Attempted (During) group, with the timeline aligned to the suicide attempt. 
At each time step, comparisons were made to $t=0$ using the paired \textit{t}-test. Asterisks ($*$) indicate statistically significant differences. The full statistical procedure is detailed in \S{\textbf{B.3}}.}
   \label{fig:temporal_topics_within}

\end{figure}

Finally, Figure~\ref{fig:temporal_topics_within} presents a \textbf{longitudinal} view of the topics \textit{Mental Health Struggles} and \textit{YouTube Engagement}, demonstrating their temporal variation (\textit{within-group}), which illustrates how the prevalence of the topics \textit{YouTube Engagement} and \textit{Mental Health Struggles} significantly increased after the attempt. 

\clearpage

\begin{table}[ht!]
\scriptsize
\begin{tabular}{>{\raggedright\arraybackslash}p{0.15\linewidth}>{\raggedright\arraybackslash}p{0.35\linewidth}>{\raggedright\arraybackslash}p{0.2\linewidth}>{\raggedright\arraybackslash}p{0.2\linewidth}}
\textbf{Category}                    & \textbf{Sub-Category}                             & \textbf{Attempted (During) (\%, n=42)}& \textbf{Attempted (Before) (\%, n=139)}\\ \hline
\textit{\textbf{Risk Groups}}        &                                                   &                                  &                                  \\ \cline{1-1}
                                     & Teens                                             & 38.1                             & 47.5                             \\
                                     & Immigrants                                        & 33.3                             & 27.3                             \\
                                     & Men                                               & 26.2                             & 26.6                             \\
                                     & LGBTQ+                                            & 11.9                             & 13.7                             \\
                                     & Suicide Survivor Family                           & 0                                & 2.9                              \\
\textit{\textbf{Mental Illness}}     &                                                   &                                  &                                  \\ \cline{1-1}
                                     & Depression                                        & 78.6                             & 79.9                             \\
                                     & Anxiety                                           & 31                               & 30.2                                                 \\
                                     & Eating Disorder                                   & 14.3                             & 5.8                              \\
                                     & PTSD                                              & 2.4                              & 8.6                              \\
                                     & Substance Abuse                                   & 4.8                              & 5.8                              \\
\textit{\textbf{Video Motivation}}   &                                                   &                                  &                                  \\ \cline{1-1}
                                     & Desire to Help Others& 38.1 \textcolor{red}{*}                             & 77     \textcolor{red}{*}                          \\
                                     & Personal Recovery Process & 66.7 \textcolor{red}{*}                             & 40.3  \textcolor{red}{*}                           \\
                                     & Raising Awareness                                 & 21.4                             & 20.1                             \\
                                     & Seeking Help                                      & 2.4                              & 1.4                              \\
\textit{\textbf{Resilience Factors}} &                                                   &                                  &                                  \\ \cline{1-1}
                                     & Social/Family Support                             & 40.5                             & 37.4                             \\
                                     & Responsibility for others                         & 14.3                             & 16.5                             \\
                                     & Treatment                                         & 14.3                             & 15.8                             \\
                                     & Fear of Death                                     & 2.4                              & 11.5                             \\
                                     & Self Disclosure Ability                           & 9.5                              & 7.9                              \\
\textit{\textbf{Risk Factors}}       &                                                   &                                  &                                  \\ \cline{1-1}
                                     & Hopelessness/Despair                              & 71.4                             & 64                               \\
                                     & Loneliness                                        & 47.6                             & 59                               \\
                                     & Unbearable Mental Pain                            & 47.6                             & 59                               \\
                                     & Life Events                                       & 45.2                             & 43.9                             \\
                                     & Comorbidity                                       & 19                               & 25.2                             \\
                                     & Impulsivity and Aggression                        & 26.2                             & 20.1      \\
                                     & Self-Harm                                         & 21.4                             & 18.7                                \\
                                     
                                     & Bullying                                          & 9.5                              & 23                               \\
                                     & Access to Lethal Means                            & 26.2                             & 17.3                             \\
                                     & Self  Disclosure Difficulties                     & 19                               & 18                               \\
                                     & Lack of Belonging                                 & 19                               & 18                               \\
                                     & Abuse                                             & 16.7                             & 16.5                             \\
                                     & Cognitive Distortions                             & 4.8                              & 5.8                              \\
\textit{\textbf{Suicide Attempt Validation}}&                                                   &                                  &                                  \\ \cline{1-1}
                                     & Suicide Attempt                                   & 100                              & 86.3                             \\
                                     & Suicidal Ideation& 54.8                             & 70.5                             \\
                                     & Non-Suicidal Self-Harm                            & 0                                & 1.4                             
                                    
\end{tabular}

 \caption{\small\textbf{Psychological Factors Analysis} conducted by a clinical psychologist through narrative analysis of attempt stories. The assessment comprised six sections: Risk Groups, Mental Illness, Video Motivation, Resilience Factors, Risk Factors, and Suicide Attempt Validation. The table displays the percentage of individuals in each group, with each condition determined by the psychologist based on the content described in the attempt stories. The data indicates "similar confounding" factors across most features, except for the motivations for posting the video, marked with \textcolor{red}{*}, and validated as significant. The table includes only factors that could be extracted from a sufficient number of users (>5\%). See \S{\ref{human_eval_protocol_full}} for the assessment protocol and \S{\ref{topic_model_attempts}}, which describes an {\color{blue}embedding-based with LLM support} topic model trained on the narrative stories.}
\label{psychological_stats_table}
\end{table}

\clearpage
\section{Appendix D - Statistical and Robustness Analyses}
This appendix presents the full statistical analyses supporting the findings described in the results sections and Appendix~\S\ref{appendix_supp_results} (Supplementary Results). It includes mathematical formulation of the GLMM model (\S\ref{appendix_effect_analysis}), and a series of robustness and validation analyses (\S\ref{validation_analysis}) that examine potential confounding factors, engagement metrics, and upload activity patterns across groups. Together, these results quantify and validate the effects reported, ensuring that the observed behavioral and linguistic differences are statistically reliable and methodologically sound.

\setcounter{table}{0}
\renewcommand{\thetable}{D.\arabic{table}}
\setcounter{figure}{0}
\renewcommand{\thefigure}{D.\arabic{figure}}

\subsection{Effects Analysis} \label{appendix_effect_analysis}

As part of both the bottom-up and hybrid approaches, we estimated a generalized linear mixed model (GLMM) using a Beta likelihood and a logit link function to model the bounded response $y_{ij} \in (0,1)$, corresponding to the mean topic probability for user~$i$ at time $j$. We calculated the aggregated mean topic probabilities before and after the reference event. 
The model included fixed effects for \textit{Time} (0 = Before, 1 = After), \textit{Group} (Attempted (During), Attempted (Before), Control (Matches), Control (Major Life Event)), 
and their interaction (\textit{Group} $\times$ \textit{Time}), together with continuous and demographic covariates (Age, Gender, Ethnicity, Upload activity). Random intercepts were specified at the user level to account for repeated measures. The \textit{Attempted (During)} group and the pre‐event period (\textit{Before}) served as reference categories. Thus, each coefficient represents the change in logit‐transformed topic prevalence relative to this baseline. Group effects were parameterized as contrasts against the \textit{Attempted (During)} group:

\begin{equation}
\text{logit}\big(E[y_{ij}]\big) 
= \beta_0 
+ \beta_1 \,\text{Time}_{ij} 
+ \sum_{g=1}^{G-1} \beta_{2g}\,\text{Group}_{g,i}
+ \sum_{g=1}^{G-1} \beta_{3g}\,(\text{Group}_{g,i} \times \text{Time}_{ij})
+ \boldsymbol{\beta_4}^\top \mathbf{X}_{ij}
+ b_{0i},
\label{eq:glmm}
\end{equation}

\noindent
where $\mathbf{X}_{ij}$ represents the matrix of user‐level covariates, and $b_{0i}$ denotes user‐specific random intercepts, assumed to follow $b_{0i} \sim \mathcal{N}(0, \sigma_b^2)$.

\subsubsection{Temporal Effects: Mental Health Struggles and YouTube Engagement }
For the \textit{Mental Health Struggles} topic, the \textit{Attempted (During)} group demonstrated a significant increase in topic prevalence following the event ($\beta=0.55$, $z=3.82$, $p_{adj}<.001$), corresponding to a 74\% increase in the odds of mental health content post-event. The \textit{Attempted (Before)} group showed a significantly smaller temporal increase relative to the \textit{Attempted (During)} group ($\beta=-0.44$, $z=-2.72$, $p_{adj}=.016$), indicating that their post-event change was attenuated compared to the reference group. The control groups also showed patterns of smaller temporal increases (\textit{Control (Matches)}: $\beta=-0.36$, $z=-2.08$, $p=.038$; \textit{Control (Major Life Event)}: $\beta=-0.40$, $z=-1.99$, $p=.046$)

The \textit{YouTube Engagement} topic exhibited a significant post-event increase in the \textit{Attempted (During)} group ($\beta=0.52$, $z=4.80$, $p_{adj}<.001$), reflecting a 67\% increase in the odds of YouTube engagement content. All comparison groups demonstrated significantly smaller temporal increases relative to the \textit{Attempted (During)} group. The \textit{Attempted (Before)} group showed a significantly attenuated post-event increase ($\beta=-0.35$, $z=-2.90$, $p_{adj}=.016$). The \textit{Control (Matches)} group demonstrated the most pronounced differential effect ($\beta=-0.53$, $z=-4.13$, $p_{adj}<.001$), with substantially less temporal change than the reference group. The \textit{Control (Major Life Event)} group similarly exhibited a significantly smaller post-event increase ($\beta=-0.42$, $z=-2.78$, $p_{adj}=.027$) compared to the \textit{Attempted (During)} group.

\subsubsection{Baseline Group Differences: YouTube Engagement and Fashion and Style }

Independent of temporal changes, baseline differences in \textit{YouTube Engagement} were observed across groups. The \textit{Control (Matches)} group ($\beta=0.83$, $z=6.05$, $p_{adj}<.001$) and \textit{Control (Major Life Event)} group ($\beta=0.76$, $z=3.75$, $p_{adj}=.001$) both exhibited markedly higher baseline prevalence compared to the \textit{Attempted (During)} group, corresponding to 129\% and 115\% higher odds, respectively. Male users were associated with significantly lower baseline engagement ($\beta=-0.31$, $z=-2.84$, $p_{adj}=.023$), corresponding to 27\% lower odds relative to female users.

The \textit{Fashion and Style} topic showed significant group and gender differences at baseline, with no significant temporal changes detected. The \textit{Control (Matches)} group demonstrated higher prevalence of fashion-related content compared to the \textit{Attempted (During)} group ($\beta=0.89$, $z=2.86$, $p_{adj}=.01$), representing 144\% higher odds. Male users showed significantly lower engagement with this content ($\beta=-0.66$, $z=-2.55$, $p_{adj}=.026$), with 48\% lower odds compared to female users.

\subsection{Robustness and Validation Analyses}
\label{validation_analysis}
To evaluate whether the observed pre-attempt dip and post-attempt rise in engagement could be explained by non-behavioral factors, we conducted a series of robustness checks addressing potential factors. Specifically, we examined (1) sensitivity analysis to ambiguous cases; (2) platform-level shifts that might reflect external events such as the COVID-19 pandemic; (3) group-level differences in audience engagement or channel characteristics; and (4) upload-frequency biases at the individual channel level. 

\subsubsection{Sensitivity Analysis}

To assess the robustness of our findings to ambiguous cases, we performed a sensitivity analysis excluding 14 users (out of 139 Attempted (Before); $13.8\%$) who posted content describing a suicide attempt but were clinically evaluated as having only suicidal thoughts (see Table \ref{psychological_stats_table}). Notably, this exclusion removed 6 male users from the sample. After excluding these users, all main effects and temporal patterns remained consistent with the primary analysis (Table 3; GLMM results), except for \textit{Group} difference to Attempted (Before), which were no longer significance in any topic ($p_{adj}>0.05$). These results indicate that our primary conclusions regarding topic prevalence and group × time interactions are robust to this source of ambiguity, while the Group difference to specific groups should be interpreted with caution due to its sensitivity to sample composition.

\subsubsection{Control Measures Analysis (Platform-Level Effects)} 

To ensure that the temporal patterns identified around the suicide attempt were not artifacts of global platform dynamics, we examined topic trends using an external, time-anchored reference event unrelated to individual users—the onset of the COVID-19 pandemic in the United States (March~1,~2020). The same temporal alignment and analysis procedures as in ~\ref{temporal_align_full} were applied to all four research groups.

\label{control_measures_analysis}

\begin{figure}[h]
    \centering
    \includegraphics[width=0.85\textwidth]{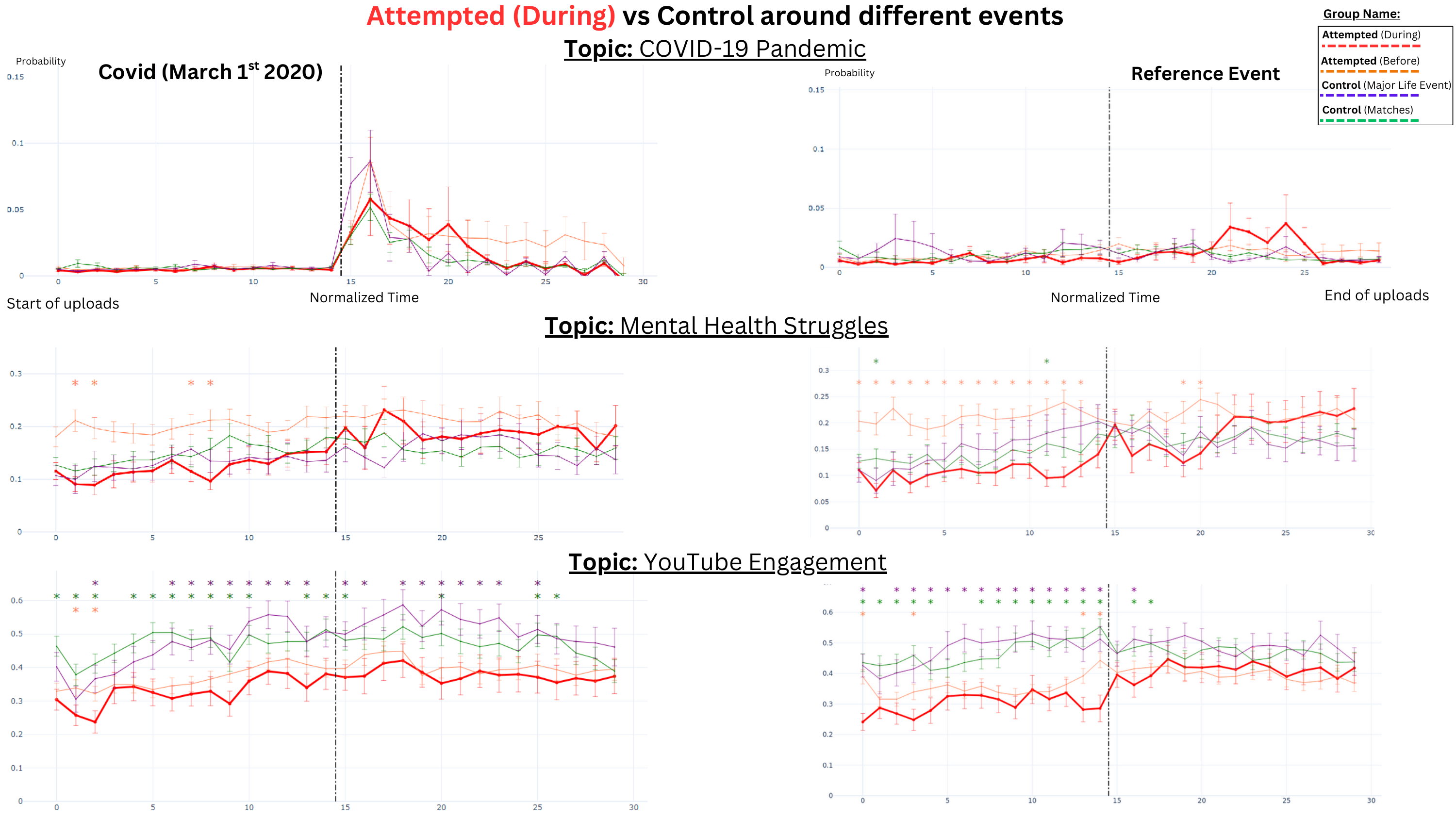}
\caption{
Temporal dynamics of topic prevalence around two reference points: the onset of the COVID-19 pandemic in the United States (March~1,~2020; \textbf{left}) and the individual-level reference events (suicide attempt, major life event, or midpoint of upload; \textbf{right}). 
Each panel displays three topics—\textit{COVID-19 Pandemic} (top) and \textit{Mental Health Struggles} (middle), \textit{YouTube Engagement} —illustrating that the global COVID-19 event elicited parallel changes across all groups, whereas the attempt-related dynamics exhibit distinct within-group patterns.}

    \label{fig:covid_plots}
\end{figure}

As shown in Figure~\ref{fig:covid_plots}, both the treatment and control groups exhibited an expected increase in the topic \textit{COVID-19 Pandemic} around March~2020, consistent with a global shift in YouTube content during that period. Crucially, there were no significant between-group differences in this topic, indicating that all groups were equally affected by this exogenous event. Moreover, while a gradual increase was observed in the topic \textit{Mental Health Struggles} during the pandemic, its temporal dynamics differed markedly from those observed around the suicide attempt event.

 \subsubsection{Engagement Metrics Analysis (Audience and Channel Composition Effects)} \label{appendix_engagment_metrics}

Next, we tested whether the engagement pattern could be explained by audience composition or content popularity rather than behavioral change. To this end, we compared aggregate engagement metrics across all four research groups, including the number of \textbf{uploads}, \textbf{likes}, \textbf{comments}, and average \textbf{video duration}. These measures were analyzed both cross-sectionally and longitudinally relative to each group’s reference event.

\begin{figure}[ht!]
    \centering
    \includegraphics[width=0.85\textwidth]{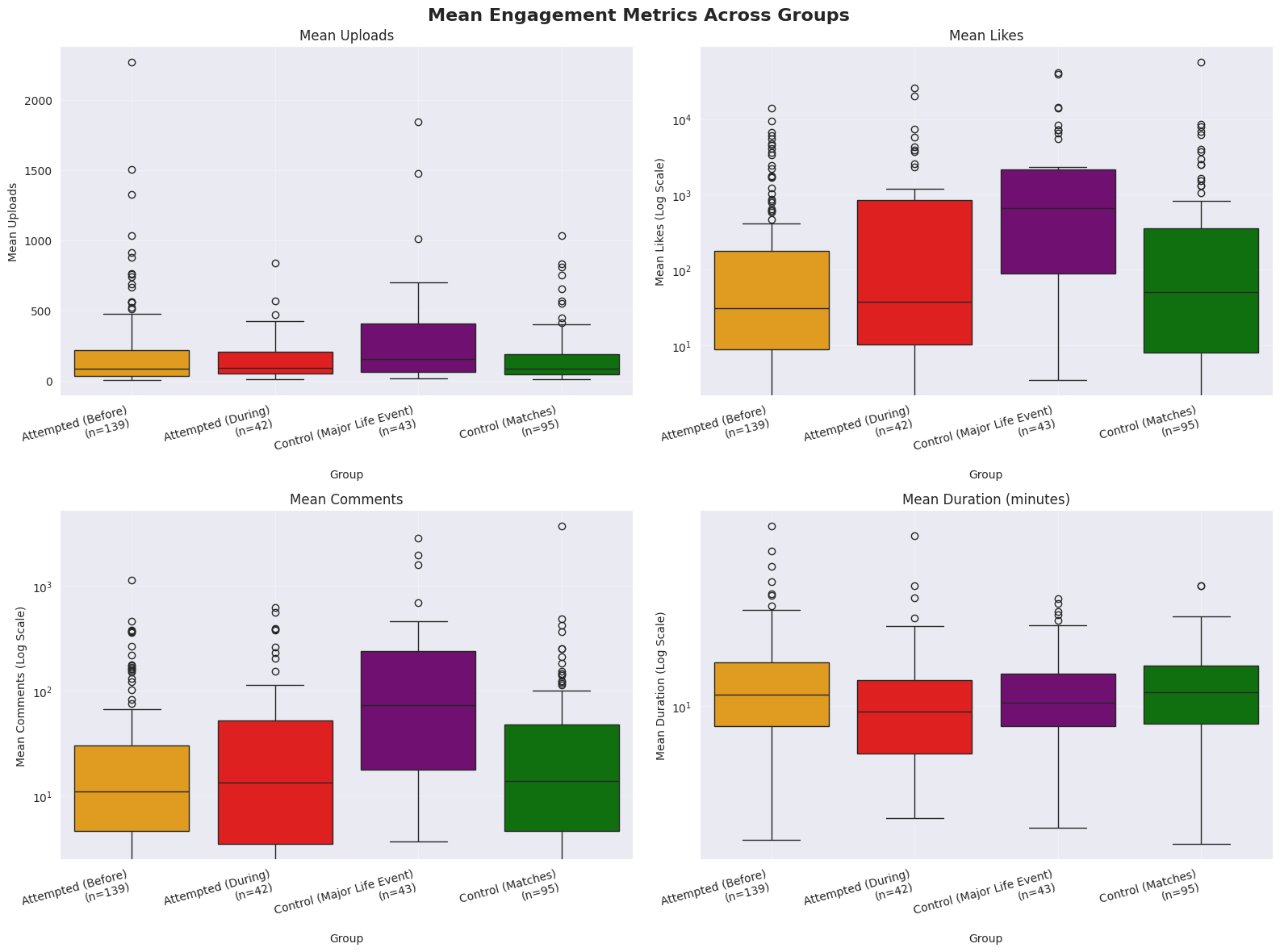}
    \caption{Box plots showing the distribution of uploads, likes, comments, and video duration, aggregated across all channel videos. Each box-plot represents the distribution across users in each group.}
    \label{fig:engagement_boxplots}
\end{figure}

\begin{figure}[h]
    \centering
    \includegraphics[width=0.85\textwidth]{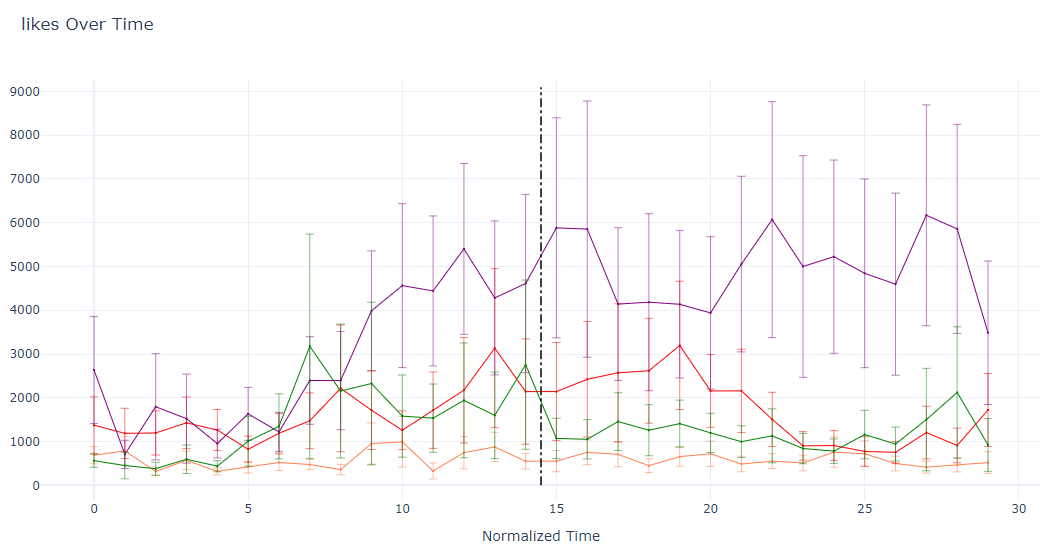}
    \caption{Number of likes over time around the reference event, comparing the four research groups.}
    \label{fig:likes_over_time}
\end{figure}

As illustrated in Figure~\ref{fig:engagement_boxplots}, the distributions of engagement metrics for the \textit{Attempted (During)}, \textit{Attempted (Before)}, and \textit{Control (Matches)} groups were largely similar. The \textit{Control (Major Life Event)} group showed somewhat higher mean values for uploads, likes, and comments, reflecting greater general channel activity. A temporal comparison of likes over time (Figure~\ref{fig:likes_over_time}) further confirmed that these differences were stable, rather than reflecting dynamic changes around the event period. 

\subsubsection{Activity Analysis (Upload Frequency)}

Finally, to assess whether the engagement pattern could be explained by disproportionate posting activity, we analyzed upload behavior at the channel level. Specifically, we examined (a) temporal gaps between uploads before and after the event, and (b) total number of uploads per user in the pre- and post-event periods. This analysis focused on the \textit{Attempted (During)} and \textit{Control (Major Life Event)} groups, as both included individually dated reference events (in contrast to synthetic midpoints used for the other groups).

Figure \ref{fig:temporal_intervals_combined} summarizes the temporal gaps between uploads for both groups. 
We analyzed the number of days between the last upload before the reference event (suicide attempt or major life event), the event itself, and the first upload after the event. In the \textit{Attempted (During)} group, the median interval between the last and first upload surrounding the attempt was 20 days, whereas in the \textit{Control (Major Life Event)} group it was 6 days.
Although some numerical differences were observed, Welch’s t-tests comparing the three interval types (before-to-event, event-to-after, and before-to-after), excluding missing values, revealed no statistically significant group differences, indicating similar variability in posting behavior across the two groups.

\begin{figure}[ht]
    \centering
    \includegraphics[width=0.85\textwidth]{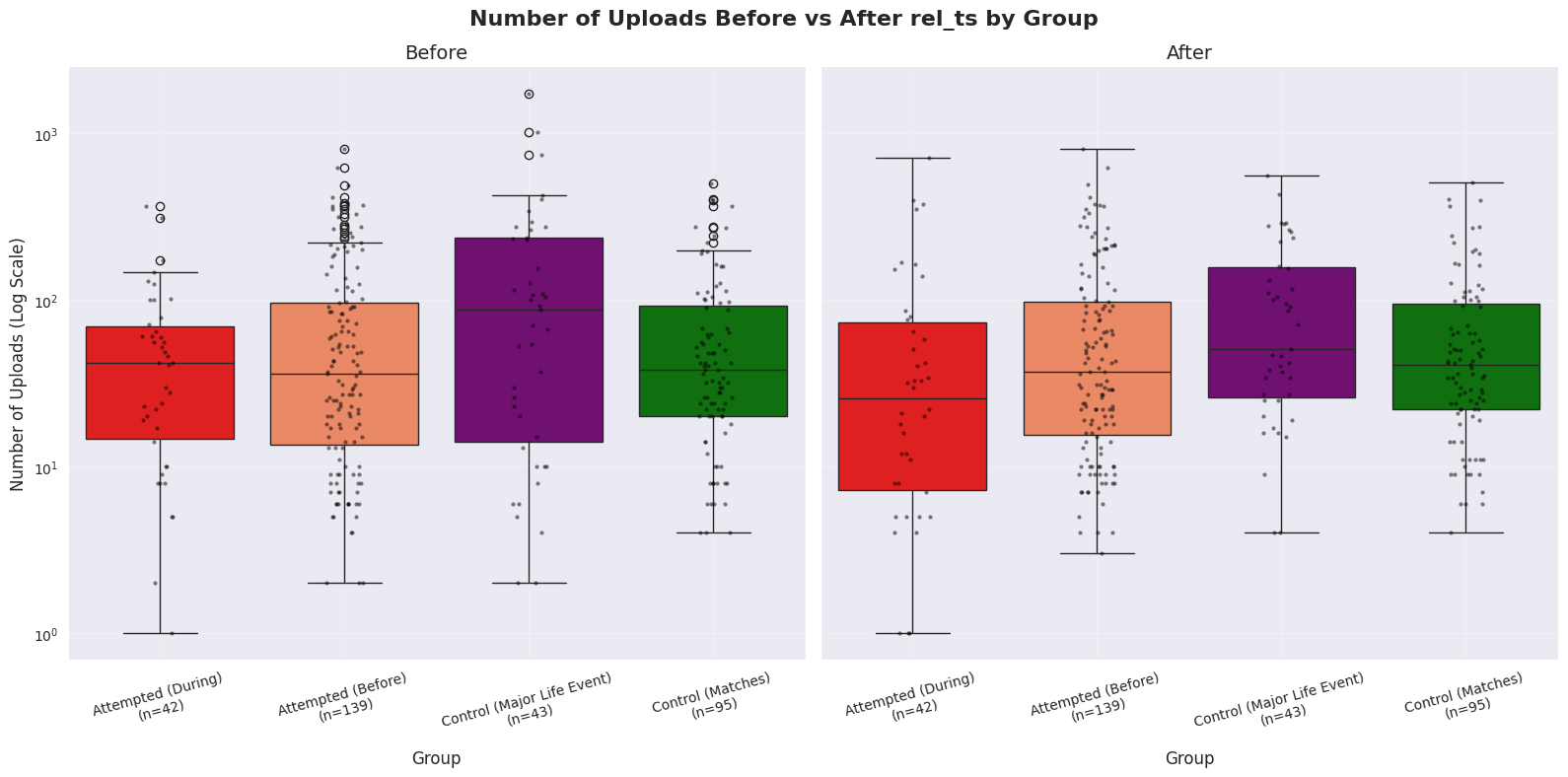}
    \vspace{0.3em}

    \vspace{1em}
    \footnotesize
    \resizebox{0.98\textwidth}{!}{%
    \begin{tabular}{|l|l|c|ccc|ccc|l|}
    \hline
    \textbf{Group} & \textbf{Event Type} & \textbf{\# Users} 
    & \multicolumn{3}{c|}{\textbf{Pre-Event}} 
    & \multicolumn{3}{c|}{\textbf{Post-Event}} 
    & \textbf{Comments} \\ 
    \cline{4-10}
     &  &  & Mean & Median & SD & Mean & Median & SD &  \\ \hline
    \textbf{Attempted (During)} & Suicide Attempt & 42 
    & 65 & 42 & 80 & 84 & 29 & 148 
    & \begin{tabular}[c]{@{}l@{}}4 out of 42 users do not have\\ valid uploads after the attempt\end{tabular} \\ \hline
    \textbf{Attempted (Before)} & Midpoint of Uploads & 139 
    & 102 & 44 & 159 & 98 & 39 & 153 
    & \begin{tabular}[c]{@{}l@{}}Reference event corresponds to the midpoint\\ of the upload period, determined by sorting\\ unique upload dates and selecting the median\end{tabular} \\ \hline
    \textbf{Control (Major Life Event)} & Major Life Event & 43 
    & 192 & 88 & 330 & 119 & 50 & 132 
    & \begin{tabular}[c]{@{}l@{}}Examples include death or suicide attempt\\ of a close relative (parent, child, sibling)\end{tabular} \\ \hline
    \textbf{Control (Personal)} & Midpoint of Uploads & 95 
    & 78 & 42 & 99 & 76 & 42 & 97 
    & \begin{tabular}[c]{@{}l@{}}Synthetic reference midpoint calculated using\\ the median of upload dates\end{tabular} \\ \hline
    \end{tabular}%
    }
    \vspace{0.3em}

    \caption{
    Pre- and post-event upload activity across research groups. 
    Top figure visualizes boxplots total uploads per user before and after the reference event, and the bottom summarizes corresponding descriptive statistics and methodological notes. 
    }
    \label{fig:before_after_uploads_combined}
\end{figure}

\begin{figure}[ht]
    \centering
    \includegraphics[width=0.85\textwidth]{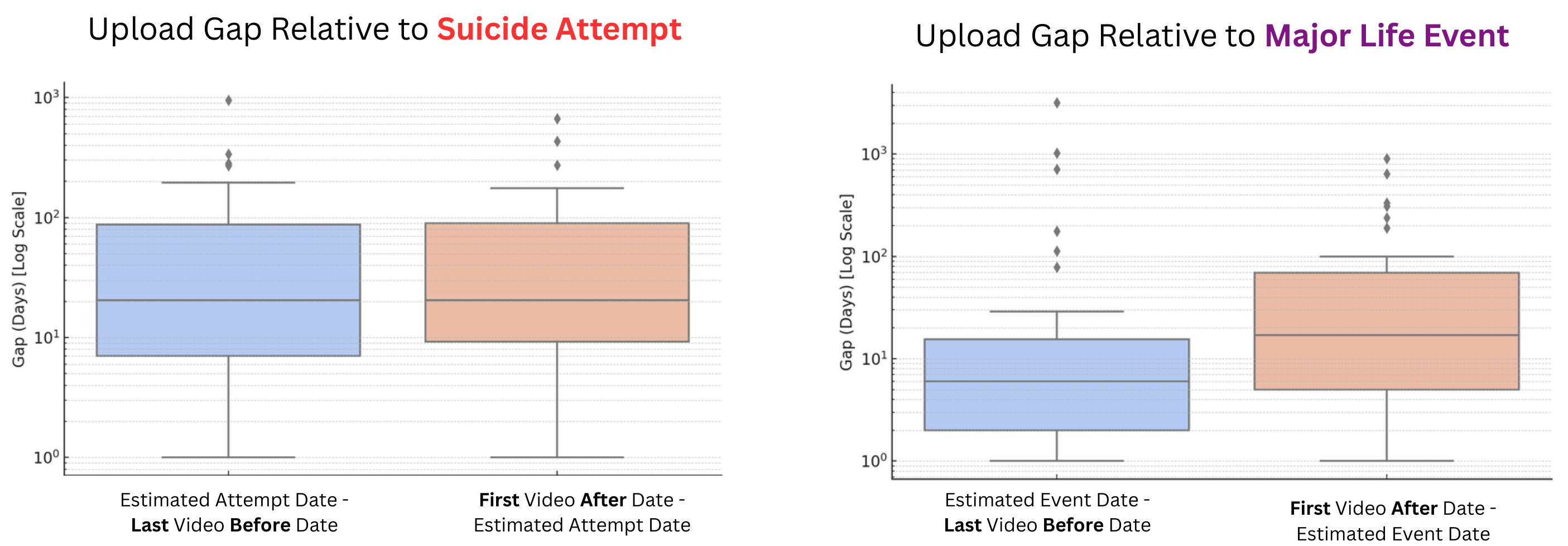}
    \vspace{0.6em}

    \footnotesize
    \resizebox{0.98\textwidth}{!}{%
    \begin{tabular}{|l|ccccc|ccccc|}
    \hline
    \multirow{2}{*}{\textbf{Interval Description}} 
    & \multicolumn{5}{c|}{\textbf{Attempted (During) Group}} 
    & \multicolumn{5}{c|}{\textbf{Control (Major Life Event) Group}} \\ 
    \cline{2-11}
    & \textbf{Mean} & \textbf{Median} & \textbf{SD} & \textbf{Min} & \textbf{Max} 
    & \textbf{Mean} & \textbf{Median} & \textbf{SD} & \textbf{Min} & \textbf{Max} \\ 
    \hline
    Last upload before $\rightarrow$ First upload after 
    & 131.74 & 59.0 & 161.48 & 5 & 702 
    & 214.26 & 39.0 & 565.95 & 2 & 3255 \\ 
    \hline
    Last upload before $\rightarrow$ Event 
    & 83.02 & 20.5 & 160.84 & 1 & 953 
    & 129.77 & 6.0 & 506.89 & 1 & 3187 \\ 
    \hline
    Event $\rightarrow$ First upload after 
    & 72.42 & 20.5 & 129.74 & 1 & 669 
    & 84.49 & 17.0 & 172.63 & 1 & 903 \\ 
    \hline
    \end{tabular}
    }
    \vspace{0.4em}

    \caption{
    Temporal gaps between uploads surrounding reference events. 
    Box plots (top) show the distribution of inactivity intervals for the \textbf{Attempted (During)} and \textbf{Control (Major Life Event)} groups (log scale). 
    The table (bottom) summarizes descriptive statistics for each interval type. 
    The reference event corresponds to a \textit{suicide attempt} in the Attempted group and a \textit{major life event} in the Control group.
    }
    \label{fig:temporal_intervals_combined}
\end{figure}

We also compared the total number of uploads before and after the reference event (Figure~\ref{fig:before_after_uploads_combined}). Across all groups, upload frequency remained relatively stable, with the \textit{Control (Major Life Event)} group showing slightly higher activity, consistent with the elevated engagement metrics observed in \S{\ref{appendix_engagment_metrics}}.

\clearpage
\newpage
\section{Appendix E -  Topic Information} \label{topic_information}
\setcounter{table}{0}
\renewcommand{\thetable}{E.\arabic{table}}
\setcounter{figure}{0}
\renewcommand{\thefigure}{E.\arabic{figure}}

This section includes two tables with prevalent topics identified in this study. Table \ref{tab:topic_info} describes the topics identified in our bottom-up  (\S{3.1}) and hybrid approaches (\S{3.4}). Table \ref{tab:correlation_matrix} presents topics that are highly correlated and have similar semantic meaning. Some information has been reduced or anonymized.

\renewcommand{\arraystretch}{1.1}

{\tiny

\begin{longtable}
    [c]{>{\raggedright\arraybackslash}p{0.05\linewidth}
        >{\centering\arraybackslash}p{0.53\linewidth}
        >{\centering\arraybackslash}p{0.15\linewidth}
        >{\raggedright\arraybackslash}p{0.15\linewidth}}
        
\caption{Overview of topic information, including: \textbf{Count} - the number of transcript segments assigned to each topic, \textbf{Llama2 Name} - the name generated by the Llama-7b model (see Section~\ref{prompts}), \textbf{Representative Transcript Segment} - the transcript segment closest to the topic's center and most representative of it, and \textbf{Top Words} - the 10 words with the highest TF-IDF scores. \textcolor{red}{Red} topic names indicate topics that were identified in our bottom-up approach. \textcolor{blue}{Blue} topics were found in our hybrid approach, and  \textcolor{brown} {brown} indicates topics that were found by both approaches.  Corresponding * indicates significance with regards to temporal changes in Attempted (During).}

\label{tab:topic_info} \\
\toprule
\textbf{Count} & \textbf{Representative Transcript Segment} & \textbf{Llama2 Name} & \textbf{Top Words} \\
\midrule
\endfirsthead

\toprule
\textbf{Count} & \textbf{Representative Transcript Segment} & \textbf{Llama2 Name} & \textbf{Top Words} \\
\midrule
\endhead

\midrule \multicolumn{4}{r}{{\small Continued on next page}} \\
\endfoot

\bottomrule
\endlastfoot
    11815 & quite what to do and I turn to self harm and suicide attempts to cope with mine which was not the best way to cope but I didn't know what else to do, I was hopeless and I just
felt like so horrible and alone and I never want to go back to feeling like that.
You need to think that number one it's not the end of the world, people do get back together,&  \textcolor{brown} {Mental Health Struggles *}  &suicide,self harm,harm, depression, suicidal,bipolar, trauma,self, mental, scars\\
    \midrule

    19822 & again y'all i enjoyed doing this video for y'all hope that you enjoyed watching it
if you made it to the end give me those thumbs up make sure you like comment subscribe to my
channel hit the bell twice to be notified if you're an old subscriber make sure you're
on share my videos and like i always always say god only gives you one life y'all please live it
bye y'all&  \textcolor{red}{YouTube Engagement *} & youtube,subscribers, thumbs, channel,subscribe,videos, hey people, subscribe channel,views,live\\
    \midrule

    19387 & As you can see, like this white part, it's a little bit like, like just like the strings you can like see through it kind of. So, that's pretty awesome. I absolutely love this bikini. And honestly, I didn't think this color would look good on me, but I actually really, really like it. Oh, and by the way, these bags are cute. Like they're very good quality bags and like they look really nice.&  \textcolor{red}{Fashion and Style} &wear,size,dress, jeans,cute,shirt, wearing,really nice,shorts,   pants\\
        \midrule

    6729 & I want to try to come up against the scope. In Jesus' name, Lord, I pray and Lord, I thank you that you get the glory from this scope, God. You get the glory from this Lord, not me. In Jesus' name, I pray. Amen. Okay and ignore my hair. I just got done washing my hair
because my, my dandruff was acting up so don't mind me but as I was, you know, looking& \textcolor{blue} {Religious Prayer and Faith}  &jesus,god, lord, church,  pray,  faith, prayer, praying, jesus lord, spirit, god god, bible, religion,spiritual,verse\\
    \midrule

    10944 & What's up guys? What is life as a cocaine addict like? What is it like to be a cocaine addict? I was addicted to cocaine. My addiction was to the point where it was one to two eight balls of coke a day. If you don't know what an eight ball of coke is, that is 3.5 to almost four grams of coke, depending on if you know the dealer or not and depending on what it's cut with.& \textcolor{blue} {Addiction Recovery}  &sober,nofap, addiction, cocaine, drinking, addict, alcohol,sobriety, relapse,coke\\
        \midrule

    1795& uh feeling tired like I would go to bed early and I just wanted to sleep like I really really
 needed to sleep week nine day three I was so happy because I've been waking up really late
 every day like almost 9 a.m because I feel so tired and so fatigued I made my bed this
 morning for the first time since like a week because I I didn't even have energy to make the& \textcolor{blue} {Sleep Disorders and Habits} &sleep,wake, waking, bed, morning, sleeping, morning routine, hours sleep, sleep schedule, hours\\
    \midrule

 3363& Just look at it, man. Right now what we have going on on the internet, we got black men against black women.
 Then on the streets we got black men against black men, and then black women against black women.
 We all talk about wanting unity, but where is it? We definitely have to do better.& \textcolor{blue} {Black community} &black,black women,black men,black people,black woman,white,racist,black man,white people,women\\
    \midrule

    4124& If you like this video, I would love it if you'd give it a thumbs up.
 And if you're interested in all things autism,
 I'd love it if you'd hit subscribe and I'll see you tomorrow. Bye.& \textcolor{blue} {Autism and Disability}& autism, autistic, trans, transgender, disabled, things autism,interested things, autism love, thumbs interested, love hit\\
    \midrule

    952& A lot of people just say that I'm gay or they assume I'm gay when I'm bi.
They're like oh no you can't be bi you're so gay.
I'm bi I think.
Seven what's your favorite thing about the LGBT community?
I just love how we're all really caring for each other.
There are hateful people but most 
of the LGBT community is really loving 
and accepting & \textcolor{brown} {LGBTQ+ Identity and Acceptance} \textcolor{red}{*} &gay,bisexual,lesbian,lgbtq,
sexuality,gay people,homophobic,queer,
pansexual,lgbtq community,lgbt,
straight,gay gay,bi,gays\\
    \midrule

    221 & Okay, yes, I know that there are a lot of women out here who are falsely accusing men of sexual assault and rape.
I know this, I know this, I know this, but don't go into the situation thinking that's what's going on.
You have to believe a female when she says that she's been raped.& \textcolor{brown} {Rape and Sexual Assault} \textcolor{red}{*} &rape,raped , consent,considered rape,rape know,  consented, victims,rapist,assault,sexual assault, men raped,rape\\
    \midrule

    1041&  I'm going to pop in something really sour into my mouth, or I'm going to smell
 something that's going to get my brain out of here.
 Any tips on, uh, anxiety panic attacks, panic attacks are hard to ground yourself
 from and they're hard to stop yourself.
 There is huge differences between anxiety attacks and panic attacks.
 One of the biggest ones is there's no trigger necessarily behind a panic attack.& \textcolor{blue} {Anxiety and Panic Attacks} &anxiety,panic,attacks,panic attacks,panic attack,attack,anxiety attack,anxious,anxiety attacks,anxiety like\\
    \midrule

    763& and year 9, not a lot of people knew about me.
 I was just kind of this loser.
 I was called emo, but no one really bullied me and even if people tried to, it never
 fazed me.
 I've never been bullied in that sense.
 When I was in school I've never been bullied and it doesn't mean that other people haven't
 tried to bully me, it's just that I don't take it as bullying.& \textcolor{blue} {Bullying}& bullied, bullying, bully, bullies, like bullied, bullied like, getting bullied, ve bullied, school, cyber\\
    \midrule

    194& Just tell them what you just told us.
 Okay.
 This is a situation that most people don't see when they see the homeless people.
 They see us as a problem.
 You have homeless people that are on the street as well as in the shelters,
 but the shelters don't usually work out for us, so we end up coming back out to the street.
 If you get in an argument with somebody, you're kicked out of the shelter,& \textcolor{blue} {Homelessness}&homeless,homeless people,homeless person,people homeless,say homeless,homeless like,homeless population,tourists,know homeless\\
        \midrule

    1262& finding out what works for you when it comes to the time of medication because
 you can hear horror stories of other people with Adderall I mean I know
 people that have said that they popped Adderall like Skittles and I don't
 comprehend that because I take 10 milligrams in the morning I take 10
 milligrams in the afternoon and there's other people out there that are like you& \textcolor{blue} {Medication Use and Addiction} & adderall, xanax, medication, effects, prescribed, milligrams, 10 milligrams, adhd, ritalin, prozac\\
    \midrule

    250&  there's signs out front that say, you know, a vet lives here, please be cautious and courteous with
 your fireworks. For some, it's slamming doors, it's loud voices, it's slamming objects. That
 triggers your flashback. That sparks and flares up their PTSD. There is that big difference. So
 let me really break down the differences between the two. PTSD, one or few trauma, short-lived
 trauma. Visual and somatic flashbacks, avoidance of triggers, isolating, nightmares about trauma,& \textcolor{blue} {Trauma-Induced Mental Health Disorders}& ptsd, flashbacks, cptsd, triggers, smell blood, emdr therapy, emdr, complex ptsd, ptsd ptsd\\
        \midrule

    192&   Honest chat.
 Honest chat.
 So one of you actually asked the question,
 what is BPD and is there a way to get over it?
 So I thought that would be a good starting point to explain what BPD is.
 BPD is what I've been diagnosed with along with PTSD, anxiety and depression.
 BPD is borderline personality disorder.
 It can also be called EUPD, emotionally unstable personality disorder.
 Personally I think that's a little bit offensive.& \textcolor{blue} {Living with Borderline Personality Disorder (BPD)} & bpd, people bpd, personality disorder, borderline personality, borderline, personality, disorder, bpd like, diagnosed bpd, diagnosed\\
    \midrule

 3363&  So, how are you today?
 I'm doing okay.
 So I can tell you a little bit about Obsessive-Compulsive Disorder, or OCD.
 It's a mental illness and it's one of the anxiety disorders.
 And so there's two parts. There's the obsessions and there's the compulsions.
 So the obsessions, those are repetitive and intrusive thoughts.
 It can be urges or images that cause a lot of anxiety in an individual.
 And then the second part are the compulsions.& \textcolor{blue} {Obsessive-Compulsive Disorder (OCD)} & ocd, compulsions, intrusive, intrusive thoughts, people ocd, ocd like, obsessions, ocd just, obsessive compulsive, obsessive\\
    \midrule

    1036&  Hi guys welcome to ... I'm recovering from an eating
 disorder...  I'm recovering from an eating disorder. I've been in this chair for like two. Yeah well it won't be two hours after editing.& \textcolor{blue} {Eating Disorder Recovery} & [this topic contains private names]\\
    \midrule

    219& She becomes a rebellious sure goes off to college and Cal basically just goes to high school and does his own thing .. 
 It's so weird ... I would say traumatized a little bit& \textcolor{blue} {Loss and Trauma in a Small Community} & [this topic contains private names]\\
    \midrule

    186& I make an effort to never do anything like that.
 I do not want to set unrealistic standards for you guys.
 And when I say unrealistic, I don't mean that those things aren't attainable.
 What I mean is that you shouldn't be aiming for that.
 You know, you shouldn't be striving for perfection because perfection doesn't exist.
 Your imperfections are what make you perfect.& \textcolor{blue} {Perfectionism and the Pursuit of Imperfection}& perfect, perfection, imperfect, imperfections, perfectionism, perfect perfect, perfectionist, know perfect, perfect person, perfect don
\end{longtable}
}
\begin{table}[h]
\centering
\scriptsize
\caption{Correlation Matrix of 30 Topics Based on Cosine Similarity: The matrix, computed using cosine similarities of the topic embeddings, demonstrates that topics with high correlation are also semantically similar, as anticipated.}

\begin{tabular}{>{\raggedright\arraybackslash}p{0.4\linewidth}>{\centering\arraybackslash}p{0.4\linewidth}>{\centering\arraybackslash}p{0.15\linewidth}}
    \toprule
    Topic 1 & Topic 2 & Correlation \\
    \midrule
    Online Therapy & Mental Health Struggles & 0.756 \\
    Mental Health Struggles & Online Therapy & 0.756 \\
    Singing and Going & Mental Health Struggles & 0.750 \\
    Mental Health Struggles & Singing and Going & 0.750 \\
    YouTube Engagement & Vlog End and Daily Uploads & 0.744 \\
    Vlog End and Daily Uploads & YouTube Engagement & 0.744 \\
    Underage Smoking and Drug Use & Addiction Recovery & 0.743 \\
    Addiction Recovery & Underage Smoking and Drug Use & 0.743 \\
    Social Media Addiction & Mindful Social Media Use & 0.742 \\
    Mindful Social Media Use & Social Media Addiction & 0.742 \\
    Tumultuous online interaction & Loss and Trauma in a Small Community & 0.741 \\
    Loss and Trauma in a Small Community & Tumultuous online interaction & 0.741 \\
    Mental Health Awareness & YouTube Engagement & 0.728 \\
    YouTube Engagement & Mental Health Awareness & 0.728 \\
    Anxiety and Panic Attacks & Fears and Growth & 0.725 \\
    Fears and Growth & Anxiety and Panic Attacks & 0.725 \\
    Mental Health Awareness & Mental Health Struggles & 0.723 \\
    Mental Health Struggles & Mental Health Awareness & 0.723 \\
    Self-Confidence and Overcoming Adversity & Self-perception and beauty standards & 0.718 \\
    Self-perception and beauty standards & Self-Confidence and Overcoming Adversity & 0.718 \\
    YouTube Engagement & Social media content creation and scheduling & 0.716 \\
    Social media content creation and scheduling & YouTube Engagement & 0.716 \\
    Yoga Practice and Philosophy & Mindfulness and Meditation & 0.715 \\
    Mindfulness and Meditation & Yoga Practice and Philosophy & 0.715 \\
    Christmas Tree Decorating & Toxic Christmas & 0.711 \\
    Toxic Christmas & Christmas Tree Decorating & 0.711 \\
    Vaccine Beliefs and Misconceptions & COVID-19 Pandemic & 0.709 \\
    COVID-19 Pandemic & Vaccine Beliefs and Misconceptions & 0.709 \\
    Mental Health Struggles & Emotional Resistance & 0.709 \\
    Emotional Resistance & Mental Health Struggles & 0.709 \\
    \bottomrule
\end{tabular}
\label{tab:correlation_matrix}
\end{table}

\clearpage
\end{document}